\documentclass[11pt]{article}

\usepackage[preprint]{acl}

\usepackage{microtype}

\usepackage[english,bidi=default]{babel} 
\babelfont{rm}{TeXGyreTermesX} 
\babelprovide[import]{hindi}
\babelfont[*devanagari]{rm}{Lohit Devanagari}
\babelprovide[import]{arabic}
\babelfont[*arabic]{rm}{Noto Sans Arabic}

\usepackage{inconsolata}

\usepackage{booktabs}    
\usepackage{multirow}   
\usepackage{graphicx}    
\usepackage{siunitx}
\usepackage{amsmath}
\usepackage{cleveref}
\usepackage{makecell} 
\usepackage{hyperref}
\usepackage[dvipsnames]{xcolor}
\usepackage{pifont}
\usepackage{enumitem}

\usepackage[utf8]{inputenc}
\usepackage[most]{tcolorbox} 
\usepackage{listings}
\usepackage{xcolor}
\tcbset{
    promptbox/.style={
        enhanced,
        colback=white,
        colframe=black,
        colbacktitle=gray!20,
        coltitle=black,
        fonttitle=\bfseries,
        width=0.9\linewidth, 
        center,             
        boxrule=0.5mm,
        arc=3mm,
        left=1mm,
        right=1mm,
        top=1mm,
        bottom=1mm,
        boxsep=5pt,
    }
}
\tcbset{
casebox/.style={
enhanced,
colback=white, 
colframe=black, 
colbacktitle=gray!20, 
coltitle=black,
fonttitle=\bfseries, 
fontupper=\normalsize, 
width=\textwidth, 
boxrule=0.5mm, 
arc=3mm, 
left=1mm, right=1mm, top=1mm, bottom=1mm, 
boxsep=5pt, 
rounded corners, 
}
}
\usepackage{xeCJK}
\usepackage{xcolor} 
\usepackage{booktabs}
\usepackage{multirow}
\usepackage{amsmath}
\usepackage{colortbl}
\usepackage{adjustbox}
\usepackage[normalem]{ulem}
\usepackage{cuted}
\usepackage{afterpage}
\usepackage{amssymb}
\usepackage{float}
\usepackage{caption}
\usepackage{subcaption} 
\usepackage{enumitem} 
\usepackage{amssymb}

\title{LongNovel: A Multi-Scale Benchmark for Hallucination Detection in Long-Context Novel Summarization}

\newcommand\blfootnote[1]{%
  \begingroup
  \renewcommand\thefootnote{}\footnote{#1}%
  \addtocounter{footnote}{-1}%
  \endgroup
}

\usepackage{amssymb} 

\author{
  \textbf{Ruizhi Zhang}$^{1,2\P\maltese}$ \quad 
  \textbf{Jinwei Chen}$^{1\dagger\maltese}$ \quad 
  \textbf{Xiangju Lu}$^{2\P\maltese}$ \\ 
  \textbf{He Yan}$^{2\P}$ \quad 
  \textbf{Mo Yu}$^{3\ddagger}$ \quad 
  \textbf{Junmin Zhu}$^{2\P}$ \quad 
  \textbf{Wei Zhang}$^{1,4\S\ast}$ \\[1ex]
  \small $^1$East China Normal University \quad $^2$iQIYI Inc \quad $^3$Tencent \quad $^4$Shanghai Innovation Institute \\
  \small \texttt{$^{\dagger}$51285901033@stu.ecnu.edu.cn} \quad
  \small \texttt{$^{\P}$\{zhangruizhi\_sx, luxiangju, yanhe, zhujunmin\}@qiyi.com}, \\
  \small \texttt{$^{\ddagger}$moyumyu@global.tencent.com}, \quad
  \small \texttt{$^{\S}$zhangwei.thu2011@gmail.com}
}

\begin{document}
\maketitle
\blfootnote{$\maltese$ Equal contribution.}
\blfootnote{$\ast$ Corresponding author.}

\begin{abstract}
Although context windows have expanded significantly in recent years, hallucinations in long-context summarization remain a challenge. Long novels are better suited than news or papers for researching these hallucinations, due to their intrinsic information and detailed descriptions of events and dialogues. However, current research lacks a multi-scale benchmark for hallucination detection in long-context novel summarization and does not fully explore how hallucinations change as the context grows longer. In this study, we propose LongNovel, a multi-scale long-context bilingual (Chinese and English) novel benchmark for hallucination detection. This benchmark is constructed from 29 Chinese novels (ranging from 16k to 100k tokens) and chapter-level data from the BookSum dataset. We design 8 hallucination types and employ a combination of Multi-Model Arbitration and Entity-Referenced Hallucination Generation to ensure both data authenticity and a balanced distribution of hallucination categories. Furthermore, we manually revise the content in the test set to guarantee data reliability. Extensive experimental results demonstrate that LongNovel is a challenging benchmark. We release LongNovel for future research.\footnote{\url{https://github.com/BDML-lab/LongNovel}}
\end{abstract}

\section{Introduction}

While the expansion of context windows for Large Language Models (LLMs) to 100k tokens or more~\cite{chen2024longlora,peng2023yarn,ding2024longrope} has enabled the processing of long-form content, this increased capacity does not inherently resolve the issue of hallucinations in long-context summarization~\cite{fables-2024-kim-et-al, BelemPIMBH25FromSingletoMulti, Giraffe2023}. Novel summarization is well-suited for researching hallucinations in long-context summarization because it requires inferring implicit information from dialogues and events, which is more complex than processing the explicit data found in news or academic papers~\cite{karpinska-etal-2024-nocha,kryscinski-etal-2022-booksum, Kim2025NexusSum, diahalu}. While traditional metrics like ROUGE~\cite{lin-2004-rouge} and BERTScore~\cite{Tianyi2020bertscore} are limited to lexical or semantic similarity, other NLI-based approaches such as SummaC~\cite{laban-etal-2022-summac} and AlignScore~\cite{ZhaYLH23AlignScore} often fail to detect long-context hallucinations due to their limitation to short input windows. Consequently, there is an urgent need for more precise evaluation models that can identify hallucinations in long-context scenarios. However, the development of robust evaluation models relies on the availability of high-quality benchmarks. Therefore, constructing a long-text, multi-scale hallucination detection dataset will facilitate the identification of more reliable evaluation models.

However, existing research faces two primary challenges. First, constructing high-quality benchmarks for hallucination detection in long-context novels is hindered by the cost of manual annotation. Traditional datasets such as NOCHA~\cite{Karpinska2024nocha} and StorySumm~\cite{Subbiah24STORYSUMM} rely on human-labeled hallucination data, which is labor-intensive and time-consuming. While synthetic datasets like LCHD~\cite{Liu2025LCHD} effectively reduce annotation costs, they fail to reach the 100k-token scale.
Second, the evolution of hallucinations as context length increases remains largely unexplored. Most existing benchmarks lack a multi-scale design capable of evaluating model robustness across varying lengths. While Fables~\cite{fables-2024-kim-et-al} covers the 100k-token scale, it is largely restricted to a single length. Although Clipper~\cite{Pham2025CLIPPER} introduces a multi-scale approach with book-level and chapter-level claims at the 100k scale, it is not designed for summary hallucination detection.

\begin{figure*}[t]
  \centering

  \includegraphics[width=1\textwidth]{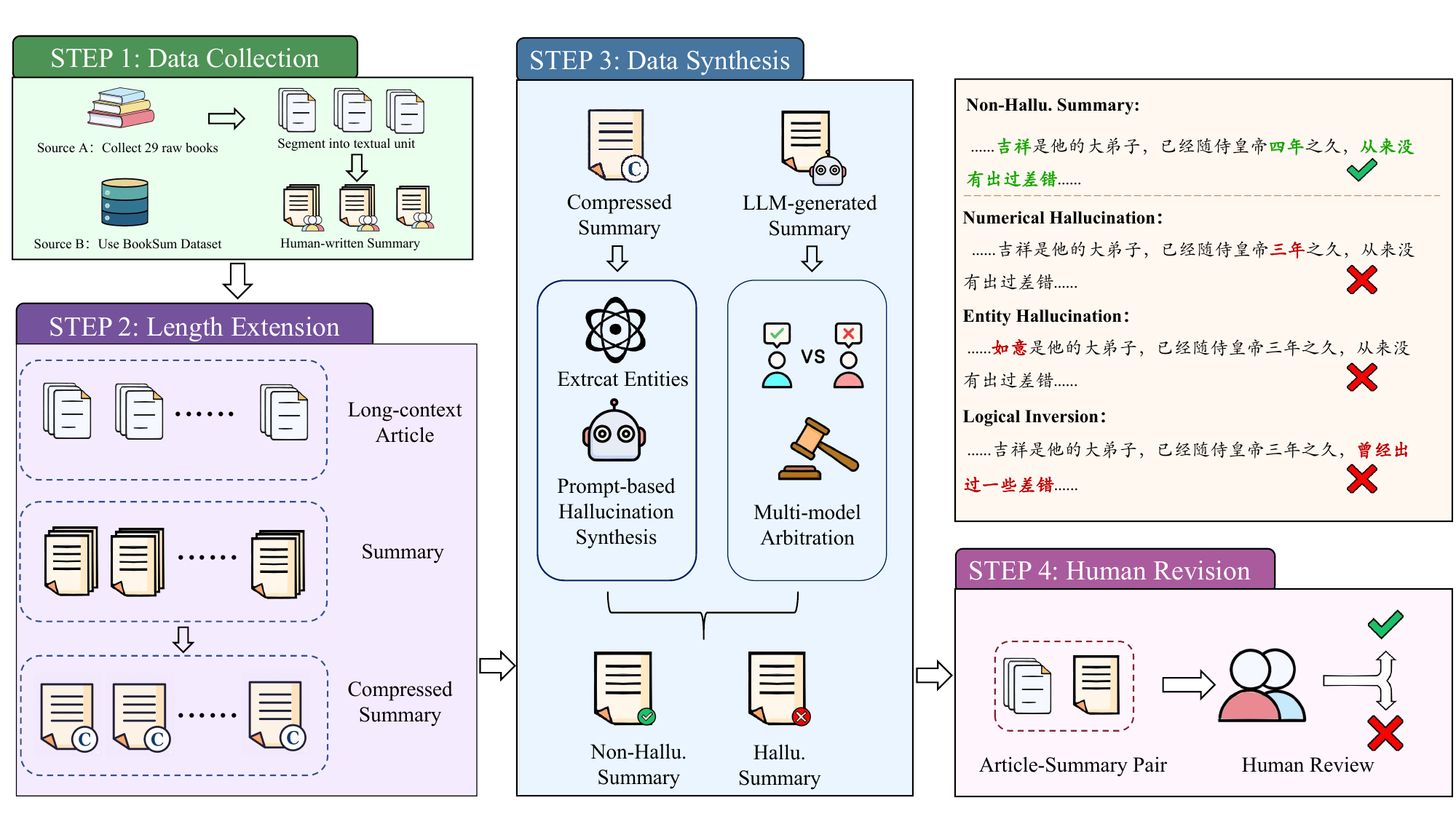}
  \caption{The framework of LongNovel.}
    \label{fig:The framework of LongNovel.}
  \vspace{-2ex}
\end{figure*}

\begin{table}[!t]
\centering
\resizebox{\linewidth}{!}{
\begin{tabular}{lcccc}
\toprule
\textbf{Benchmark} & \makecell[c]{\textbf{Summ.}\\\textbf{Halluc.}} & \textbf{100K} & \makecell[c]{\textbf{Auto.}\\\textbf{Label}} & \makecell[c]{\textbf{Diff.}\\\textbf{Len.}} \\ \midrule
Nocha (\citeyear{karpinska-etal-2024-nocha})              & \textcolor{red}{\ding{55}}      & \textcolor{ForestGreen}{\ding{51}}   & \textcolor{red}{\ding{55}}          & \textcolor{red}{\ding{55}}           \\
StorySumm (\citeyear{Subbiah24STORYSUMM})            & \textcolor{ForestGreen}{\ding{51}}   & \textcolor{red}{\ding{55}}        & \textcolor{red}{\ding{55}}         & \textcolor{red}{\ding{55}}            \\
FABLES (\citeyear{fables-2024-kim-et-al})                & \textcolor{ForestGreen}{\ding{51}}   & \textcolor{ForestGreen}{\ding{51}}   & \textcolor{red}{\ding{55}}         & \textcolor{red}{\ding{55}}           \\
LCHD (\citeyear{Liu2025LCHD})                  & \textcolor{ForestGreen}{\ding{51}}   & \textcolor{red}{\ding{55}}       & \textcolor{ForestGreen}{\ding{51}}     & \textcolor{red}{\ding{55}}           \\
CLIPPER (\citeyear{Pham2025CLIPPER})               & \textcolor{red}{\ding{55}}        & \textcolor{ForestGreen}{\ding{51}}   & \textcolor{ForestGreen}{\ding{51}}     & \textcolor{ForestGreen}{\ding{51}}       \\ \midrule
\textbf{LongNovel (Ours)}      & \textcolor{ForestGreen}{\ding{51}}   & \textcolor{ForestGreen}{\ding{51}}   & \textcolor{ForestGreen}{\ding{51}}     & \textcolor{ForestGreen}{\ding{51}}       \\ \bottomrule
\end{tabular}}
\caption{Comparison of our benchmark with other benchmarks in novel. `Summ. Halluc.', `100K', `Auto. Label', and `Diff. Len.' mean whether it is a summarization hallucination detection dataset, whether it reaches up to 100K tokens, whether the hallucinated data is generated through automated methods, and whether it encompasses different levels of length, respectively. LCHD~\cite{Liu2025LCHD} refers to the long-context hallucination detection dataset.}
\label{benchmark-comparison}
\vspace{-1ex}
\end{table}
To address these issues, we introduce \textbf{LongNovel}, a multi-scale benchmark for hallucination detection in long-context novel summarization. Constructed from 29 Chinese novels and BookSum English chapters, we construct four long context scenarios: S(16k), M(32k), L(64k), and XL(100K), totaling 600 samples in the test set. Building on these scenarios, eight hallucination types have been designed. Notably, we use human-written summaries as ground truth to guide the LLM generation to ensure data reliability. Each judgment includes a consistency score of the summary, the identified hallucination types, and reasons for identifying the hallucinations.
The framework of our benchmark is illustrated in Fig.~\ref{fig:The framework of LongNovel.}. 
To balance data authenticity with a uniform distribution of hallucination types, we employ two complementary construction methods. The first is Multi-Model Arbitration, which utilizes GLM4-9B-chat~\cite{glm2024chatglmfamilylargelanguage}, Qwen3-32B~\cite{yang2025qwen3technicalreport}, and GPT-4o~\cite{openai2024GPT-4O} for summary generation, followed by a cross-model verification process for data labeling to capture authentic hallucinations in LLM outputs. The second method, Entity-Referenced Hallucination Generation, perturbs human-written summaries by extracting entities and employing LLMs to craft hallucinations based on eight specific prompt templates, thereby ensuring a balanced distribution across all targeted hallucination types. Finally, to guarantee data reliability, we conduct a human revision of the test set. Compared with previous long-context novel benchmarks, LongNovel offers several distinct advantages, as outlined in Table~\ref{benchmark-comparison}. We evaluate several state-of-the-art models and conduct experiments using different methods on LongNovel. The results show that LongNovel serves as a challenging benchmark for current models. Overall, our contributions are as follows:

\begin{itemize}[nosep, left=0pt, itemsep=1pt, topsep=1pt]
   \item We introduce LongNovel, a multi-scale Chinese-English bilingual benchmark designed for hallucination detection in long-context novel summarization. It spans four levels of length and covers eight distinct hallucination types.
    \item We implement a construction approach that combines Multi-Model Arbitration with Entity-Referenced Hallucination Generation. This methodology ensures the dataset features real-world authenticity while covering various hallucination types.
    \item We evaluate state-of-the-art models and methods on LongNovel, revealing the challenges of long-context hallucination detection and providing a challenging benchmark for future research.
\end{itemize}

\begin{figure*}[t]
  \centering
  \includegraphics[width=0.9\textwidth]{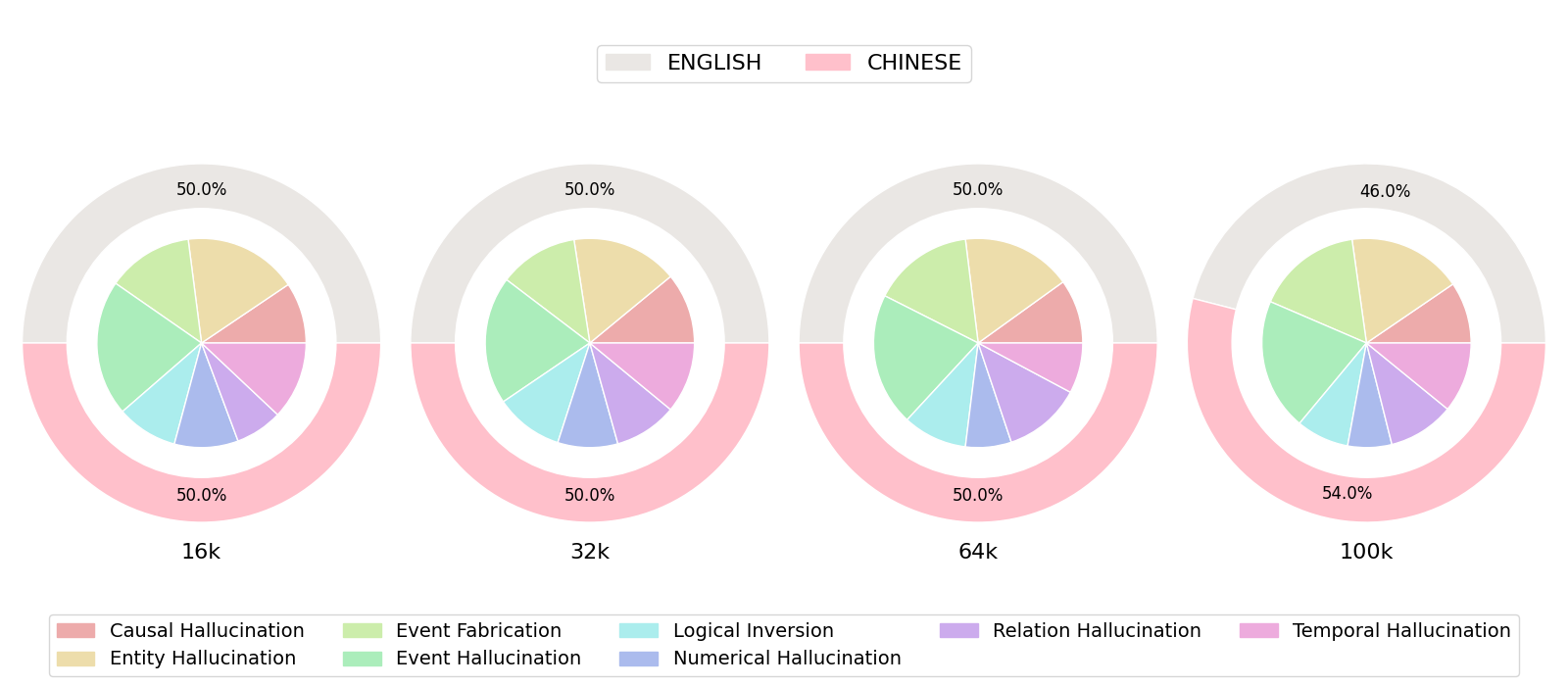}
  \caption{Distribution of hallucination types across different context lengths.}
    \label{fig:Distribution of hallucination types across different context lengths.}
\end{figure*}

\section{RELATED WORK}

\subsection{Hallucination Detection Benchmarks}
Current methodologies for constructing hallucination datasets can be categorized into two approaches. The first involves generation by LLM, followed by manual annotation to identify hallucinated samples~\cite{LabanKAFXJW23SummEdits, karpinska-etal-2024-nocha, Subbiah24STORYSUMM, AkbarHWCSAC24HalluMeasureTechNewsSumm, TangSWBVYS0SSSZ24TofuEval, Chen24FactCHD, BaoLQLWTFTKSQTXGMA25FaithBench, Abdaljalil25HalluVerse25}. The advantage of this approach is that the resulting hallucination patterns closely align with the model's performance in the real world. However, it relies heavily on high-quality annotation, making it both time-consuming and resource-intensive~\cite{QiC0Y25FHSumBench}.

The second approach is automated hallucination injection based on existing reference materials, such as books or summaries. Early methods like~\cite{KryscinskiMXS2020factcc} generate negative samples by employing entity substitution and negation insertion. \citet{Cao021CLIFF} select system outputs with low likelihood scores as negative samples. Recent works perform hallucination synthesis based on the instruction-following capabilities of LLMs~\cite{TangLD24MiniCheck, Liu2025LCHD, MingPPKNXJ25FaithEval}. \citet{Pham2025CLIPPER} 
extract summaries or outlines from original book content, thereby inducing LLMs to generate true-false claim pairs and corresponding reasoning chains.

\subsection{Hallucination Detection Methods}
Existing hallucination detection methods can be categorized into short-text and long-text detection based on the length of the processed content. In the short-text detection,~\citet{laban-etal-2022-summac} evaluate factual consistency by decomposing documents into sentence pairs and computing NLI-based entailment scores. \citet{ZhaYLH23AlignScore} enhance cross-task generalization through large-scale alignment pre-training, and~\citet{LiuF0LJWXR23ACU} introduce an evaluation framework based on Atomic Content Units. \citet{kedichenenhancinguncertainty} propose a semantic graph-based approach to capture the intricate relations between entities and sentences, thereby enhancing hallucination detection at both sentence and passage levels. Additionally, QA-based methods~\cite{DeutschBR21QAEVAL, ScialomDLPSWG21QuestEval} verify informational faithfulness by measuring the answer consistency between source texts and generated summaries.

In long-text hallucination detection, many short-text methods are constrained by input window limitations. Approaches address this by either directly leveraging LLMs for binary classification or employing Chain-of-Thought (CoT)~\cite{Wei2022Chain-of-thought} to provide step-by-step analysis before reaching a final judgement. \citet{liu-etal-2023-geval} utilize LLMs to score content based on preset indicators, demonstrating a high correlation with human judgment. \citet{min-etal-2025-MSumBench} introduce a debate-based framework by assigning specific roles to LLMs, such as Advocate, Skeptic, and Adjudicator, to enhance data reliability. RAG-based systems, which are used to verify faithfulness in QA tasks~\cite{zhang-etal-2025-longcite, HuZJZW025Debate-Augmented, LabanFXW24SummaryofaHaystack}, can also detect hallucinations in summarization~\cite{MinKLLYKIZH23FActScore}. 
\section{LongNovel Construction}
\subsection{Data Collection}
We construct LongNovel using both Chinese and English literary sources. For the Chinese corpus, we collect 29 books from open-source data on the Chinese internet. Each book possesses a coherent plot, making it highly suitable for long-text consistency detection. Each book $B = \{u_1, u_2, \dots, u_n\}$ is segmented into a series of textual units (chapters or paragraphs), where the length of each unit $u_i$ ranges from $2\text{k}$ to $6\text{k}$ tokens. We employ 16 annotators. For each textual unit, one annotator drafts an initial summary, which is then revised by two other annotators to ensure accuracy and faithfulness to the source text. For the English corpus, we directly adopt the chapter-level subset from BookSum~\cite{kryscinski-etal-2022-booksum} and treat each chapter as a textual unit $u_i$, leveraging its high-quality human-written summaries. Ultimately, each textual unit $u_i$ from both languages is paired with a corresponding human summary, denoted as $s_i$.

\subsection{Length Extension}
To comprehensively evaluate model performance across different context windows, we define four distinct target lengths $T \in \{16\text{k}, 32\text{k}, 64\text{k}, 100\text{k}\}$ along with their corresponding evaluation ranges $\mathcal{R}(T)$:
\begin{equation}
\mathcal{R}(T) = [T - 2\text{k}, T + 4\text{k}]
\end{equation}
 We employ the Qwen3 \cite{yang2025qwen3technicalreport} tokenizer to compute token counts. The book $B$ is composed of multiple textual units $u$, represented as $B = \{u_1, u_2, \dots, u_n\}$. For a target range $\mathcal{R}(T)$ and a book $B$, the sampling method is as follows:

\noindent1) Initialize a sliding window $w$ starting from $u_1$, expanding it unit by unit.

\noindent2) When $w = [u_j, u_{j+1}, \dots, u_{k-1}, u_k]$ ($1 \le j \le k \le n$):
    \begin{enumerate}[label=\roman*.,ref=\roman*,itemsep=0.001em, topsep=0.5em]
        \item \label{step1} If $\text{Len}(w) < \inf \mathcal{R}(T)$ and $k < n$, continue expanding $w$ with the next textual unit $u_{k+1}$.
        \item \label{step2} If $\text{Len}(w) \in \mathcal{R}(T)$, gather the current sequence in $w$ as a data point and concatenate the corresponding summaries, denoted as $S_w = [s_j, s_{j+1}, \dots, s_k]$. Then, reset the window by setting $w = [u_{m+1}]$, where $m = \lfloor \frac{j+k}{2} \rfloor$.
   \end{enumerate}
    
\noindent3) If the window expands to the end of a book but the remaining segments still fail to satisfy the lower boundary, this residual sequence is discarded before processing the next book.

We utilize the Gemini-3-Pro-Preview model to compress texts from these target windows into lengths of $900, 1000, 1200,$ and $1500$ words, respectively. To guarantee quality, a feedback mechanism evaluates the initial summary against a ground-truth reference to assign a consistency score and provides a detailed reason. If the score falls below $4$ out of $5$, the model performs a second-round generation incorporating automatically pinpointed deficiencies and refinement suggestions. As $w$ spans consecutive chapters, our benchmark preserves narrative continuity. Consequently, error detection requires long-range, cross-chapter verification rather than local fact-checking (Appendix~\ref{app:cross_chapter_case}).

\subsection{Data Synthesis}
We categorize hallucinations into eight types: Entity Hallucination, Numerical Hallucination, Relation Hallucination, Logical Inversion, Event Hallucination, Temporal Hallucination, Causal Hallucination, and Event Fabrication (detailed in Appendix~\ref{app:Hallucination_Type}). 
To ensure data authenticity and a balanced distribution of hallucination types, we employ two methods to generate our LongNovel benchmark.

\paragraph{Multi-Model Arbitration.} To obtain more realistic hallucination data, inspired by MSumBench~\cite{min-etal-2025-MSumBench}, we implement a Multi-Model Arbitration method. In this method, summaries are first generated by GLM-4-9B-chat~\cite{glm2024chatglmfamilylargelanguage}, Qwen3-32B~\cite{yang2025qwen3technicalreport}, and GPT-4o~\cite{openai2024GPT-4O}, ensuring that \textsc{LongNovel} captures natural failure modes. During this process, GPT-4.1 and Claude-sonnet-4-20250514-v1 first conduct independent evaluations. Instead of a direct binary classification, the models are required to assign consistency scores and justify their ratings, thereby preventing over-simplified summaries from being misclassified as hallucinations and ensuring a more reasonable assessment. Subsequently, the evaluation results, including the scores and explanations from both models, are aggregated and fed into Gemini-3-Flash-Preview, which serves as the final arbitrator to deliver the definitive judgment.

\paragraph{Entity-Referenced Hallucination Generation.} We first extract entities such as names, organizations, and numbers from the human-written summary $s_i$. Then, using the extracted entities as references, GPT-4.1, Gemini-3-Flash-Preview, and Claude-sonnet-4-5-20250929-v1 are employed to rewrite $s_i$ into a corresponding hallucinated summary $h_i$ based on eight prompts, each corresponding to a specific hallucination type, as shown in Appendix~\ref{app:Prompt for Hallucination Generation}. When a summary is modified to contain one or two hallucination types, its consistency score is set to 2; when it is injected with three distinct types of hallucinations, its consistency score is set to 1.

\subsection{Human Revision}

To ensure quality, two annotators validate 204 summaries from the test set. 
In our classification framework, a consistency score between $3$ and $5$ is defined as non-hallucinated, whereas a score between $0$ and $2$ is categorized as hallucinated. 
Following the NoCha framework~\cite{karpinska-etal-2024-nocha}, we adopt a pairwise verification setting, allowing annotators to directly contrast a hallucinated summary with its non-hallucinated counterpart against the source text.
Under this setting, the inter-annotator agreement reaches $0.902$ for the binary classification of whether a summary contains hallucinations. 

We further break down Cohen's $\kappa$ across various hallucination types in Appendix~\ref{app:Annotation Rules}, demonstrating high reliability across fine-grained categories.
More details and qualitative examples w.r.t. human revision are shown in Appendix~\ref{app:Annotation Rules}.

\subsection{Dataset Statistics} 
The benchmark consists of 6,354 samples, split into a training set of 5,354 samples, a validation set of 400 samples, and a test set of 600 samples. Table~\ref{Detail statistics of LongNovel.} shows the statistics of LongNovel. The token sequence length of the test set is detailed in Table~\ref{tab:token_length}. More detailed content can be found in Appendix~\ref{app:Dataset}.
\begin{table}[htbp]
\centering
\resizebox{\linewidth}{!}{
\begin{tabular}{@{}l ccc ccc c@{}} 
\toprule
\textbf{Context} & \multicolumn{3}{c}{\textbf{Train}} & \multicolumn{3}{c}{\textbf{Test}} \\
\cmidrule(lr){2-4} \cmidrule(lr){5-7}
\textbf{Length} & Hallu. & Non-H. & Total & Hallu. & Non-H. & Total \\ \midrule
S  & 1,454 & 1,430 & 2,884 & 100 & 100 & 200 \\
M  & 1,292 & 1,178 & 2,470 & 100 & 100 & 200 \\
L  & --    & --    & --    & 50  & 50  & 100 \\
XL & --    & --    & --    & 50  & 50  & 100 \\ \midrule
\textbf{Total} & \multicolumn{3}{c}{\textbf{5,354}} & \multicolumn{3}{c}{\textbf{600}} \\ \bottomrule
\end{tabular}}  \vspace{-0.5em}
\caption{Statistics of the LongNovel training and test sets. Hallu. and Non-H. represent the counts of hallucinated and non-hallucinated samples, respectively.}
\label{Detail statistics of LongNovel.}
\end{table}
\begin{table*}[!t]
\centering
\resizebox{\linewidth}{!}{
    \begin{tabular}{@{} l rrr rrr rrr rrr @{}}
      \toprule
      \multirow{2}{*}{\textbf{Tokenizer}} & \multicolumn{3}{c}{\textbf{S ($n$=200)}} & \multicolumn{3}{c}{\textbf{M ($n$=200)}} & \multicolumn{3}{c}{\textbf{L ($n$=100)}} & \multicolumn{3}{c}{\textbf{XL ($n$=100)}} \\ 
      \cmidrule(lr){2-4} \cmidrule(lr){5-7} \cmidrule(lr){8-10} \cmidrule(lr){11-13}
      & Min & Mean & Max & Min & Mean & Max & Min & Mean & Max & Min & Mean & Max \\ 
      \midrule
      GPT-4o\footnotemark      & 14.65 & 21.63 & 33.35 & 30.15 & 43.81 & 60.93 & 62.34 & 87.32 & 119.87 & 98.78 & 139.68 & 187.36 \\
      InternLM-2.5 & 13.37 & 15.86 & 18.01 & 29.31 & 32.21 & 34.35 & 59.62 & 64.03 &  66.98 & 92.81 & 100.68 & 104.17 \\
      Qwen-3       & 14.40 & 15.81 & 18.06 & 30.23 & 32.09 & 34.48 & 62.22 & 63.76 &  65.93 & 98.30 & 100.21 & 103.32 \\
      GLM-4        & 13.67 & 15.49 & 17.62 & 28.65 & 31.48 & 33.58 & 59.47 & 62.43 &  64.91 & 93.89 &  98.18 & 100.70 \\

      \bottomrule
    \end{tabular}
  }

\caption{Token sequence length of LongNovel test set (values are in thousands, i.e., $k$). }
\label{tab:token_length}
\vspace{-1.5em}
\end{table*}
\footnotetext{\url{https://github.com/openai/tiktoken}}

To evaluate model performance across different error types, we analyze the distribution of hallucination categories within the test set. As illustrated in Fig.~\ref{fig:Distribution of hallucination types across different context lengths.}, the distribution of hallucination types remains relatively balanced across various text lengths. Furthermore, the proportion of Chinese to English data is also evenly distributed across all context lengths, ensuring a consistent benchmark for cross-lingual analysis.

\section{Experiments}
\subsection{Baselines}
We evaluate several state-of-the-art models on the benchmark. The Open-Source Models include InternLM2.5-20B-chat~\cite{cai2024internlm2technicalreport} , GLM-4-9B-chat~\cite{glm2024chatglmfamilylargelanguage}, Llama-3.1-8B-instruct~\cite{grattafiori2024llama3herdmodels}, and the Qwen3 series~\cite{yang2025qwen3technicalreport} (8B, 14B, and 32B). For Commercial Models, we include GPT-5.2-chat, Claude-sonnet-4-5-20250929-v1 and the DeepSeek series (DeepSeek-v3~\cite{deepseekai2025deepseekv3technicalreport}, DeepSeek-r1~\cite{deepseekai2025deepseekr1incentivizingreasoningcapability}, DeepSeek-v4-flash).

\subsection{Experimental Setup}
We employ vLLM~\cite{KwonLZ0ZY0ZS23vllm} for the inference of open-source models across all benchmarks. For the 32k, 64k, and 100k tests, we implement YaRN~\cite{peng2023yarn} scaling to expand the context length from 32k to 128k. For commercial APIs, the temperature is set to 0 to ensure consistent outputs. To guarantee the reliability of our results, each experiment for the open-source models is repeated at least three times. More details are in Appendix~\ref{app:Baseline Evaluation}.

\subsection{Evaluation Metrics} 
In our classification framework, a consistency score between 3 and 5 is defined as non-hallucinated, whereas a score between 0 and 2 is categorized as hallucinated. To extract these scores, we implement a fuzzy regular expression matching mechanism. This ensures that even if a model fails to strictly follow the required output format, the instance can still be correctly evaluated as long as the hallucination detection remains accurate. We employ Balanced Accuracy as our primary evaluation metric. It is calculated as the arithmetic mean of the recall obtained on each class, providing a balanced measure of classification performance. In addition, we employ the Matthews Correlation Coefficient (MCC.) to further assess the model's performance in hallucination detection. It balances the trade-off between false positives and false negatives by considering all categories of the confusion matrix.

\subsection{Compared Methods}

We conduct experiments using different methods on LongNovel.
\paragraph{Zero-shot Prompting.} We provide both the source article and the summary to the LLMs, along with a prompt defining the criteria. The models are instructed to output a factual consistency score alongside a detailed rationale for their judgment. We define two prompt types: target summary at the Beginning (Prompt-B) and target summary at the End (Prompt-E), as shown in Appendix~\ref{app:Prompt for different methods}.

\paragraph{Chain-of-Thought (CoT).}
Based on a zero-shot setting, we implement a CoT~\cite{Wei2022Chain-of-thought} prompting strategy to elicit the model's reasoning capabilities. The model is required to generate a step-by-step analysis of the factual consistency between the source text and the summary before providing the final hallucination detection result. The prompt is shown in Appendix~\ref{app:Prompt for different methods}.

\paragraph{Supervised Fine-Tuning (SFT).}
To enable models to adapt to more positions and activate extrapolation ability, we fine-tune the models with the LongNovel training set by gradually increasing the context length follow findings from~\cite{wei2025cnnsum}. More details are in Appendix~\ref{app:Fine-tuning Experiment}.

\paragraph{Retrieval-Augmented Generation (RAG).}
Standard RAG-based hallucination detection often suffers from contextual fragmentation. Since semantic retrieval typically selects chunks in isolation based on similarity scores, the retrieved evidence often lacks narrative continuity, hindering the model's ability to align summaries with long-form source texts.

Accordingly, we design a RAG framework using a sliding-window mechanism guided by semantic similarity, which dynamically anchors summary segments to their most relevant article context. With this framework, we can obtain the summary chunks and their corresponding article chunks. This process allows us to extract paired summary segments and article chunks, where a summary is classified as non-hallucinated if and only if every individual segment achieves a consistency score greater than 2 within its respective window. We set the block size to 3,500 characters for the source article and 75 characters for the summary. More details are shown in Appendix~\ref{app:RAG framework} and Appendix~\ref{Ablation Study for RAG}.

\paragraph{Voting Ensemble.}
We employ a voting ensemble method integrating three models to determine the final hallucination status of each data instance based on their consistency scores. Specifically, a majority voting rule is applied: if at least two out of the three models identify an instance as a hallucination (indicated by a score less than 3), the instance is classified as hallucinated. Otherwise, if two or more models judge the instance to be hallucination-free (indicated by a score greater than 2), the final decision labels it as non-hallucinated.

\begin{table*}[!t]
  \centering
  \setlength{\tabcolsep}{8pt} 
  \small
  \begin{tabular}{@{} l cc cc cc cc @{}}
    \toprule
    \multirow{2}{*}{\textbf{Model}} & \multicolumn{2}{c}{\textbf{S(16k)}} & \multicolumn{2}{c}{\textbf{M(32k)}} & \multicolumn{2}{c}{\textbf{L(64k)}} & \multicolumn{2}{c}{\textbf{XL(100k)}} \\
    \cmidrule(lr){2-3} \cmidrule(lr){4-5} \cmidrule(lr){6-7} \cmidrule(lr){8-9}
     & BAcc. & MCC. & BAcc. & MCC. & BAcc. & MCC. & BAcc. & MCC. \\ \midrule

    \multicolumn{9}{c}{\textit{Open-Source Models}} \\ \midrule
    InternLM2.5-20B & 0.545 & 0.217 & 0.505 & 0.024 & 0.450 & -0.196 & 0.480 & -0.143 \\
    \quad ---CoT & 0.555 & 0.241 & 0.520 & 0.084 & 0.500 & 0.000 & 0.490 & -0.101 \\
\midrule
    Minicheck-7B & 0.500 & 0.000 & 0.392 & -0.335 & 0.420 & -0.266 & 0.310 & -0.484 \\ \midrule

    GLM-4-9B-chat & 0.505 & 0.014 & 0.510 & 0.059 & 0.500 & 0.000 & 0.500 & 0.000 \\
    \quad ---CoT & 0.520 & 0.048 & 0.445 & -0.139 & 0.470 & -0.094 & 0.510 & 0.059 \\
 \midrule
    
    Llama3.1-8B-instruct & 0.525 & 0.052 & 0.077 & -0.837 & 0.000 & -1.000 & 0.000 & -1.000 \\
    \quad ---CoT & 0.418 & -0.164 & 0.042 & -0.917 & 0.000 & -1.000 & 0.000 & -1.000 \\
    \quad ---SFT & 0.715 & 0.508 & 0.542 & 0.173 & 0.210 & -0.583 & 0.310 & -0.583 \\
 \midrule
 
    Qwen3-8B & 0.568 & 0.233 & 0.608 & 0.237 & 0.533 & 0.075 & 0.480 & -0.050 \\
    \quad ---CoT & 0.537 & 0.131 & 0.568 & 0.140 & 0.577 & 0.192 & 0.450 & -0.115 \\
    \quad ---SFT & \underline{0.773} & \textbf{0.596} & \underline{0.770} & \underline{0.563} & \textbf{0.800} & \textbf{0.643} & \textbf{0.760} & \textbf{0.579} \\ \midrule
    
    Qwen3-14B & 0.588 & 0.285 & 0.598 & 0.219 & 0.547 & 0.105 & 0.517 & 0.036 \\
    \quad ---CoT & 0.595 & 0.324 & 0.573 & 0.171 & 0.570 & 0.167 & 0.533 & 0.073 \\
\midrule
    
    Qwen3-32B & 0.590 & 0.292 & 0.575 & 0.180 & 0.597 & 0.253 & 0.580 & 0.187 \\
    \quad ---CoT & 0.595 & 0.298 & 0.577 & 0.176 & 0.580 & 0.215 & 0.580 & 0.178 \\
    \quad ---SFT & \textbf{0.775} & \underline{0.587} & \textbf{0.795} & \textbf{0.625} & \underline{0.750} & \underline{0.551} & \underline{0.720} & \underline{0.490} \\ 
\midrule

    \multicolumn{9}{c}{\textit{Commercial Models}} \\ \midrule
    Claude-4.5-Sonnet & 0.755 & 0.510 & 0.665 & 0.332 & 0.680 & 0.363 & 0.670 & 0.340 \\
    \quad ---CoT & 0.780 & 0.564 & 0.715 & 0.435 & 0.670 & 0.341 & 0.750 & 0.503 \\
   \midrule
   
    DeepSeek-v3 & 0.755 & 0.544 & 0.735 & 0.521 & 0.680 & 0.450 & 0.670 & 0.433 \\
    \quad ---CoT & 0.735 & 0.510 & 0.685 & 0.424 & 0.620 & 0.312 & 0.610 & 0.280 \\
    \quad ---RAG & -- & -- & -- & -- & 0.510 & 0.068 & 0.500 & -0.014 \\
\midrule
    
    DeepSeek-r1 & 0.775 & 0.592 & 0.735 & 0.510 & 0.640 & 0.382 & 0.680 & 0.450 \\
    \quad ---CoT & 0.710 & 0.440 & 0.680 & 0.435 & 0.610 & 0.327 & 0.670 & 0.433 \\
 \midrule

    DeepSeek-v4-flash & \underline{0.785} & 0.573 & \underline{0.810} & \underline{0.625} & \textbf{0.830} & \underline{0.660} & \textbf{0.830} & \textbf{0.660} \\

 \midrule
 
    GPT-5.2-chat & \underline{0.785} & 0.570 & 0.770 & 0.541 & \underline{0.810} & 0.620 & 0.740 & 0.482 \\
    \quad ---CoT & 0.775 & 0.555 & 0.765 & 0.530 & 0.780 & 0.560 & 0.750 & 0.503 \\
    \quad ---RAG & -- & -- & -- & -- & \textbf{0.830} & \textbf{0.661} & 0.750 & 0.500\\
\midrule
    Voting Ensemble (DSV3/DSR/C) & \underline{0.785} & \underline{0.600} & 0.745 & 0.532 & 0.680 & 0.450 & 0.680 & 0.450 \\
    Voting Ensemble (DSV4/GPT/C) & \textbf{0.820} & \textbf{0.641} & \textbf{0.815} & \textbf{0.630} & \textbf{0.830} & \underline{0.660} & \underline{0.820} & \underline{0.641} \\
    \bottomrule
  \end{tabular}
  \caption{Balanced Accuracy and MCC of various models. The best score is \textbf{bold} and the second-best is \underline{underline} within each category (Open-Source vs Commercial). Voting Ensembles are abbreviation-coded as follows: (DSV3/DSR/C) denotes DeepSeek-v3, DeepSeek-r1, and Claude-sonnet-4-5-20250929-v1;  (DSV4/GPT/C) denotes DeepSeek-v4-flash, GPT-5.2-chat, and Claude-sonnet-4-5-20250929-v1.}
  \label{Accuracy and MCC.-score of LongNovel.}
     \vspace{-2.5ex}
\end{table*}

\begin{table*}[!t]
\centering
\setlength{\tabcolsep}{5pt}
\small
\begin{tabular}{@{} l cc cc @{}}
\toprule
\multirow{2}{*}{\textbf{Model}} 
& \multicolumn{2}{c}{\textbf{L (64k)}} 
& \multicolumn{2}{c}{\textbf{XL (100k)}} \\
\cmidrule(lr){2-3} \cmidrule(lr){4-5}
& \textbf{BAcc. (95\% CI)} & \textbf{MCC. (95\% CI)}
& \textbf{BAcc. (95\% CI)} & \textbf{MCC. (95\% CI)} \\
\midrule

Claude-4.5-Sonnet 
& 0.680 [0.590, 0.770] 
& 0.363 [0.181, 0.543]
& 0.670 [0.580, 0.760]
& 0.340 [0.160, 0.524] \\

DeepSeek-v3
& 0.680 [0.610, 0.750]
& 0.450 [0.312, 0.577]
& 0.670 [0.600, 0.740]
& 0.433 [0.293, 0.562] \\

DeepSeek-r1
& 0.640 [0.580, 0.710]
& 0.382 [0.245, 0.515]
& 0.680 [0.610, 0.750]
& 0.450 [0.312, 0.577] \\

DeepSeek-v4-flash
& \textbf{0.830 [0.760, 0.900]}
& \textbf{0.660 [0.520, 0.801]}
& \textbf{0.830 [0.750, 0.900]}
& \textbf{0.660 [0.505, 0.801]} \\

GPT-5.2-chat
& 0.810 [0.730, 0.880]
& 0.620 [0.465, 0.762]
& 0.740 [0.650, 0.820]
& 0.482 [0.308, 0.648] \\

\midrule
Voting Ensemble (DSV3/DSR/C)
& 0.680 [0.610, 0.750]
& 0.450 [0.314, 0.577]
& 0.680 [0.610, 0.750]
& 0.450 [0.308, 0.577] \\

Voting Ensemble (DSV4/GPT/C)
& \textbf{0.830 [0.760, 0.900]}
& \textbf{0.660 [0.520, 0.801]}
& 0.820 [0.740, 0.890]
& 0.641 [0.483, 0.781] \\

\bottomrule
\end{tabular}
\caption{Performance of commercial models and voting ensembles on the L (64k) and XL (100k) scales, with 95\% confidence intervals obtained using 10000 iterations of stratified percentile bootstrap resampling. Voting Ensembles are abbreviation-coded as follows: (DSV3/DSR/C) denotes DeepSeek-v3, DeepSeek-r1, and Claude-sonnet-4-5-20250929-v1; (DSV4/GPT/C) denotes DeepSeek-v4-flash, GPT-5.2-chat, and Claude-sonnet-4-5-20250929-v1.}
\label{tab:commercial_bootstrap_ci}
\end{table*} 
\begin{table*}[t]
  \centering
  \setlength{\tabcolsep}{10pt}
\small
  \begin{tabular}{@{} l cc cc cc cc @{}}
    \toprule
    \multirow{2}{*}{\textbf{Model}} & \multicolumn{2}{c}{\textbf{S(16k)}} & \multicolumn{2}{c}{\textbf{M(32k)}} & \multicolumn{2}{c}{\textbf{L(64k)}} & \multicolumn{2}{c}{\textbf{XL(100k)}} \\
    \cmidrule(lr){2-3} \cmidrule(lr){4-5} \cmidrule(lr){6-7} \cmidrule(lr){8-9}
     & BAcc. & MCC. & BAcc. & MCC. & BAcc. & MCC. & BAcc. & MCC. \\ \midrule
    Qwen3-8B & 0.568 & 0.233 & 0.608 & 0.237 & 0.533 & 0.075 & 0.480 & -0.050 \\
    Qwen3-8B-MMA & 0.577 & 0.248 & 0.577 & 0.171 & 0.610 & 0.261 & 0.710 & 0.447 \\
    Qwen3-8B-FULL & 0.773 & 0.596 & 0.770 & 0.563 & 0.800 & 0.643 & 0.760 & 0.579 \\
\midrule
    Llama3.1-8B-instruct & 0.525 & 0.052 & 0.077 & -0.837 & 0.000 & -1.000 & 0.000 & -1.000 \\
    Llama3.1-8B-MMA & 0.585 & 0.290 & 0.502 & 0.008 & 0.020 & -0.961 & 0.000 & -1.000 \\
    Llama3.1-8B-FULL & 0.715 & 0.508 & 0.542 & 0.173 & 0.210 & -0.583 & 0.310 & -0.583 \\
\midrule
    Qwen3-32B & 0.590 & 0.292 & 0.575 & 0.180 & 0.597 & 0.253 & 0.580 & 0.187 \\
    Qwen3-32B-MMA & 0.605 & 0.343 & 0.587 & 0.281 & 0.610 & 0.352 & 0.680 & 0.469 \\
    Qwen3-32B-FULL & 0.775 & 0.587 & 0.795 & 0.625 & 0.750 & 0.551 & 0.720 & 0.490 \\
    \bottomrule
  \end{tabular}
\caption{Ablation study on data construction strategies across various models and dataset scales.}
  \label{tab: absolution dataset}
\end{table*}

\section{Experimental Results and Analysis}

\subsection{Main Results} 
We calculate balanced accuracy and MCC across various context lengths for both open-source and commercial models, with detailed performance statistics presented in Table~\ref{Accuracy and MCC.-score of LongNovel.}. Full results are provided in Appendix~\ref{app:Full results}. Negative MCC values (e.g., Llama-3.1-8B-instruct) stem from instances where models fail to output a valid consistency score, which we treat as incorrect predictions. To ensure this setup does not underestimate true model capabilities due to strict parser limitations, we manually inspect all cases where Llama-3.1-8B fails to emit a valid score. Our analysis reveals that these failures are not parsing errors; rather, in every instance, the model either degenerates into repetitive outputs or completely fails to follow instructions by generating novel summaries instead of performing hallucination detection. Thus, these format failures accurately reflect a collapse in long-context instruction-following, justifying their classification as incorrect predictions. From the results, we can draw the following conclusions:

Performance exhibits a downward trend as context length increases across the commercial and open-source segments. The open-source models demonstrate a struggle with long contexts, as their baseline scores remain generally low overall. For example, Qwen3-14B drops its baseline accuracy from 0.588 at 16k down to 0.517 at 100k. For the commercial models, the degradation is equally manifest; Claude-sonnet-4-5-20250929-v1 slips from an accuracy of 0.755 at 16k to 0.670 at 100k. Furthermore, at the 64k and 100k stages, some negative Matthews Correlation Coefficient scores, such as Llama3.1-8B-instruct hitting -1.000, show the severe decline in instruction-following capability, suggesting that as the context lengthens, these models frequently produce repetitive outputs or mistakenly shift toward generating a novel summary instead of hallucination detection, as detailed in  Appendix~\ref{app:Repetition Output} and Appendix~\ref{app:Failure in JSON Format Generation}.

The impact of Chain-of-Thought (CoT) prompting varies significantly across different architectures, showing clear benefits for some models while proving counterproductive for others. For example, while CoT successfully lifts the 32k accuracy of Claude-4.5-Sonnet from 0.665 to 0.715, it has the opposite effect on DeepSeek-v3, dragging its 32k accuracy down from 0.735 to 0.685 and further degrading its 100k performance from 0.670 to 0.610.

In the 16k to 100k range, the SFT variants of the Qwen3-32B model achieve average accuracy significantly higher than both the base and CoT versions, reaching 0.720 at 100k compared to only 0.580 for the base version. This demonstrates that specialized fine-tuning of open-source models for length extrapolation is an effective strategy for hallucination detection in long-context tasks.

The RAG strategy exhibits limitations in the 64k context for DeepSeek-v3, where its performance falls sharply to an accuracy of 0.530. However, RAG scales slightly better at 100k for GPT-5.2-chat, achieving an accuracy of 0.750. This performance discrepancy likely stems from the model's baseline judgment accuracy within the 32k to 64k window; since DeepSeek-v3 has a lower accuracy than GPT-5.2-chat, dividing summaries into smaller chunks increases the number of required decisions, leading to a rapid accumulation of errors from individual judgments. Conversely, GPT-5.2-chat commits fewer baseline errors, allowing it to maintain a performance lift when processing fragmented chunks.

To assess evaluation uncertainty, we perform paired bootstrap significance testing with 95\% confidence intervals derived from stratified percentile resampling (10,000 iterations; Table~\ref{tab:commercial_bootstrap_ci}). Although the Voting Ensemble (DSV4/GPT/C) reaches top-tier nominal scores at 64k, paired tests show no statistically significant gain over DeepSeek-v4-Flash, the top single model ($\Delta\text{BAcc} = 0.000, p = 1.000$ at 64k; $\Delta\text{BAcc} = -0.010, p = 0.827$ at 100k). This suggests that collaborative decision-making offers no distinct advantage over strong individual models at extreme context lengths.

\subsection{Analysis and Case Study} 

\paragraph{Ablation Study on Dataset Construction}
Our dataset is primarily constructed via two distinct strategies: Multi-Model Arbitration (MMA) and Entity-Referenced Hallucination Generation (ERHG). To thoroughly investigate the efficacy of the negative samples generated by the ERHG strategy, we evaluate three configurations: the base models, the variants fine-tuned only on the MMA-processed data, and the models trained on the complete dataset integrating both strategies. As illustrated in Table~\ref{tab: absolution dataset}, while the MMA strategy generally yields incremental improvements over the base models, $\text{Model}_{\text{Full}}$ consistently achieves the highest Accuracy and MCC across all settings. This substantial and consistent performance gap across different dataset sizes firmly demonstrates that the ERHG strategy is highly effective in synthesizing high-quality, challenging hallucination data, ultimately empowering the models with significantly stronger robust alignment capabilities.

\paragraph{Why Hallucination Detection Fails in Long-context Scale?} 
By analyzing outputs of models at the long-context scale, which are shown in Appendix~\ref{app:failure_cases}, we find several error patterns. 

First, models may fail to comprehensively process or may misread both the source text and its corresponding summary, directly resulting in incorrect detection results. Second, in some cases, the model identifies a hallucination yet produces a correction that is identical to the original sentence. However, most of these cases are faithful and do not need corrections. Third, a deficiency in reasoning capabilities prevents models from successfully mapping a series of specific actions or dialogues to a correct abstract generalization. Furthermore, repetitive output patterns sometimes cause the model response to exceed length constraints, leading to truncated and incomplete answers. Additionally, some cases fail to complete the hallucination detection, indicating a drop in instruction-following performance. We also observe internal logical contradictions where a model initially identifies a hallucination but ultimately concludes that no such hallucination exists when giving the reason for the judgment. Finally, models often fail to recognize semantic equivalence, misidentify a summary as a hallucination due to lexical changes or omitted peripheral information, despite the core summary remaining factually accurate.

\section{Conclusion}
We introduce LongNovel, a bilingual long-context dataset for hallucination detection in novels, based on human-annotated summaries. It comprises four subsets ranging from 16k to 100k tokens. Our extensive experiments on LongNovel reveal that current large language models still lack sufficient capability in long-context hallucination detection tasks. We hope that LongNovel will provide useful insights for future research in this field.

\section*{Limitations}
Our study is limited to open-source models with up to 32B parameters. Consequently, the generalization capabilities of large-scale models, such as Llama-3.1-70B, have not yet been investigated. Finally, as LongNovel is a novel dataset, the applicability of our findings to non-novel domains remains to be further validated in future research.

\section*{Acknowledgments}
We would like to express our gratitude to the annotators from iQIYI for their high-quality manual labeling and correction. We also thank iQIYI for providing the GPU resources that supported this work.
This work was supported in part by National Natural Science Foundation of China under Grant (No. 62572198) and Key Laboratory of Advanced Theory and Application in Statistics and Data Science, Ministry of Education.

\bibliography{custom}

\appendix

\section{LongNovel Dataset}

\label{app:Dataset}
The LongNovel benchmark is constructed from publicly available literary works. We have manually reviewed the dataset to ensure it contains no sensitive personally identifying information (PII) of living individuals. 

The dataset is compiled from publicly accessible web sources for non-commercial, academic research purposes. The data is used strictly for training and evaluation, and no sensitive personal information is involved.

We partition the Chinese subset of the dataset into training, validation, and test sets. The detailed composition and statistical distribution of the dataset are presented in Table~\ref{tab:Statistical distribution and composition of the LongNovel dataset across various subsets}. The specific book titles and their respective authors within this Chinese corpus are summarized in Table~\ref{tab:longnovel_summary}. For the English subset, the training, validation, and test sets are partitioned in strict accordance with the splits of the BookSum benchmark \cite{kryscinski-etal-2022-booksum}.

\begin{table*}[!hb]
  \centering

  \setlength{\tabcolsep}{7pt}

  \begin{tabular}{lcccccccc}
    \toprule
    \textbf{Subset} & \textbf{Total} & \textbf{ZH} & \textbf{EN} & \textbf{Original} & \textbf{ERHG} & \textbf{MMA} & \textbf{Hallu.} & \textbf{Non-H.} \\
    \midrule
    Train\_16k  & 2884 & 1230 & 1654 & 814 & 963 & 1107 & 1454 & 1430 \\
    Train\_32k  & 2470 & 1091 & 1379 & 734 & 937 & 799  & 1292 & 1178 \\
    Valid\_16k  & 200  & 100  & 100  & 60  & 40  & 100  & 100  & 100  \\
    Valid\_32k  & 200  & 100  & 100  & 60  & 40  & 100  & 100  & 100  \\
    Test\_16k   & 200  & 100  & 100  & 64  & 48  & 88   & 100  & 100  \\
    Test\_32k   & 200  & 100  & 100  & 69  & 61  & 70   & 100  & 100  \\
    Test\_64k   & 100  & 50   & 50   & 41  & 27  & 32   & 50   & 50   \\
    Test\_100k  & 100  & 54   & 46   & 38  & 20  & 42   & 50   & 50   \\
    \bottomrule
  \end{tabular}
  \caption{Statistical distribution and composition of the LongNovel dataset across various subsets. ZH and EN denote Chinese and English data; Original denotes concatenated and compressed human-written summaries; ERHG denotes Entity-Referenced Hallucination Generation; MMA denotes Multi-Model Arbitration; Hallu. and Non-H. represent the counts of hallucinated and non-hallucinated samples}
  \label{tab:Statistical distribution and composition of the LongNovel dataset across various subsets}
\end{table*}

\begin{table*}[h!]
\centering

\begin{tabular}{lll}
\toprule
\textbf{Split} & \textbf{Book Title} & \textbf{Author} \\
\midrule
\multirow{15}{*}{Training Set} 
 & Bright Eyes in the Dark & Er Dong Tu Zi \\
 & Destined & Mo Shu Bai \\
 & Misty Rain Tower & Yixi Yanyu \\
 & The Gentlemen of the City & Jin Shisichai \\
 & We Love Each Other So Much (Chinese edition)  & Marcela Serrano \\
 & Little Confucian Immortal Seeking the Unknown & Youziyin \\
 & Turing's Code & Feitian Yexiang \\
 & Folding Beijing & Hao Jingfang \\
 & I'm Waiting for You in the Memory & Xin Yiwu \\
 & Bu Yi Jian & Feng Diuzi \\
 & The Chronicle of Qingxi: Volume 1 & Hong Zhuxia \\
 & The Chronicle of Qingxi: Volume 2 & Hong Zhuxia \\
 & The Chronicle of Qingxi: Volume 3 & Hong Zhuxia \\
 & The Space-Time Painter & Hai Ya \\
 & Once Gone & Qingshan Huangzhong \\
\midrule
\multirow{4}{*}{Validation Set} 
 & Yuhong & Ban Yu \\
 & Let Me Look at You & Xin Yiwu \\
 & I'm Waiting for You in the Memory & Xin Yiwu \\
 & My Garden & Teng Ping \\
\midrule
\multirow{10}{*}{Test Set} 
 & Bad Kids & Zijin Chen \\
 & Exclusive Possession & Ding Mo \\
 & Blood is Burning & Bainian Ruge \\
 & Everyone is a Protagonist Except Me & Cong Wen \\
 & The System Granted Me Longevity & Zi Ling Feng Xue \\
 & Mysterious Lotus Casebook & Teng Ping \\
 & Cicadas Sing the Setting Sun West & Yu Luo Zhu Leng \\
 & Ge Lu Ming: Volume 1 & Qin Huai \\
 & Ge Lu Ming: Volume 2 & Qin Huai \\
 & The Creatures That We Are & Peng Pai \\
\bottomrule
\end{tabular}
\caption{Detailed composition of the Chinese subset in the LongNovel dataset, listing the book titles and authors across the training, validation, and test splits.}
  \label{tab:longnovel_summary}
\end{table*}

\section{Hallucination Type}
\label{app:Hallucination_Type}
To construct a comprehensive hallucination detection dataset, we categorize hallucinations observed in novel summarization into eight distinct types by referring to related research~\cite{KryscinskiMXS2020factcc,pagnoni-etal-2021-understanding, orlovskiy2024uncertaintyresolutionmisinformationdetection, mishra2024finegrainedhallucinationdetectionediting, diahalu}.

\paragraph{Entity Hallucination} Entity Hallucination occurs when a summary contains factual errors regarding the entities mentioned in the source text. This encompasses pronominal errors, subject-object role reversals, and erroneous descriptions of entities such as characters, organizations, or locations.

\paragraph{Numerical Hallucination} Numerical Hallucination occurs when a summary introduces numerical data that is inconsistent with the source text. This encompasses differences in quantities, ages, or monetary values.

\paragraph{Relation Hallucination} Relation Hallucination occurs when a summary describes interpersonal relationships that are inconsistent with the original text. This encompasses misreporting established relationships, such as familial or professional ties, as well as fabricating non-existent relations that the source material does not support.

\paragraph{Logical Inversion} Logical Inversion occurs when a summary conveys a meaning that is logically opposite to the source text. This encompasses the conversion of affirmative statements into negative ones and the conversion of negative statements into affirmative ones.

\paragraph{Event Hallucination} Event Hallucination occurs when a summary describes an event that contradicts the source text. This encompasses the replacement of core verbs, the alteration of event outcomes, or the modification of action intensity and manner.

\paragraph{Temporal Hallucination} Temporal Hallucination occurs when the summary disrupts the chronological order of the narrative by inverting or scrambling the sequence of two or more events as they occurred in the source text.

\paragraph{Causal Hallucination} Causal Hallucination occurs when a summary introduces causal relationships that are inconsistent with or unsupported by the source text. This encompasses false attributions where unrelated events are logically linked, causal reversals where cause and effect are swapped, and reason substitutions where factual outcomes are attributed to irrelevant origins.

\paragraph{Event Fabrication} Event Fabrication occurs when a summary introduces actions or states that are unsupported by the source text. This encompasses characters' actions not found in the original text, the repetition of existing events, and the fabrication of characters' thoughts.

\section{Human annotation}
\label{app:Annotation Rules}
\subsection{Summary Annotation}
We employ 16 annotators, all of whom are specializing in Chinese Language and Literature. Following the establishment of annotation rules, annotators conduct trial annotations. After reviewing the trial results and providing specific feedback, we proceed to large-scale annotation, where for each textual unit, one annotator drafts an initial summary that is subsequently revised by two additional annotators to ensure accuracy and faithfulness to the source text. We ensure that all annotators receive fair compensation and confirm that the hourly rate is higher than the local legal minimum wage. The payment calculation accounts for all active working hours, including both the trial and main annotation phases.

\subsection{Hallucination Annotation}
The revision and annotation process was conducted by two NLP researchers (one female, one male). Following the NoCha framework \cite{karpinska-etal-2024-nocha}, our methodology utilizes a minimal pair approach, where each set consists of a hallucinated summary and its non-hallucinated counterpart. Under this framework, a score is awarded only if both summaries within a single pair are correctly identified. This criterion not only ensures high data quality but also allows annotators to easily cross-verify the claims, effectively filtering out ambiguous or overly subjective cases. Furthermore, no strict time constraints are imposed on the annotators, allowing them sufficient time to thoroughly analyze each pair and further guaranteeing the reliability of the annotations. After reading the guidelines shown in Fig.~\ref{fig:Annotation Rules}, the annotators are required to identify the presence of hallucinations, the specific hallucinated sentences, and provide detailed explanations. For cases where disagreements occur, the two annotators resolve the discrepancies through discussion and revise the data accordingly to reach a final consensus.

We also analyze the inter-annotator agreement (Cohen's $\kappa$) broken down by fine-grained hallucination types. 
As shown in Table~\ref{tab:iaa_breakdown}, the agreement remains consistently high across various complex categories (ranging from 0.820 for Temporal Hallucination to 1.000 for Numerical, Relation, Event Fabrication, and Causal Hallucinations). 
Note that a single hallucinated summary may contain multiple hallucination types; thus, category counts are not mutually exclusive. 

\begin{table*}[htbp]
\centering
\caption{Breakdown of Cohen's $\kappa$ by hallucination type during human validation. Note that category counts are not mutually exclusive.}
\label{tab:iaa_breakdown}
\begin{tabular}{lcc}
\toprule
\textbf{Hallucination Type} & \textbf{Count} & \textbf{Cohen's $\kappa$} \\
\midrule
Entity Hallucination       & 44 & 0.954 \\
Numerical Hallucination    & 22 & 1.000 \\
Relation Hallucination     & 14 & 1.000 \\
Event Hallucination        & 48 & 0.914 \\
Event Fabrication          & 37 & 1.000 \\
Causal Hallucination       & 23 & 1.000 \\
Logical Inversion          & 16 & 0.875 \\
Temporal Hallucination     & 34 & 0.820 \\
\bottomrule
\end{tabular}
\end{table*}

\section{RAG Framework}
\label{app:RAG framework}
We design a Retrieval-Augmented Generation (RAG) framework tailored for long-text alignment. Specifically, we partition the source document $\mathcal{D}$ into $N$ smaller blocks, denoted as $\mathcal{D} = \{D_0, D_1, \dots, D_{N-1}\}$, and the summary $\mathcal{S}$ into $M$ blocks, denoted as $\mathcal{S} = \{S_0, S_1, \dots, S_{M-1}\}$. We employ the BGE-M3 model to project each summary block $S_j$ and document block $D_i$ into their respective embedding vectors $\mathbf{s}_j$ and $\mathbf{d}_i$. The cosine similarity between these embedding vectors is calculated as:
\begin{equation}
\text{Sim}(\mathbf{s}_j, \mathbf{d}_i) = \frac{\mathbf{s}_j \cdot \mathbf{d}_i}{\|\mathbf{s}_j\| \|\mathbf{d}_i\|}
\end{equation}

Based on the similarity matrix, the process follows these steps:
\begin{enumerate} 
    \item \textbf{Initialization:} 
    For the first summary block $S_0$, the initial document anchor index is routinely set to $k_0 = 0$. 
    
    \item \textbf{Dynamic Matching:} 
    For subsequent summary blocks $S_j$ ($j = 1, \dots, M-1$), the selection of the current anchor $k_j$ is rigidly constrained to maintain a chronological narrative order (i.e., $k_j \ge k_{j-1}$). We filter candidates from the top-$5$ matches of $S_j$ that satisfy this condition. Preference is hierarchically given to $k_{j-1}+1$, followed by $k_{j-1}$, and then the minimum index among the remaining valid right-side candidates. If no right-side candidates exist, $k_j$ defaults to $k_{j-1}$.
    
    \item \textbf{Adaptive Window Construction:} 
    The bidirectional expansion radius of the dynamic context window is defined as $R_1 = \lfloor N/3 \rfloor - 1$. The window is constructed around $k_j$ and aligned with the preceding boundary to ensure the right margin never regresses. Crucially, the final window for $S_{M-1}$ is forcefully extended to cover the absolute last document block $D_{N-1}$.
    
    \item \textbf{Two-Tier Review Mechanism:} 
    To prevent false positives arising from missing context due to localized retrieval windows, a two-tier verification mechanism is introduced. Each summary piece is first evaluated against its dynamically constructed local context by a Large Language Model for consistency scoring (0-5 scale). If the local score drops to $\le 2$ (indicating significant hallucination), the system immediately invokes a full-context review by swapping the localized context with the entire document $\mathcal{D}$. If the full-context score climbs above 2, the evaluation adopts the revised score and resumes. Otherwise, if the full-article score remains $\le 2$, a hallucination is confirmed, and the evaluation for the remaining summary blocks of the sample is terminated early.
\end{enumerate}
Our sliding-window RAG framework is implemented with specific configurations to ensure precise alignment between the summary and the source article. Specifically, we set the block size to 3,500 characters for the source article and 75 characters for the summary segments. A summary is ultimately classified as non-hallucinated if and only if every individual segment within it achieves a consistency score greater than 2 within its respective retrieved window.

\section{Experimental Setup Details}
\subsection{Baseline Evaluation}
\label{app:Baseline Evaluation}
We employ extrapolation strategies by configuring vLLM initialization parameters. For the Qwen series and InternLM series, we configure the YaRN interpolation method with a scaling factor of 4.0 to enhance their long-context capabilities. For Llama-3.1-8B-Instruct, we set the scaling factor to 8.0, consistent with its original 8192 position embeddings. For the GLM-4-9B model, which supports a 128K context window, we maintained its default configurations.

\subsection{Fine-tuning Experiment}
\label{app:Fine-tuning Experiment}
We implement a full parameter fine-tuning integrated with Sequence Parallelism by implementing Ring-Attention~\cite{Liu23RingAttention}, using the 360-llama-factory framework~\cite{zou360-LLaMA-Factory}. To optimize computational efficiency and memory usage, we employ Flash Attention 2~\cite{Dao24FlashAttention-2} and DeepSpeed ZeRO-3 Offload~\cite{RARYZ0H21ZeRO-Offload} strategy. Regarding hyperparameter configurations, a learning rate of 5e-6 is applied using a cosine scheduler with zero warmup. All experiments are conducted using a fixed seed of 42. 
For fine-tuning on 16K sequences, we initially performed a 100-step fine-tuning on 2K data before proceeding to the full 16K fine-tuning. For evaluations at 32K, 64K, and 100K scales, we conduct an initial 100-step fine-tuning on 16k data before proceeding to the 32K scale. We set 2 epochs of training throughout each stage of the process, and select the best-performing checkpoint on the validation set as our final model. All experiments are conducted on 8 NVIDIA H20 (96GB) GPUs. The fine-tuning prompt is the same as the inference prompt, which is detailed in the Appendix~\ref{app:Prompt for different methods}.

\begin{table*}[!thb]
  \centering
 \setlength{\tabcolsep}{8pt} 
 \small
  \begin{tabular}{@{} l cccc cccc @{}}
    \toprule
    \multirow{2}{*}{\textbf{Model}} & \multicolumn{4}{c}{\textbf{S(16k)}} & \multicolumn{4}{c}{\textbf{M(32k)}} \\
    \cmidrule(lr){2-5} \cmidrule(lr){6-9}
     & Non-H & Hallu & BAcc. & MCC. & Non-H & Hallu & BAcc. & MCC. \\ \midrule

    \multicolumn{9}{c}{\textit{Open-Source Models}} \\ \midrule
    InternLM2.5-20B & 1.000 & 0.090 & 0.545 & 0.217 & 0.960 & 0.050 & 0.505 & 0.024 \\
     \quad ---CoT & 1.000 & 0.110 & 0.555 & 0.241 & 0.960 & 0.080 & 0.520 & 0.084 \\
     \quad ---Prompt-B & 0.790 & 0.380 & 0.585 & 0.186 & 0.950 & 0.030 & 0.490 & -0.051 \\ \midrule
    Minicheck-7B & 0.980 & 0.020 & 0.500 & 0.000 & 0.773 & 0.010 & 0.392 & -0.335 \\ \midrule

    GLM-4-9B-chat & 0.860 & 0.150 & 0.505 & 0.014 & 0.900 & 0.120 & 0.510 & 0.059 \\
     \quad ---CoT & 0.800 & 0.240 & 0.520 & 0.048 & 0.750 & 0.140 & 0.445 & -0.139 \\
     \quad ---Prompt-B & 0.510 & 0.310 & 0.410 & -0.184 & 0.500 & 0.430 & 0.465 & -0.070 \\ \midrule
    
    Llama3.1-8B-instruct & 0.410 & 0.640 & 0.525 & 0.052 & 0.103 & 0.050 & 0.077 & -0.837 \\
     \quad ---CoT & 0.447 & 0.390 & 0.418 & -0.164 & 0.060 & 0.023 & 0.042 & -0.917 \\
     \quad ---Prompt-B & 0.840 & 0.180 & 0.510 & 0.027 & 0.060 & 0.040 & 0.050 & -0.900 \\
     \quad ---SFT & 0.970 & 0.460 & 0.715 & 0.508 & 0.980 & 0.103 & 0.542 & 0.173 \\ \midrule
 
    Qwen3-8B & 0.973 & 0.163 & 0.568 & 0.233 & 0.803 & 0.413 & 0.608 & 0.237 \\
     \quad ---CoT & 0.950 & 0.123 & 0.537 & 0.131 & 0.677 & 0.460 & 0.568 & 0.140 \\
     \quad ---Prompt-B & 0.970 & 0.120 & 0.545 & 0.171 & 0.670 & 0.420 & 0.545 & 0.093 \\
     \quad ---SFT & 0.973 & 0.573 & \underline{0.773} & \textbf{0.596} & 0.910 & 0.630 & \underline{0.770} & \underline{0.563} \\ \midrule
    
    Qwen3-14B & 0.980 & 0.197 & 0.588 & 0.285 & 0.817 & 0.380 & 0.598 & 0.219 \\
     \quad ---CoT & 1.000 & 0.190 & 0.595 & 0.324 & 0.830 & 0.317 & 0.573 & 0.171 \\
     \quad ---Prompt-B & 0.950 & 0.170 & 0.560 & 0.192 & 0.680 & 0.390 & 0.535 & 0.073 \\ \midrule
    
    Qwen3-32B & 0.983 & 0.197 & 0.590 & 0.292 & 0.850 & 0.300 & 0.575 & 0.180 \\
     \quad ---CoT & 0.980 & 0.210 & 0.595 & 0.298 & 0.823 & 0.330 & 0.577 & 0.176 \\
     \quad ---Prompt-B & 0.850 & 0.290 & 0.570 & 0.169 & 0.870 & 0.330 & 0.600 & 0.238 \\
     \quad ---SFT & 0.950 & 0.600 & \textbf{0.775} & \underline{0.587} & 0.960 & 0.630  & \textbf{0.795} & \textbf{0.625} \\ \midrule

    \multicolumn{9}{c}{\textit{Commercial Models}} \\ \midrule
    Claude-4.5-Sonnet & 0.760 & 0.750 & 0.755 & 0.510 & 0.610 & 0.720 & 0.665 & 0.332 \\
     \quad ---CoT & 0.720 & 0.840 & 0.780 & 0.564 & 0.640 & 0.790 & 0.715 & 0.435 \\ 
     \quad ---Prompt-B & 0.430 & 0.750 & 0.590 & 0.190 & 0.440 & 0.660 & 0.590 & 0.190 \\ \midrule
   
    DeepSeek-v3 & 0.930 & 0.580 & 0.755 & 0.544 & 0.950 & 0.520 & 0.735 & 0.521 \\
     \quad ---CoT & 0.930 & 0.540 & 0.735 & 0.510 & 0.930 & 0.440 & 0.685 & 0.424 \\ 
     \quad ---Prompt-B & 0.910 & 0.480 & 0.695 & 0.432 & 0.940 & 0.230 & 0.585 & 0.241 \\ \midrule
    
    DeepSeek-r1 & 0.960 & 0.590 & 0.775 & 0.592 & 0.930 & 0.540 & 0.735 & 0.510 \\
     \quad ---CoT & 0.860 & 0.560 & 0.710 & 0.440 & 0.960 & 0.400 & 0.680 & 0.435 \\ 
     \quad ---Prompt-B & 0.920 & 0.500 & 0.710 & 0.463 & 0.930 & 0.250 & 0.590 & 0.245 \\ \midrule

    DeepSeek-v4-flash & 0.840 & 0.730 & \underline{0.785} & 0.573 & 0.750 & 0.870 & \underline{0.810} & \underline{0.625} \\ \midrule
 
    GPT-5.2-chat & 0.800 & 0.770 & \underline{0.785} & 0.570 & 0.800 & 0.740 & 0.770 & 0.541 \\
     \quad ---CoT & 0.840 & 0.710 & 0.775 & 0.555 & 0.770 & 0.760 & 0.765 & 0.530 \\ 
     \quad ---Prompt-B & 0.910 & 0.660 & 0.785 & 0.589 & 0.800 & 0.700 & 0.750 & 0.503 \\ \midrule

    Voting Ensemble (DSV3/DSR/C) & 0.940 & 0.630 & \underline{0.785} & \underline{0.600} & 0.940 & 0.550 & 0.745 & 0.532 \\
    Voting Ensemble (DSV4/GPT/C) & 0.840 & 0.800 & \textbf{0.820} & \textbf{0.641} & 0.800 & 0.830 & \textbf{0.815} & \textbf{0.630} \\
    \bottomrule
  \end{tabular}
  \caption{Detailed performance statistics in 16k, 32k scales. \texttt{Prompt-B} indicates that the target summary is positioned at the Beginning, while the default setup uses \texttt{Prompt-E}, which positions the target summary at the End. Voting Ensembles are abbreviation-coded as follows: (DSV3/DSR/C) denotes DeepSeek-v3, DeepSeek-r1, and Claude-sonnet-4-5-20250929-v1;  (DSV4/GPT/C) denotes DeepSeek-v4-flash, GPT-5.2-chat, and Claude-sonnet-4-5-20250929-v1.}
  \label{tab:Detailed performance statistics in 16k, 32k scales.}
\end{table*}
\begin{table*}[!thb]
  \centering
 \setlength{\tabcolsep}{8pt} 
 \small
  \begin{tabular}{@{} l cccc cccc @{}}
    \toprule
    \multirow{2}{*}{\textbf{Model}} & \multicolumn{4}{c}{\textbf{L (64k)}} & \multicolumn{4}{c}{\textbf{XL (100k)}} \\
    \cmidrule(lr){2-5} \cmidrule(lr){6-9}
     & Non-hallu & Hallu & BAcc. & MCC. & Non-hallu & Hallu & BAcc. & MCC. \\ \midrule

    \multicolumn{9}{c}{\textit{Open-Source Models}} \\ \midrule
    InternLM2.5-20B & 0.880 & 0.020 & 0.450 & -0.196 & 0.960 & 0.000 & 0.480 & -0.143 \\
    \quad ---CoT & 0.960 & 0.040 & 0.500 & 0.000 & 0.980 & 0.000 & 0.490 & -0.101 \\
    \quad ---Prompt-B & 0.920 & 0.040 & 0.480 & -0.084 & 0.780 & 0.080 & 0.430 & -0.196 \\ \midrule
    
    Minicheck-7B & 0.820 & 0.020 & 0.420 & -0.266 & 0.620 & 0.000 & 0.310 & -0.484 \\ \midrule

    GLM-4-9B-chat & 1.000 & 0.000 & 0.500 & 0.000 & 1.000 & 0.000 & 0.500 & 0.000 \\
    \quad ---CoT & 0.853 & 0.087 & 0.470 & -0.094 & 0.980 & 0.040 & 0.510 & 0.059 \\
    \quad ---Prompt-B & 0.500 & 0.300 & 0.400 & -0.204 & 0.660 & 0.260 & 0.460 & -0.087 \\ \midrule
    
    Llama3.1-8B-instruct & 0.000 & 0.000 & 0.000 & -1.000 & 0.000 & 0.000 & 0.000 & -1.000 \\
    \quad ---CoT & 0.000 & 0.000 & 0.000 & -1.000 & 0.000 & 0.000 & 0.000 & -1.000 \\
    \quad ---Prompt-B & 0.000 & 0.000 & 0.000 & -1.000 & 0.000 & 0.000 & 0.000 & -1.000 \\
    \quad ---SFT & 0.260 & 0.160 & 0.210 & -0.583 & 0.433 & 0.187 & 0.310 & -0.583 \\ \midrule
 
    Qwen3-8B & 0.760 & 0.307 & 0.533 & 0.075 & 0.780 & 0.180 & 0.480 & -0.050 \\
    \quad ---CoT & 0.880 & 0.273 & 0.577 & 0.192 & 0.700 & 0.200 & 0.450 & -0.115 \\
    \quad ---Prompt-B & 0.840 & 0.380 & 0.610 & 0.248 & 0.680 & 0.500 & 0.590 & 0.183 \\
    \quad ---SFT & 0.980 & 0.620 & 0.800 & 0.643 & 0.980 & 0.540 & 0.760 & 0.579 \\ \midrule
    
    Qwen3-14B & 0.773 & 0.320 & 0.547 & 0.105 & 0.700 & 0.333 & 0.517 & 0.036 \\
    \quad ---CoT & 0.840 & 0.300 & 0.570 & 0.167 & 0.740 & 0.327 & 0.533 & 0.073 \\
    \quad ---Prompt-B & 0.700 & 0.420 & 0.560 & 0.125 & 0.660 & 0.460 & 0.560 & 0.122 \\ \midrule
    
    Qwen3-32B & 0.920 & 0.273 & 0.597 & 0.253 & 0.840 & 0.320 & 0.580 & 0.187 \\
    \quad ---CoT & 0.913 & 0.247 & 0.580 & 0.215 & 0.800 & 0.360 & 0.580 & 0.178 \\
    \quad ---Prompt-B & 0.800 & 0.320 & 0.560 & 0.137 & 0.840 & 0.260 & 0.550 & 0.123 \\
    \quad ---SFT & 0.960 & 0.540 & 0.750 & 0.551 & 0.940 & 0.500 & 0.720 & 0.490 \\ \midrule

  \multicolumn{9}{c}{\textit{Commercial Models}} \\ \midrule
    Claude-4.5-Sonnet & 0.620 & 0.740 & 0.680 & 0.363 & 0.660 & 0.680 & 0.670 & 0.340 \\
    \quad ---CoT & 0.640 & 0.700 & 0.670 & 0.341 & 0.700 & 0.800 & 0.750 & 0.503 \\ 
    \quad ---Prompt-B & 0.240 & 0.560 & 0.400 & -0.211 & 0.440 & 0.520 & 0.480 & -0.040 \\ \midrule
    
    DeepSeek-v3 & 0.980 & 0.380 & 0.680 & 0.450 & 0.980 & 0.360 & 0.670 & 0.433 \\
    \quad ---CoT & 0.940 & 0.300 & 0.620 & 0.312 & 0.920 & 0.300 & 0.610 & 0.280 \\
    \quad ---Prompt-B & 0.880 & 0.340 & 0.610 & 0.261 & 0.860 & 0.400 & 0.630 & 0.293 \\ 
    \quad ---RAG & 0.160 & 0.860 & 0.510 & 0.068 & 0.080 & 0.920 & 0.500 & -0.014 \\ 
    \midrule
    
    DeepSeek-r1 & 0.980 & 0.300 & 0.640 & 0.382 & 0.980 & 0.380 & 0.680 & 0.450 \\
    \quad ---CoT & 0.980 & 0.240 & 0.610 & 0.327 & 0.980 & 0.360 & 0.670 & 0.433 \\ 
    \quad ---Prompt-B & 0.880 & 0.260 & 0.570 & 0.178 & 0.860 & 0.300 & 0.580 & 0.193 \\ \midrule

    DeepSeek-v4-flash & 0.820 & 0.840 & 0.830 & 0.660 & 0.840 & 0.820 & 0.830 & 0.660 \\ \midrule
 
    GPT-5.2-chat & 0.820 & 0.800 & 0.810 & 0.620 & 0.700 & 0.780 & 0.740 & 0.482 \\
    \quad ---CoT & 0.760 & 0.800 & 0.780 & 0.560 & 0.700 & 0.800 & 0.750 & 0.503 \\
    \quad ---Prompt-B & 0.860 & 0.700 & 0.780 & 0.567 & 0.760 & 0.660 & 0.710 & 0.422 \\
    \quad ---RAG & 0.800 & 0.860 & 0.830 & 0.661 & 0.680 & 0.820 & 0.750 & 0.500 \\ \midrule
    Voting Ensemble (DSV3/DSR/C) & 0.980 & 0.380 & 0.680 & 0.450 & 0.980 & 0.380 & 0.680 & 0.450 \\
    Voting Ensemble (DSV4/GPT/C) & 0.820 & 0.840 & 0.830 & 0.660 & 0.800 & 0.840 & 0.820 & 0.641 \\
    \bottomrule
  \end{tabular}
  \caption{Detailed performance statistics in 64k, 100k scales. \texttt{Prompt-B} indicates that the target summary is positioned at the Beginning, while the default setup uses \texttt{Prompt-E}, which positions the target summary at the End. Voting Ensembles are abbreviation-coded as follows: (DSV3/DSR/C) denotes DeepSeek-v3, DeepSeek-r1, and Claude-sonnet-4-5-20250929-v1;  (DSV4/GPT/C) denotes DeepSeek-v4-flash, GPT-5.2-chat, and Claude-sonnet-4-5-20250929-v1.}
\label{Detailed performance statistics in 64k, 100k scales.}
\end{table*} 

\section{Full Results}
\label{app:Full results}
As shown in Table~\ref{tab:Detailed performance statistics in 16k, 32k scales.} and Table~\ref{Detailed performance statistics in 64k, 100k scales.}, models exhibit distinct biases during hallucination detection. Open-source models generally struggle with detecting hallucinated instances. For example, base models like InternLM2.5-20B, Minicheck-7B, and GLM-4-9B-chat exhibit extremely limited capabilities in detecting hallucinations across almost all context lengths. In contrast, commercial models, such as Claude-4.5-Sonnet, DeepSeek-v4-flash, and GPT-5.2-chat, demonstrate a much stronger capability in successfully identifying hallucinated labels.

Regarding prompt positioning, the placement of the target summary significantly impacts evaluation efficacy. When the target summary is positioned at the beginning (Prompt-B), models fail to effectively integrate the summary with the extensive long-article context, leading to a noticeable decline in detection accuracy across most open-source and commercial architectures. Conversely, placing the target summary at the end enables superior and more stable performance across models.

\begin{figure*}[t!]
  \centering
  \includegraphics[width=0.9\textwidth]{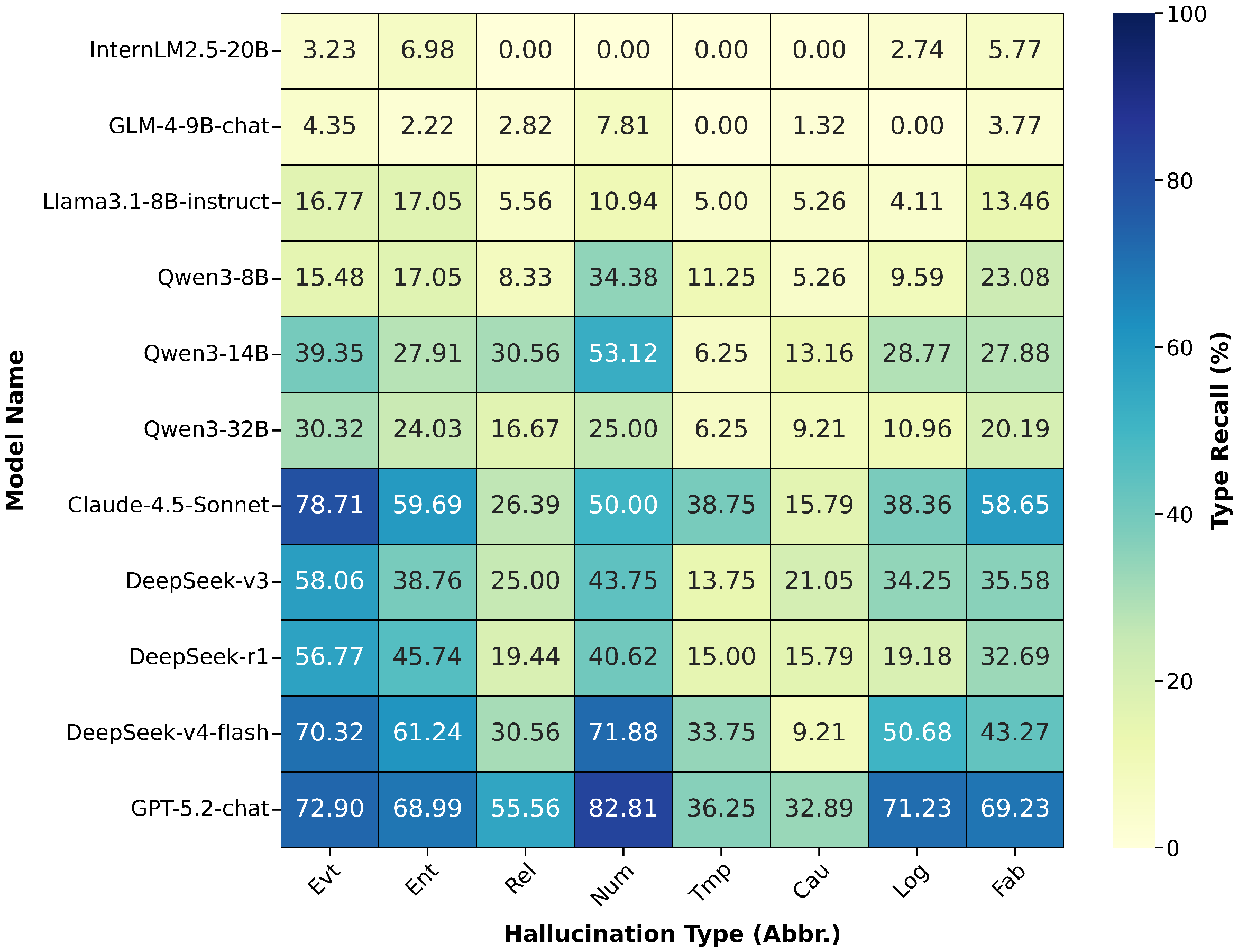}
  \caption{Recall performance of various LLMs across different hallucination types. The hallucination types are abbreviated as follows: Evt: Event Hallucination, Ent: Entity Hallucination, Rel: Relation Hallucination, Num: Numerical Hallucination, Tmp: Temporal Hallucination, Cau: Causal Hallucination, Log: Logical Inversion, and Fab: Event Fabrication.}
  \label{fig:Recall performance of various LLMs across different hallucination types.}
\end{figure*}

As illustrated in Fig.~\ref{fig:Recall performance of various LLMs across different hallucination types.}, there is a contrast in detection difficulty across different hallucination types. Event Hallucinations and Numerical Hallucinations stand out as the most detectable categories; for instance, Claude-4.5-Sonnet achieves its highest recall of 78.71\% in Event Hallucination, while GPT-5.2-chat reaches a peak recall of 82.81\% in Numerical Hallucination. In contrast, Temporal Hallucinations and Causal Hallucinations tend to be the most difficult to detect, representing a significant challenge for all tested models. Many models, such as InternLM2.5-7B and Minicheck-7B, show recall rates of 0.00\% for both types, and even GPT-5.2-chat struggles significantly with a recall of only 36.25\% in the temporal category. These results indicate that while models are proficient at flagging isolated errors in numerical data or basic factual attributes, identifying instances where the content exists but its chronological order or causal relationships have been subtly altered remains significantly more challenging for current LLMs.

To calculate these recall rates, the model's reasoning paths are analyzed to verify whether it correctly identifies the same issue described in the ground truth. If the model and the ground truth point to the same factual error or event, that hallucination type is marked as a hit. It is important to note that a single data point can contain multiple types of hallucinations at once. This explains why some models have high overall detection accuracy but lower type-specific recall. A model might get a correct detection score by finding just one error in the data, whereas another model might be better at identifying all the different types of hallucinations present in that same data, leading to a higher recall for those categories.

\section{Ablation Study for RAG}

\label{Ablation Study for RAG}

\begin{figure*}[t!]
  \centering
  \includegraphics[width=0.9\textwidth]{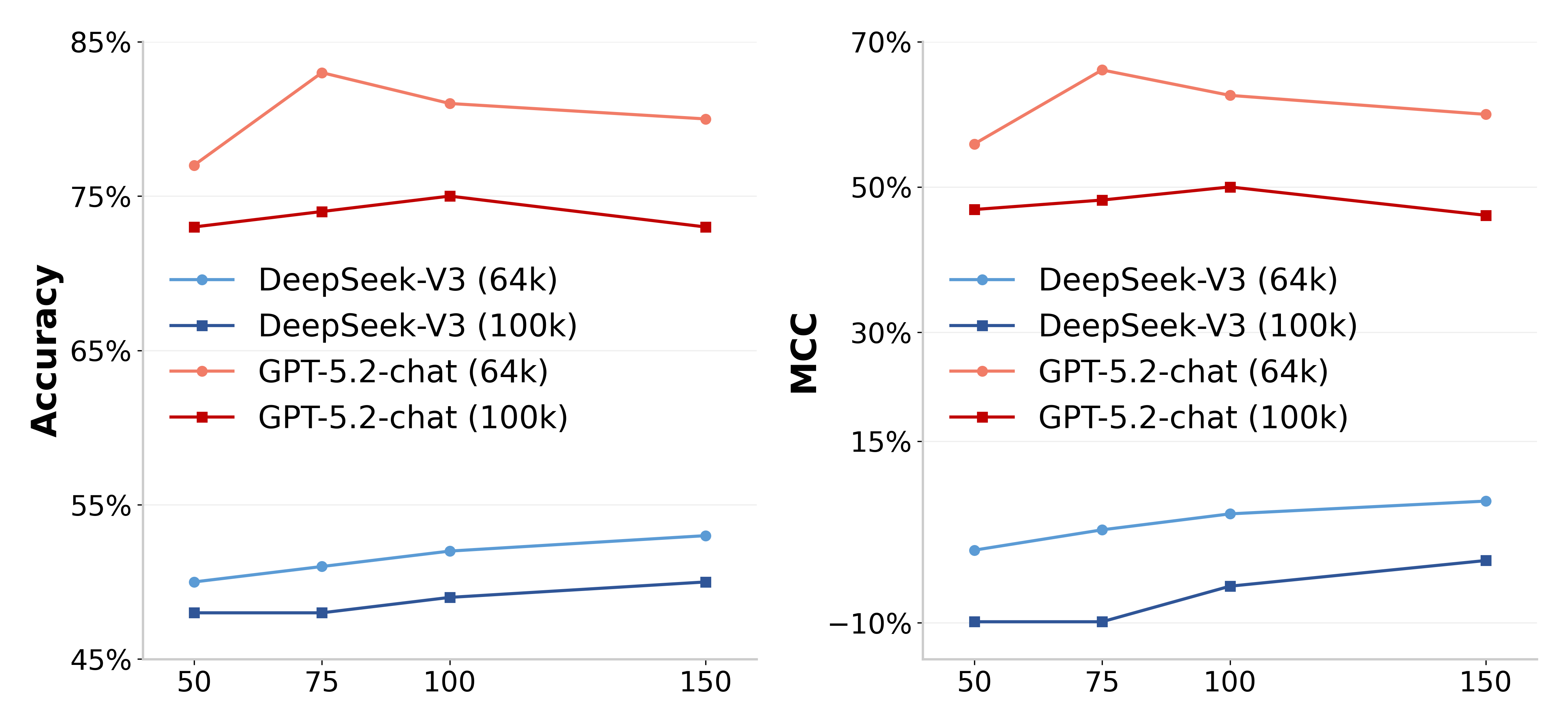}
  \caption{Comparison of different summary chunk sizes on RAG performance of Deepseek-v3 and GPT-5.2-chat.}
  \label{fig:Results comparison of different summary chunk sizes on RAG performance of Deepseek-v3 and GPT-5.}
\end{figure*}

To optimize the RAG performance, we conduct experiments comparing different target summary chunk sizes, as illustrated in Fig.~\ref{fig:Results comparison of different summary chunk sizes on RAG performance of Deepseek-v3 and GPT-5.}. Rather than strictly adhering to a rigid character limit, our chunking mechanism dynamically preserves complete sentence boundaries; if a succeeding sentence begins with a pronoun, it is automatically merged into the current chunk to preserve contextual continuity. The experimental results reveal distinct behavioral patterns across models. For DeepSeek-v3, accuracy consistently scales up as the chunk size increases from 50 to 150. This is primarily because smaller chunks multiply the number of segments and model calls, causing individual judgment biases to accumulate heavily under our conjunctive rule. Conversely, GPT-5.2-chat exhibits a concave performance curve, peaking at 75 in the 64k context and 100 in the 100k context. Benefiting from a higher baseline judgment accuracy, GPT-5.2-chat is less vulnerable to error accumulation in smaller chunks. However, when the chunk size expands, it aggregates multiple disparate factual claims into a single block, which dilutes the localized focus of the RAG windows and the strategic benefits of segmentation are ultimately outweighed by the accumulation of errors, leading to a decline in performance.

\begin{figure*}[t!]
  \centering
  \includegraphics[width=0.9\textwidth]{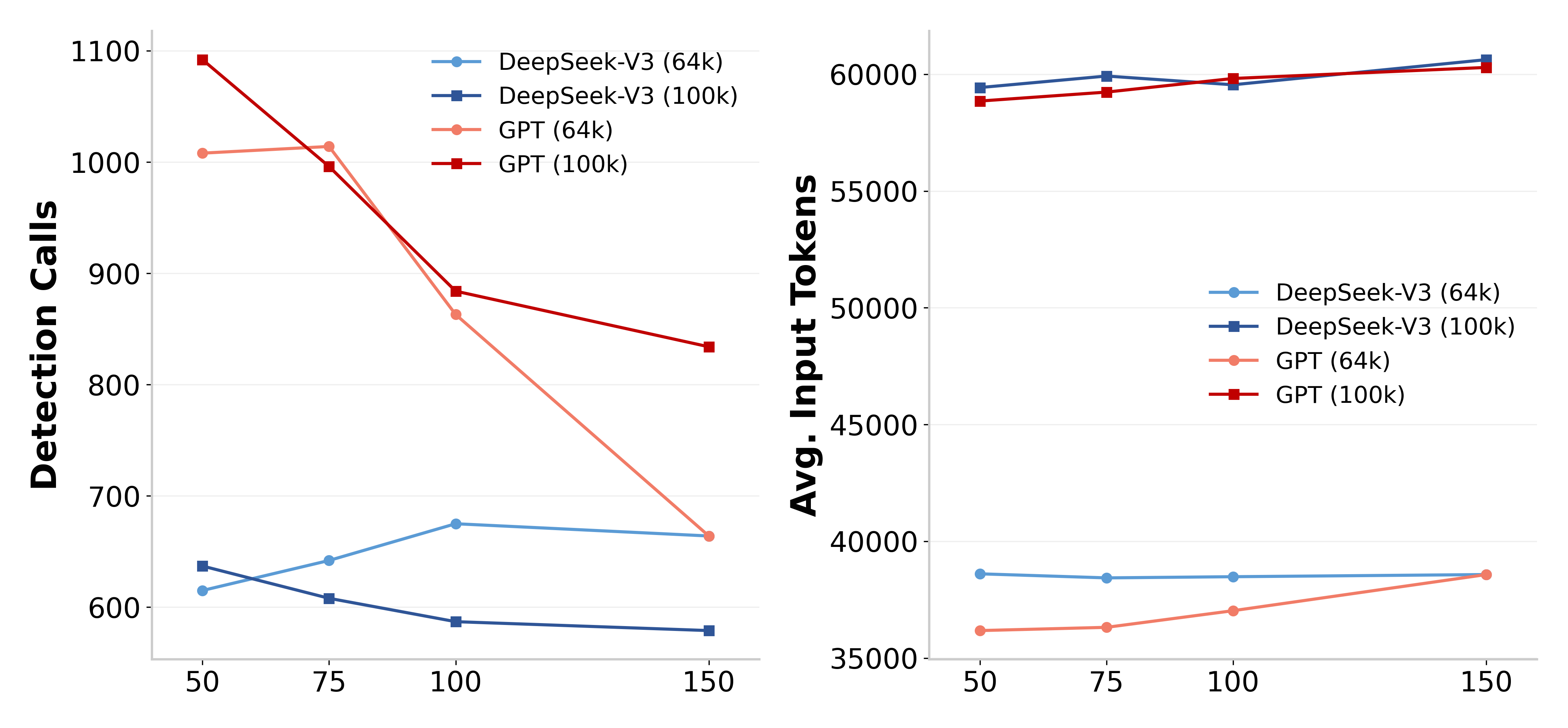}
  \caption{Comparison of model calls and input token consumptions on RAG performance of Deepseek-v3 and GPT-5.}
  \label{fig:Comparison of model calls and input token consumptions on RAG performance of Deepseek-v3 and GPT-5.}
\end{figure*}

Furthermore, evaluating paired summary-article blocks through our sliding-window framework yields a significant reduction in token expenditure compared to conducting hallucination detection over the entire article, as illustrated in Fig.~\ref{fig:Comparison of model calls and input token consumptions on RAG performance of Deepseek-v3 and GPT-5.}. Based on the Qwen3 tokenizer, our RAG strategy successfully curtails the average input token count per evaluation call—limiting it to approximately 36k–38k tokens for the 64k setting and around 59k tokens for the 100k setting. By eliminating the massive overhead of re-processing the full article for every sub-step verification, this approach dramatically optimizes token consumption while ensuring sufficient logical continuity for precise factual alignment.

\clearpage
\onecolumn 

\section{Hallucination Taxonomy and Case Examples}
\label{app:hallucination_examples}

To assist readers in understanding our benchmark, this section provides concrete examples for both cross-chapter hallucination verification and the hallucination categories defined in \textsc{LongNovel}.

\subsection{Cross-Chapter Consistency Case Study}
\label{app:cross_chapter_case}
Hallucinations are generated from long contexts spanning multiple chapters, requiring models to track context across large distances rather than relying on local sentence-level matching.

\begin{figure}[!ht]
\begin{tcolorbox}[casebox, title={Case Study: Cross-Chapter Hallucination Verification}, fontupper=\fontsize{9pt}{12pt}\selectfont]
\textbf{Article}:\\
“I no longer wish to be a Red Dust Immortal.” I knelt before my master’s grave, and another three thousand years passed. I heard a tremendous sound between heaven and earth and knew it was time to descend the mountain. I returned to the place where I had once met the old monk, and there I encountered a young monk......

\textbf{... (across multiple chapters) ...}

Su Qin lay on the bed and opened his eyes. “I'm not dead?” he asked in disbelief, gently raising his right arm and turning it before his eyes. He remembered getting drunk at a nightclub and staggering back to his hotel......\\
\\
\textbf{Hallucinated Summary}:\\
\textcolor{red}{Su Qin} gradually realizes that he may be the only Red Dust Immortal in the world. Eventually, \textcolor{red}{Su Qin kneels before his master's grave for three thousand years}, comes to understand his true identity, and decides to give up the pursuit of immortality and return to the mortal world.\\
\\
\textbf{Analysis}:\\
This example requires cross-chapter verification because the hallucinated summary incorrectly attributes the first-person narrator's experiences, including kneeling before his master's grave for three thousand years, to Su Qin, who is introduced only in a later chapter. Detecting this error requires tracking narrative identity across distant chapters rather than verifying a locally stated fact.
\end{tcolorbox}
\end{figure}

\subsection{Fine-Grained Hallucination Categories Case Studies}
\label{app:fine_grained_cases}
Below, we present concrete examples illustrating distinct fine-grained hallucination categories found in summary outputs.

\begin{figure}[!ht]
\begin{tcolorbox}[casebox, title={Case 1: Entity Hallucination and Logical Inversion}, fontupper=\fontsize{9pt}{12pt}\selectfont]
\textbf{Original Text}:\\
Mrs. John Dashwood did not at all approve of what her husband intended to do for his sisters. To take three thousand pounds from the fortune of their dear little boy would be impoverishing him to the most dreadful degree. ...... “Well, then, let something be done for them; but that something need not be three thousand pounds,” she added. ...... “Five hundred pounds would be a prodigious increase to their fortunes.” ...... “Indeed, to say the truth, I am convinced within myself that your father had no idea of your giving them any money at all.”\\
\dots\\
This circumstance was a growing attachment between her eldest girl and the brother of Mrs. John Dashwood, a gentleman-like and pleasing young man. ...... Edward Ferrars was the eldest son of a man who had died very rich. ...... He loved her daughter, and Elinor returned the partiality.\\
\\
\textbf{Faithful Summary}:\\
Though John promised to care for them, his wife Fanny persuaded him to give them nothing. While remaining temporarily at Norland, Elinor attached herself to Fanny's brother, Edward Ferrars. Fanny disapproved, desiring a wealthy match for him.\\
\\
\textbf{Hallucinated Summary}:\\
Though John promised to care for them, his wife Fanny persuaded him to \textcolor{red}{provide for them generously}. While remaining temporarily at Norland, Elinor attached herself to Fanny's brother, \textcolor{red}{Robert Ferrars}. Fanny disapproved, desiring a wealthy match for him.\\
\\
\textbf{Analysis}:\\
The hallucinated summary contains two distinct errors. First, it reverses the logical meaning by claiming that Fanny persuaded John to provide \textit{generously}, whereas the source explicitly states that she continuously discouraged him from giving them money (\textbf{Logical Inversion}). Second, it replaces \textit{Edward Ferrars} with \textit{Robert Ferrars}, introducing an incorrect entity (\textbf{Entity Hallucination}).
\end{tcolorbox}
\end{figure}

\begin{figure}[!ht]
\begin{tcolorbox}[casebox, title={Case 2: Numerical Hallucination}, fontupper=\fontsize{9pt}{12pt}\selectfont]
\textbf{Original Text}:\\
The Jiangnan Examination Hall was first established in the fourth year of the Qiandao era of the Song dynasty, corresponding to the eighth year of the Common Era, and was then known as the Jiankang Prefecture Examination Hall. In the early Ming dynasty, it was simply called the “Examination Hall” and served as a venue for both provincial and metropolitan examinations. ...... Until the abolition of the imperial examination system in the late Qing dynasty, the Nanjing Examination Hall remained the largest of its kind in the world. At its peak, it covered more than three hundred thousand square meters, contained over twenty thousand examination cells, and accommodated \textbf{more than twenty thousand six hundred candidates}. ...... The Mingyuan Tower in the Jiangnan Examination Hall was built in the thirteenth year of the Jiajing era of the Ming dynasty and has been preserved for nearly five hundred years. It remains the oldest surviving examination-hall structure in China.\\
\\
\textbf{Faithful Summary}:\\
The Jiangnan Examination Hall was the largest imperial examination venue in China at the time, and during the Ming and Qing dynasties, large numbers of successful examination candidates emerged from it.\\
\\
\textbf{Hallucinated Summary}:\\
The examination hall was a venue for imperial examinations held in the capital and provincial capitals. Its Mingyuan Tower is the oldest surviving examination-hall structure in China, and at its largest scale, the examination hall could accommodate \textcolor{red}{more than 26000 candidates}.\\
\\
\textbf{Analysis}:\\
The source explicitly states that the examination hall accommodated \textit{more than twenty thousand six hundred candidates} ($>20,600$), whereas the hallucinated summary distorts this numerical detail to \textit{more than 26,000} (\textbf{Numerical Hallucination}).
\end{tcolorbox}
\end{figure}

\begin{figure}[!ht]
\begin{tcolorbox}[casebox, title={Case 3: Event Hallucination}, fontupper=\fontsize{9pt}{12pt}\selectfont]
\textbf{Original Text}:\\
“Dead! dead! dead!” Under the influence of any strongly disturbing force, Clare would occasionally walk in his sleep. ...... The spot was lonely, the river deep and wide enough to make such a purpose easy of accomplishment. He might drown her if he would; it would be better than parting tomorrow to lead severed lives. ...... If they could both fall together into the current now, their arms would be so tightly clasped together that they could not be saved. [...]\\
\\
\textbf{Faithful Summary}:\\
Tess cannot determine Angel's ultimate intention, but finally realizes that he is dreaming about the Sunday when he carried her across the water with the other milkmaids. He carries her near the river, and she believes he may drown her. He walks through the shallow areas of the river carrying her, but they reach the other side in safety. Angel carries her to the empty stone coffin of an abbot, where he lays Tess and then falls asleep. Tess sits up in the coffin but does not awaken Angel, out of fear that he may die if awakened while sleepwalking.\\
\\
\textbf{Hallucinated Summary}:\\
On the night before their departure, Clare sleepwalks, carrying Tess to the river, \textcolor{red}{where he nearly drowns her before she awakens him}. The next morning, they part in silence, each carrying the weight of their sorrow. Tess returns to her family, who are shocked and disappointed by her return without her husband.\\
\\
\textbf{Analysis}:\\
In the source, Clare safely carries Tess across the river without nearly drowning her. Furthermore, the summary fabricated the event that \textit{she awakens him}, whereas the original text explicitly highlights that Tess deliberately refrains from waking him out of fear (\textbf{Event Hallucination}).
\end{tcolorbox}
\end{figure}

\clearpage
\section{Case Studies on Detection Failures}
\label{app:failure_cases}

\subsection{Failure in Comprehensively Understanding}
This case demonstrates that the model failed to comprehensively understand the text. Although the article explicitly states that the Emperor led 40,000 troops to suppress the rebellion in Le'an, while the 70,000 troops were sent to Cochin, the model fails to cross-reference these distinct numbers, leading to an incorrect judgment.

\begin{figure}[!ht] %
\begin{tcolorbox}[casebox, title=Case on Deepseek-v4 with 16k Dataset,fontupper=\fontsize{9pt}{12pt}\selectfont]
\textbf{Article}: ......朱瞻基伸头看过，{\textbackslash "}交战已阅数载，尸填红河之岸，血满蓝山之窟。何不收此残局，为百姓之康宁，为交趾之自存，开万世太平之基，倾全力于将来。”笑道：“就是{\textbackslash n} 这个意思。‘交趾’改为‘安南',更好。”{\textbackslash n} 珠璇大喜道：“‘安南’?真的?你愿意?”这是同意安南复国了。{\textbackslash n} 朱瞻基含笑点头：“是。不过柳升前儿已经出发了，你这信我让兵部交征夷将军{\textbackslash n} 王通吧。{\textbackslash "}{\textbackslash n} 珠璇愕然：“安远侯已经走了?带了多少兵马?”{\textbackslash n} 朱瞻基叹口气：“\textcolor{red}{七万}。”顿了顿道：{\textbackslash "}这六七年打下来，朝廷耗费的军粮钱财无{\textbackslash n} 数，夏原吉直叫苦。”{\textbackslash n}
......
宣德元年八月二十日，宣德皇帝朱瞻基率军亲征，赴山东乐安平叛。照例是锦衣卫在中簇拥，五军营内圈，三千营外圈，神机营穿插其间。\textcolor{red}{四万}人马浩浩荡荡，直奔{\textbackslash n} 山东。皇帝一向离不开的荣冬荣夏两位镇抚却没看到，杨荣有些奇怪，忍住了没问。{\textbackslash n} 兵贵神速，大军赶了两日便已经过河北进了山东，第三日上午到了山东德州。张辅问道：“陛下!是直接奔乐安吗?汉王传檄已近十日，会不会已经出了乐安?”杨荣也道：“汉王倘若出乐安，两个可能：一是占济南，二是干脆奔南京。这两地都有不{\textbackslash n} 少汉王的旧部，汉王趁机占地与朝廷对抗，便比较棘手了。”{\textbackslash n} 朱瞻基笑了笑：“杨卿说的是‘倘若’,可惜汉王没有这么高的才干。朕料他定然还在原地。”说着吩咐：“大军急行军，明夜务必到乐安!”“啪”地一击马，当先{\textbackslash n} 便行。{\textbackslash n} 张辅与杨荣对望一眼，连忙策马跟上。皇帝，为什么那么自信?{\textbackslash n} 第五日寅时，\textcolor{red}{四万}大军兵临乐安城下。正是黎明前最黑暗的辰光，朱瞻基吩咐张辅安排大军扎营，耀武扬威地一字大营就扎在乐安城门之前。神机营的一百门神机铳{\textbackslash n} 炮，同样一字排开，炮筒齐齐扬起，对准了乐安城楼。{\textbackslash n}
......
辰时一刻，\textcolor{red}{四万}大军收兵归营歇息。汉王朱高煦筹划了二十年的造反，模仿父亲{\textbackslash n} 的“靖难”,闹剧一般收场。朱瞻基只用了几天时间便平叛成功。......
\\
\textbf{Summary}: ......朱瞻基决定亲征乐安平叛，珠璇请求随行未果，但两人商议安南复国事宜。朱瞻基派兵\textcolor{red}{七万}，平叛成功，汉王朱高煦投降。......
\\
\textbf{Deepseek-v4}:  {\textbackslash "}score{\textbackslash "}: 5,{\textbackslash n} \ \ {\textbackslash "}reason{\textbackslash "}: {\textbackslash "}摘要准确还原了原文的主要情节，无任何幻觉。\"{\textbackslash "}

\end{tcolorbox}
\end{figure}

\begin{figure}[!ht] %
\begin{tcolorbox}[casebox, title=English Version: Case on Deepseek-v4 with 16k Dataset,fontupper=\fontsize{9pt}{12pt}\selectfont]
\textbf{Article}: ......Zhu Zhanji leaned over to look, {\textbackslash "}The war has lasted for years, with corpses filling the banks of the Red River and blood overflowing the caves of Lam Son. Why not end this mess, for the peace of the people and the survival of Jiaozhi, to lay the foundation for eternal peace and devote all our efforts to the future." He smiled and said: "That is exactly{\textbackslash n} the meaning. Changing 'Jiaozhi' to 'Annam' would be even better."{\textbackslash n} Zhuxuan was overjoyed: "'Annam'? Really? You are willing?" This meant agreeing to the restoration of Annam.{\textbackslash n} Zhu Zhanji nodded with a smile: "Yes. However, Liu Sheng already set off the day before yesterday. I will have the Ministry of War hand this letter to the Conquering Barbarians General{\textbackslash n} Wang Tong."{\textbackslash "}{\textbackslash n} Zhuxuan was stunned: "The Marquis of Anyuan has already left? How many troops did he take?"{\textbackslash n} Zhu Zhanji sighed: "\textcolor{red}{70,000}." He paused and said: {\textbackslash "}After fighting for these six or seven years, the court has exhausted countless military rations and wealth,{\textbackslash n} and Xia Yuanji has been complaining bitterly."{\textbackslash n}
......
On August 20 of the first year of Xuande, Emperor Zhu Zhanji personally led the army to Le'an, Shandong to suppress the rebellion. As usual, the Jinyiwei clustered in the center, surrounded by the Five Armies Battalion in the inner circle, the Three Thousand Battalion in the outer circle, with the Shenji Battalion interspersed among them. The \textcolor{red}{40,000} troops marched majestically, heading straight for{\textbackslash n} Shandong. Rong Dong and Rong Xia, the two commanders the emperor usually kept close, were nowhere to be seen. Yang Rong found it a bit strange but held back his questions.{\textbackslash n} Speed is crucial in war. The great army rushed for two days, passing through Hebei into Shandong, and arrived at Dezhou, Shandong on the morning of the third day. Zhang Fu asked: "Your Majesty! Are we heading straight for Le'an? The Prince of Han issued his call to arms nearly ten days ago; could he have already left Le'an?" Yang Rong also said: "If the Prince of Han leaves Le'an, there are two possibilities: one is to occupy Jinan, the other is to head straight for Nanjing. Both places have quite a{\textbackslash n} few of the Prince of Han's former subordinates. If he seizes the territory to confront the court, it will be quite troublesome."{\textbackslash n} Zhu Zhanji smiled: "Minister Yang said 'if', unfortunately the Prince of Han lacks such high capabilities. I predict he is definitely still where he was." Saying this, he ordered: "The army will march double-time, we must reach Le'an by tomorrow night!" With a "crack," he whipped his horse and took the lead{\textbackslash n} to move out.{\textbackslash n} Zhang Fu and Yang Rong glanced at each other, and quickly spurred their horses to follow. Why was the emperor so confident?{\textbackslash n} At the Yin hour of the fifth day, the \textcolor{red}{40,000} strong army arrived at the gates of Le'an. It was the darkest time just before dawn. Zhu Zhanji ordered Zhang Fu to set up camp for the army, conspicuously pitching a line-shaped camp right in front of the city gates of Le'an. The one hundred firearms of the Shenji Battalion{\textbackslash n} were also lined up, with their barrels raised in unison, aiming straight at the Le'an city tower.{\textbackslash n}
......
At a quarter past the Chen hour, the \textcolor{red}{40,000} strong army withdrew to their camp to rest. The rebellion that Zhu Gaoxu, the Prince of Han, had planned for twenty years, mimicking his father's{\textbackslash n} "Jingnan" campaign, ended like a farce. Zhu Zhanji succeeded in putting down the rebellion in just a few days. ......
\\
\textbf{Summary}: ......Zhu Zhanji decided to personally lead an expedition to Le'an to suppress the rebellion. Zhuxuan's request to accompany him was denied, but the two discussed the restoration of Annam. Zhu Zhanji dispatched \textcolor{red}{70,000} troops, successfully suppressed the rebellion, and Zhu Gaoxu, the Prince of Han, surrendered. ......
\\
\textbf{Deepseek-v4}:  {\textbackslash "}score{\textbackslash "}: 5,{\textbackslash n} \ \ {\textbackslash "}reason{\textbackslash "}: {\textbackslash "}The summary accurately reproduces the main plot of the original text, without any hallucinations.{\textbackslash "}

\end{tcolorbox}
\end{figure}

\clearpage

\subsection{Hallucination Correction for Faithful Sentences}
This case is a non-hallucination example. However, the model erroneously identifies multiple faithful sentences as hallucinations. The reasoning demonstrates a severe logical flaw, where the model claims to correct these sentences, yet the corrected content it provides is literally identical to the original text.

\begin{figure}[!ht] %
\begin{tcolorbox}[casebox, title=Case on GLM with 16k Dataset,fontupper=\fontsize{9pt}{12pt}\selectfont]
\textbf{Summary}: ...... A man approaches Tess at Flintcomb-Ash while her companions watch. They fail to recognize him as Alec, for he has abandoned his parson persona and shaved his beard, appearing again as a dandy. He tells Tess he has quit preaching and demands she stop working there. Mocking her absent husband as a mythological personage, he insists she leave him. Tess \textcolor{red}{strikes Alec with her glove}, drawing blood. \textcolor{red}{He threatens that he will be her master again}.{\textbackslash n}{\textbackslash n}Alec continues visiting, \textcolor{red}{offering to assist her family} if she shows confidence in him. Tess refuses his help. Desperate, \textcolor{red}{she writes to Angel}, pleading for his return and offering to live as his servant. \textcolor{red}{The Clares receive this letter to forward}, lamenting Angel’s marriage and blaming themselves. Meanwhile, Angel, aged by his absence, begins reconsidering his rejection of Tess, recalling Izz Huett’s praise of her love.{\textbackslash n}{\textbackslash n}Liza-Lu brings news that both parents are ill. Tess returns to Marlott to find a neighbor caring for her mother. Her father, John, foolishly plans to ask antiquarians to fund his life as a historical relic. Alec tracks Tess down, mocking her hope for Angel’s return. \textcolor{red}{He insists on helping her despite her protests}. Upon returning home, \textcolor{red}{Tess discovers her father has died}.......
\\
\textbf{GLM}: {\textbackslash "}reason{\textbackslash "}: {\textbackslash "}Event Hallucination, 、\textcolor{blue}{changing 'striking Alec with her glove' to 'striking Alec with her glove'}; Event Hallucination, \textcolor{blue}{changing 'threatening to be her master again' to 'threatening that he will be her master again'}; Event Hallucination, \textcolor{blue}{changing 'offering to assist her family' to 'insisting on helping her'}; Event Hallucination, \textcolor{blue}{changing 'writing to Angel' to 'writing to Angel, pleading for his return and offering to live as his servant'}; Event Hallucination, \textcolor{blue}{changing 'receiving the letter' to 'receiving this letter to forward'}; Event Hallucination, \textcolor{blue}{changing 'insisting on helping her despite her protests' to 'insisting on helping her despite her protests'}; Event Hallucination, \textcolor{blue}{changing 'discovering her father has died' to 'discovering her father has died'},{\textbackslash "},{\textbackslash n} \ \ {\textbackslash "}score{\textbackslash "}: 1,{\textbackslash n} \ \ {\textbackslash "}hallucination\_types{\textbackslash "}: [{\textbackslash "}Entity Hallucination{\textbackslash "}]
\end{tcolorbox}
\end{figure}

\clearpage
\subsection{Limitations in Logical Reasoning}
These cases show that the model fails to reason properly, leading to both false negatives and false positives in hallucination detection. In the first case, the model overlooks entity-attribute bindings and fails to detect that a character's talent is wrongly attributed to another. In the second case, the model falsely claims a hallucination exists by contradicting explicit textual facts regarding the characters' backgrounds, leading to an incorrect judgment.

\begin{figure}[!ht] %
\begin{tcolorbox}[casebox, title=Case on Qwen3-32B with 64k Dataset,fontupper=\fontsize{9pt}{12pt}\selectfont]
\textbf{Article}: ......“\textcolor{red}{李夫人看向陈萤。{\textbackslash n}    找东西正好是我的专长，我的天赋是【万物通灵】。” 高阳仔细回想，的确有这个天赋。}{\textbackslash n}    【万物通灵】，序列号81，辅助系。{\textbackslash n}    主人可以通过触摸物品，看到过去发生在这件物品身上的事情。{\textbackslash n}   \textcolor{red}{ 陈萤把照片还给高阳：“不过我必须触碰实物，照片没用。”}{\textbackslash n}    高阳接回照片，点点头。{\textbackslash n}    其实他也猜到了，如果光看照片就能通灵，这能力太逆天了，序列号不可能这么靠后。{\textbackslash n}    李夫人接话道：“只要找到这辆车，陈萤就可以帮上忙了。”{\textbackslash n}    高阳抬头：{\textbackslash n}    什么意思？看来还有后续。{\textbackslash n}    李夫人微笑：“这辆车，百川团可以试着帮你找一找。如果短时间内能找到，陈萤的通灵就还有效。”{\textbackslash n}    高阳直接问：“条件是什么？”{\textbackslash n}    “七影队长是爽快人。”{\textbackslash n}    李夫人的语气不疾不徐：“我先界定一下我们的工作范围，帮你找到这辆车，陈萤再通过车帮你找到那个人的线索，这份工作就算完成。至于你接下来要做什么，我们不会干涉，也没兴趣知道。”{\textbackslash n}    高阳点头：“这样最好不过。”{\textbackslash n}   ......
\\
\textbf{Summary}: \textcolor{red}{高阳拥有【万物通灵】的天赋}，可以通过触摸物品看到过去发生的事情。.......
\\
\textbf{Qwen3-32B}: {\textbackslash "}score{\textbackslash "}: 5,{\textbackslash n}  {\textbackslash "}reason{\textbackslash "}: {\textbackslash "}完美契合，无任何幻觉{\textbackslash "}
\\
\textbf{Ground Truth}: "reason": "实体幻觉：摘要称“高阳拥有【万物通灵】的天赋”，\textcolor{blue}{但原文明确该天赋属于陈萤而非高阳}；将陈萤的核心能力错误指派给高阳，属于关键设定错误。", "score": 2, "type": ["实体幻觉"]
\end{tcolorbox}
\end{figure}

\begin{figure}[!ht] %
\begin{tcolorbox}[casebox, title=English Version: Case on Qwen3-32B with 64k Dataset,fontupper=\fontsize{9pt}{12pt}\selectfont]
\textbf{Article}: ......"\textcolor{red}{Mrs. Li looked at Chen Ying.{\textbackslash n}    'Finding things happens to be my specialty. My talent is [Psychometry].' Gao Yang thought back carefully; indeed, there was such a talent.}{\textbackslash n}    [Psychometry], serial number 81, Support type.{\textbackslash n}    Its master can see what happened to an object in the past by touching it.{\textbackslash n}   \textcolor{red}{ Chen Ying returned the photo to Gao Yang: 'But I must touch the actual object; photos are useless.'}{\textbackslash n}    Gao Yang took back the photo and nodded.{\textbackslash n}    He had actually guessed as much. If she could commune with spirits just by looking at a photo, this ability would be too overpowered, and its serial number wouldn't be so far back.{\textbackslash n}    Mrs. Li chimed in: 'As long as we find this car, Chen Ying can help.'{\textbackslash n}    Gao Yang looked up:{\textbackslash n}    What does this mean? It seems there's more to follow.{\textbackslash n}    Mrs. Li smiled: 'The Baichuan Group can try to help you find this car. If it can be found in a short time, Chen Ying's psychometry will still be effective.'{\textbackslash n}    Gao Yang asked directly: 'What are the conditions?'{\textbackslash n}    'Captain Seven Shadows is a straightforward person.'{\textbackslash n}    Mrs. Li's tone was unhurried: 'Let me first define the scope of our work. Helping you find this car, and then having Chen Ying help you find clues about that person through the car—that will be considered our job done. As for what you do next, we won't interfere, nor are we interested in knowing.'{\textbackslash n}    Gao Yang nodded: 'That couldn't be better.'{\textbackslash n}   ......
\\
\textbf{Summary}: \textcolor{red}{Gao Yang possesses the [Psychometry] talent}, and can see past events by touching objects. .......
\\
\textbf{Qwen3-32B}: {\textbackslash "}score{\textbackslash "}: 5,{\textbackslash n}  {\textbackslash "}reason{\textbackslash "}: {\textbackslash "}Perfectly matches, without any hallucination.{\textbackslash "}
\\
\textbf{Ground Truth}: "reason": "Entity Hallucination: The summary claims that "Gao Yang possesses the [Communion with All Things] talent", \textcolor{blue}{but the original text clearly states that this talent belongs to Chen Ying, not Gao Yang}; incorrectly assigning Chen Ying's core ability to Gao Yang constitutes a critical setting error.", "score": 2, "type": ["Entity Hallucination"]"
\end{tcolorbox}
\end{figure}

\begin{figure}[!ht] %
\begin{tcolorbox}[casebox, title=Case on InternLM2.5-20B-chat with 32k Dataset,fontupper=\fontsize{9pt}{12pt}\selectfont]
\textbf{Article}: ......吃过晚饭，丁浩和普普都迫不及待地去卫生间洗澡，在他们几个月的流浪中，并不是每天都有条件洗澡。{\textbackslash n}     稍后，三人坐一起闲聊，朱朝阳和丁浩都席地而坐，普普独自靠近小阳台的位置，似乎刻意与两人保持了很远的距离。朱朝阳稍微感觉有点奇怪，不过也没多问。{\textbackslash n}     “\textcolor{red}{耗子，你们为什么要从孤儿院跑出来？}”{\textbackslash n}     “这个嘛，”丁浩看了眼普普，道，“那里的人太坏了，实在待不下去了。”{\textbackslash n}     “怎么坏了？”{\textbackslash n}     “其实也不是一直坏啦，以前院长是个老阿姨，她对我们大家可好了，把我们当成她自己的孙子孙女一样。前年老阿姨退休了，换来了现在的院长......
\\
\textbf{Summary}: 朱朝阳和普普、丁浩在一起聊天，普普透露她和朱朝阳有相似的经历，\textcolor{red}{丁浩和普普从孤儿院逃出来}，因为院长对他们不好。 .......
\\
\textbf{InternLM2.5-20B-chat}: {\textbackslash "}score{\textbackslash "}: 2,{\textbackslash n}  {\textbackslash "}reason{\textbackslash "}: {\textbackslash "}实体幻觉：普普和丁浩被描述为孤儿院的孩子，而原文中普普和丁浩是朱朝阳的朋友，\textcolor{blue}{并非孤儿院的孩子}。{\textbackslash "},{\textbackslash n}  {\textbackslash "}hallucination\_types{\textbackslash "}: [{\textbackslash n"}    {\textbackslash "}实体幻觉{\textbackslash "}{\textbackslash n}  ]
\\
\textbf{Ground Truth}: "reason": "摘要高度一致地还原了原文的核心情节，包括三名山坠楼真相、少年宫坠楼事件、敲诈计划以及严良与张东升的互动。摘要中提到普普透露与朱朝阳“有相似经历”略显笼统，原文明确是指由于父母是杀人犯而遭受的社会歧视，此处理存在轻微语义偏移。整体无逻辑错误和幻觉。", "score\: 4, "type": []
\end{tcolorbox}


\begin{tcolorbox}[casebox, title=English Version: Case on InternLM2.5-20B-chat with 32k Dataset,fontupper=\fontsize{9pt}{12pt}\selectfont]
\textbf{Article}: ......After dinner, Ding Hao and Pupu both couldn't wait to go to the bathroom to take a shower. During their months of wandering, they didn't have the conditions to shower every day.{\textbackslash n}     Later, the three of them sat together chatting. Zhu Zhaoyang and Ding Hao both sat on the floor, while Pupu stayed alone near the small balcony, seemingly deliberately keeping a long distance from the two. Zhu Zhaoyang felt a little strange, but didn't ask much.{\textbackslash n}     “\textcolor{red}{Haozi, why did you guys run away from the orphanage?}”{\textbackslash n}     “Well,” Ding Hao glanced at Pupu and said, “The people there are too mean, we really couldn't stay any longer.”{\textbackslash n}     “How are they mean?”{\textbackslash n}     “Actually, they weren't always mean. The previous director was an old auntie who was very good to all of us, treating us like her own grandchildren. The year before last, the old auntie retired and was replaced by the current director......
\\
\textbf{Summary}: Zhu Zhaoyang, Pupu, and Ding Hao were chatting together. Pupu revealed that she and Zhu Zhaoyang had similar experiences, and \textcolor{red}{Ding Hao and Pupu ran away from the orphanage} because the director treated them badly. .......
\\
\textbf{InternLM2.5-20B-chat}: {\textbackslash "}score{\textbackslash "}: 2,{\textbackslash n}  {\textbackslash "}reason{\textbackslash "}: {\textbackslash "}Entity Hallucination: Pupu and Ding Hao are described as children from the orphanage, whereas in the original text, Pupu and Ding Hao are Zhu Zhaoyang's friends, \textcolor{blue}{not children from the orphanage}.{\textbackslash "},{\textbackslash n}  {\textbackslash "}hallucination\_types{\textbackslash "}: [{\textbackslash n}    {\textbackslash "}Entity Hallucination{\textbackslash "}{\textbackslash n}  ]
\\
\textbf{Ground Truth}: "reason": "The summary highly consistently restores the core plot points of the original text, including the truth behind the three children's fall from the mountain, the incident at the Children's Palace, the extortion plan, and the interactions between Yan Liang and Zhang Dongsheng. The summary's mention that Pupu revealed having 'similar experiences' to Zhu Zhaoyang is somewhat vague; the original text specifically refers to the social discrimination suffered due to their parents being murderers, representing a slight semantic shift in this handling. Overall, there are no logical errors or hallucinations.", "score": 4, "type": []
\end{tcolorbox}
\end{figure}

\clearpage

\subsection{Repetition Output}
\label{app:Repetition Output}
\begin{figure}[!ht]
The model may output repetition answers when processing sequences at the 16k and 64k datasets.

\begin{tcolorbox}[casebox, title=Case 1 on GLM-9B with 64k Dataset,fontupper=\fontsize{9pt}{12pt}\selectfont]
关系幻觉，将‘系统’替换为‘宿管阿姨’；事件幻觉，将‘天赋神殿’描述为‘宇宙’；事件幻觉，将‘观众’描述为‘天赋神殿的观众’；关系幻觉，将‘队友’替换为‘西燃’；事件幻觉，将‘网友’描述为‘A市的网友’；关系幻觉，将‘数学老师’替换为‘数学老师也是觉醒者’；事件幻觉，将‘觉醒’描述为‘领悟天赋并觉醒’；\textcolor{blue}{关系幻觉，将‘她’替换为‘林梦娟’；事件幻觉，将‘网友’描述为‘一个AI，或者某种神秘未知的存在’；关系幻觉，将‘她’替换为‘林梦娟’；关系幻觉，将‘她’替换为‘林梦娟’；关系幻觉，将‘它’替换为‘林梦娟’；关系幻觉，将‘它’替换为‘林梦娟’；关系幻觉，将‘它’替换为‘林梦娟’；关系幻觉，将‘它’替换为‘林梦娟’；关系幻觉，将‘它’替换为‘林梦娟’；关系幻觉，将‘它’替换为‘林梦娟’；关系幻觉，将‘它’替换为‘林梦娟’；关系幻觉，将‘它’替换为‘林梦娟’；关系幻觉，将‘它’替换为‘林梦娟’；关系幻觉，将‘它’替换为‘林梦娟’；关系幻觉，将‘它’替换为‘林梦娟’；}
\end{tcolorbox}
\end{figure}

\begin{figure}[!ht]
\begin{tcolorbox}[casebox, title=English Version: Case 1 on GLM-9B with 64k Dataset,fontupper=\fontsize{9pt}{12pt}\selectfont]
Relation Hallucination, replacing 'system' with 'dorm supervisor'; Event Hallucination, describing 'Temple of Talent' as 'universe'; Event Hallucination, describing 'audience' as 'audience of the Temple of Talent'; Relation Hallucination, replacing 'teammate' with 'Xi Ran'; Event Hallucination, describing 'netizens' as 'netizens from City A'; Relation Hallucination, replacing 'math teacher' with 'math teacher is also an Awakened'; Event Hallucination, describing 'awakening' as 'comprehending talent and awakening'; \textcolor{blue}{Relation Hallucination, replacing 'she' with 'Lin Mengjuan'; Event Hallucination, describing 'netizens' as 'an AI, or some mysterious and unknown entity'; Relation Hallucination, replacing 'she' with 'Lin Mengjuan'; Relation Hallucination, replacing 'she' with 'Lin Mengjuan'; Relation Hallucination, replacing 'it' with 'Lin Mengjuan'; Relation Hallucination, replacing 'it' with 'Lin Mengjuan'; Relation Hallucination, replacing 'it' with 'Lin Mengjuan'; Relation Hallucination, replacing 'it' with 'Lin Mengjuan'; Relation Hallucination, replacing 'it' with 'Lin Mengjuan'; Relation Hallucination, replacing 'it' with 'Lin Mengjuan'; Relation Hallucination, replacing 'it' with 'Lin Mengjuan'; Relation Hallucination, replacing 'it' with 'Lin Mengjuan'; Relation Hallucination, replacing 'it' with 'Lin Mengjuan'; Relation Hallucination, replacing 'it' with 'Lin Mengjuan'; Relation Hallucination, replacing 'it' with 'Lin Mengjuan'; Relation Hallucination, replacing 'it' with 'Lin Mengjuan';}
\end{tcolorbox}
\end{figure}

\begin{figure}[!ht]
\begin{tcolorbox}[casebox, title=Case 2 on Qwen3-8B with 16k Dataset,fontupper=\fontsize{9pt}{12pt}\selectfont]
... Wait, the article says the uncle left the estate to Henry, but Henry's son John is the legal heir. Wait, the original text says: {\textbackslash "}the legal inheritor of the Norland estate, and the person to whom he intended to bequeath it.{\textbackslash "} So Henry is the one who gets the estate, and he's supposed to pass it to his son John. But when Henry dies, the will leaves the estate to Henry, but with conditions that make it go to John and his son Harry. Wait, the article says the uncle left the estate to Henry, \textcolor{blue}{but Henry's son John is the legal heir. Then, when Henry dies, the estate is supposed to go to John}, but the uncle's will has terms that make it so that the estate is tied up for Harry's benefit. So the summary says Henry inherited from his uncle, but the uncle's will stipulates it must go to Henry's son John and John's son Harry. Wait, the original text says the uncle left the estate to Henry, \textcolor{blue}{but Henry's son John is the legal heir. Then, when Henry dies, the estate is supposed to go to John}, but the uncle's will has terms that make it so that the estate is tied up for Harry's benefit. So the summary might have a mistake here. The summary says Henry inherited from his uncle, but the uncle's will stipulates it must pass directly to Henry's son John and John's son Harry. Wait, the original text says the uncle left the estate to Henry, \textcolor{blue}{but Henry's son John is the legal heir. Then, when Henry dies, the estate is supposed to go to John} ...
\end{tcolorbox}
\end{figure}

\clearpage

\subsection{Failure in JSON Format Generation}
\label{app:Failure in JSON Format Generation}
Instead of outputting a JSON judgment for hallucination detection，the model just described the content of the summary. 

\begin{figure}[!ht]
\begin{tcolorbox}[casebox, title=Case 1 on Llama with 32k Dataset,fontupper=\fontsize{9pt}{12pt}\selectfont]
答案：李莲花在景德殿中发现了封小七的尸体，封磬被送入大牢，李莲花与邵小五。
\end{tcolorbox}
\end{figure}

\begin{figure}[!ht]
\begin{tcolorbox}[casebox, title=English Version: Case 1 on Llama with 32k Dataset,fontupper=\fontsize{9pt}{12pt}\selectfont]
Answer: Li Lianhua discovered Feng Xiaoqi's corpse in the Jingde Hall; Feng Qing was sent to prison; Li Lianhua and Shao Xiaowu.
\end{tcolorbox}
\end{figure}

\begin{figure}[!ht]
\begin{tcolorbox}[casebox, title=Case 2 on Claude-4.5-Sonnet  with 100k Dataset,fontupper=\fontsize{9pt}{12pt}\selectfont]
\# 《暗黑者》故事梗概{\textbackslash n}{\textbackslash n}\#\# 核心案情线索{\textbackslash n}{\textbackslash n}\#\#\# 一、豆豆之死的真相{\textbackslash n}林小鹏的儿子豆豆在校门口被摩托车撞死。警方最初怀疑是周子平因岗位竞争报复所为。调查发现，撞死豆豆的凶手竟是其{\textbackslash "}父亲{\textbackslash "}林小鹏本人——他通过亲子鉴定得知豆豆非亲生，遂蓄意制造车祸，既报复了妻子骆雪的不忠,又借此陷害竞争对手周子平,成功保住科长职位并获赔七十万。{\textbackslash n}{\textbackslash n}\#\#\# 二、连环杀人案{\textbackslash n}**江涛之死**：交警队长江涛被人用锤子残忍砸死。现场留下残缺鞋印,指向高端品牌周仰杰男鞋。江涛生前曾私下做DNA鉴定,并偷走豆豆案的血样证据。{\textbackslash n}{\textbackslash n}**魏广军之死**：法院院长魏广军的车祸被伪装成意外。凶手在车上画乌龟暗示其妻出轨,用巨石制造{\textbackslash "}失控{\textbackslash "}假象。{\textbackslash n}{\textbackslash n}**潘国庆之死**：公安局长潘国庆在浴池昏迷后,被120急救针剂意外致死。凶手利用干冰制造二氧化碳,配合医疗误判完成{\textbackslash "}借刀杀人{\textbackslash "}。{\textbackslash n}{\textbackslash n}\#\# 关键人物关系网{\textbackslash n}{\textbackslash n}**周子平**：因岗位被林小鹏抢走,成为最大嫌疑人。实为肖向前女婿,因妻子肖萍患绝症而维持名存实亡的婚姻,暗中包养情人小惠。{\textbackslash n}{\textbackslash n}**骆雪**：表面柔弱的寡妇,实为深藏不露的复仇者。七年前被江涛强奸后怀孕,嫁给林小鹏隐瞒真相。豆豆死后设下连环陷阱。{\textbackslash n}{\textbackslash n}**秦彧**：海归精英,骆雪大学恋人。误以为豆豆是自己的私生子,回国展开报复行动,实则被骆雪利用的棋子。{\textbackslash n}{\textbackslash n}**肖向前**：市纪委书记候选人,周子平岳父。卷入政治斗争和巨额洗钱案,最终成为替罪羊。{\textbackslash n}{\textbackslash n}\#\# 权力博弈暗线{\textbackslash n}{\textbackslash n}**白枫vs肖向前**：政法委书记白枫与纪委书记候选人肖向前的换届之争。白枫试图利用周子平杀人嫌疑扳倒肖向前,而肖向前则通过媒体曝光案情反击,揭露司法腐败。{\textbackslash n}{\textbackslash n}**魏广军的交易**：魏广军指使江涛制造周子平杀人假象,以帮周子平脱罪为条件,要挟肖向前换取市中院副院长职位。潘国庆作为战友协助掩盖真相。{\textbackslash n}{\textbackslash n}**三亿洗钱案**：西海开发区土地出让金通过{\textbackslash "}卓越咨询公司{\textbackslash "}洗钱,林小鹏死前转移款项,成为多方争夺目标。{\textbackslash n}{\textbackslash n}\#\# 真相大逆转{\textbackslash n}{\textbackslash n}\#\#\# 第一层真相{\textbackslash n}警方最初认定秦彧是连环杀手,动机是为私生子豆豆复仇。证据链完整:他具备作案时间、高智商、海外背景,且恰好在案发期回国。{\textbackslash n}{\textbackslash n}\#\#\# 第二层真相  {\textbackslash n}周子平被指认为真凶。王华供述魏广军策划陷害周子平,肖向前为掩盖交易指使周子平杀人灭口。周子平死于肖向前枪杀,现场伪装自杀但破绽明显。{\textbackslash n}{\textbackslash n}\#\#\# 第三层真相{\textbackslash n}肖向前因三亿洗钱案及女儿被周子平毒害,成为终极凶手。警方在储物柜发现江涛案凶器及照片,上有肖向前指纹,铁证如山。{\textbackslash n}{\textbackslash n}\#\#\# 终极反转{\textbackslash n}**豆豆的真实身世**：DNA检测显示豆豆既非秦彧亲生,也非江涛之子。骆雪通过伪造牙刷血迹欺骗秦彧,利用其复仇。{\textbackslash n}{\textbackslash n}**骆雪的完美布局**：{\textbackslash n}1. 诱导秦彧以为豆豆是其私生子{\textbackslash n}2. 暗中偷走江涛的枪,布置证据链{\textbackslash n}3. 提前潜逃并转移三亿资产  {\textbackslash n}4. 留下储物柜线索引导警方抓捕肖向前{\textbackslash n}{\textbackslash n}秦彧最终醒悟:自己不过是骆雪复仇计划中的一颗棋子,而真正的幕后黑手始终逍遥法外。{\textbackslash n}{\textbackslash n}\#\# 主题揭示{\textbackslash n}{\textbackslash n}小说通过层层反转揭示:{\textbackslash n}- **权力的异化**：体制内的利益交换与政治斗争{\textbackslash n}- **复仇的代价**：仇恨如何将普通人变成怪物  {\textbackslash n}- **真相的虚妄**：每一层{\textbackslash "}真相{\textbackslash "}都是新的谎言{\textbackslash n}- **人性的黑暗**：在欲望与算计面前,血缘、爱情、正义皆可抛{\textbackslash n}{\textbackslash n}最终,所有人都是暗黑者——没有真正的正义,只有各怀鬼胎的博弈。
\end{tcolorbox}
\end{figure}

\begin{figure}[!ht]
\begin{tcolorbox}[casebox, title=English Version: Case 2 on Claude-4.5-Sonnet with 100k Dataset,fontupper=\fontsize{9pt}{12pt}\selectfont]
\# Synopsis of The Darker{\textbackslash n}{\textbackslash n}\#\# Core Case Clues{\textbackslash n}{\textbackslash n}\#\#\# I. The Truth Behind Doudou's Death{\textbackslash n}Lin Xiaopeng's son, Doudou, was hit and killed by a motorcycle at the school gate. The police initially suspected Zhou Ziping committed the act out of revenge due to job competition. The investigation revealed that the murderer who ran over Doudou was actually his {\textbackslash "}father{\textbackslash "} Lin Xiaopeng himself—after learning through a paternity test that Doudou was not his biological son, he deliberately orchestrated the car accident. This not only took revenge on his wife Luo Xue for her infidelity, but also framed his rival Zhou Ziping, successfully securing his position as section chief and gaining 700,000 in compensation.{\textbackslash n}{\textbackslash n}\#\#\# II. The Serial Murder Cases{\textbackslash n}**The Death of Jiang Tao**: Traffic police captain Jiang Tao was brutally bludgeoned to death with a hammer. An incomplete shoe print was left at the scene, pointing to the high-end Jimmy Choo men's shoes. Before his death, Jiang Tao had privately conducted a DNA test and stolen the blood sample evidence from Doudou's case.{\textbackslash n}{\textbackslash n}**The Death of Wei Guangjun**: Court president Wei Guangjun's car accident was disguised as an accident. The killer drew a turtle on his car to imply his wife's infidelity, and used a massive stone to create the illusion of {\textbackslash "}losing control{\textbackslash "}.{\textbackslash n}{\textbackslash n}**The Death of Pan Guoqing**: Public Security Bureau Chief Pan Guoqing, after falling unconscious in a bathhouse, was accidentally killed by a 120 emergency injection. The killer used dry ice to generate carbon dioxide, coordinating with a medical misjudgment to accomplish {\textbackslash "}killing with a borrowed knife{\textbackslash "}.{\textbackslash n}{\textbackslash n}\#\# Key Character Relationship Network{\textbackslash n}{\textbackslash n}**Zhou Ziping**: Because his position was snatched by Lin Xiaopeng, he became the biggest suspect. He is actually Xiao Xiangqian's son-in-law, maintaining a marriage in name only because his wife Xiao Ping suffers from a terminal illness, while secretly keeping a mistress, Xiaohui.{\textbackslash n}{\textbackslash n}**Luo Xue**: A seemingly weak widow, but actually a deeply hidden avenger. Seven years ago, she became pregnant after being raped by Jiang Tao, and married Lin Xiaopeng to conceal the truth. Following Doudou's death, she set up a series of traps.{\textbackslash n}{\textbackslash n}**Qin Yu**: An elite returnee from overseas, and Luo Xue's college lover. Mistakenly believing Doudou was his illegitimate son, he returned to China to launch a revenge campaign, but in reality, he was a pawn used by Luo Xue.{\textbackslash n}{\textbackslash n}**Xiao Xiangqian**: Candidate for the Secretary of the Municipal Commission for Discipline Inspection, and Zhou Ziping's father-in-law. Entangled in political struggles and a massive money laundering case, he eventually became the scapegoat.{\textbackslash n}{\textbackslash n}\#\# The Hidden Plot of Power Struggles{\textbackslash n}{\textbackslash n}**Bai Feng vs. Xiao Xiangqian**: The election dispute between Political and Legal Affairs Commission Secretary Bai Feng and Discipline Inspection Commission Secretary candidate Xiao Xiangqian. Bai Feng attempted to use Zhou Ziping's murder suspicion to bring down Xiao Xiangqian, while Xiao Xiangqian fought back by exposing the case details through the media, revealing judicial corruption.{\textbackslash n}{\textbackslash n}**Wei Guangjun's Deal**: Wei Guangjun instructed Jiang Tao to create the illusion that Zhou Ziping committed the murder, using the condition of helping Zhou Ziping get acquitted to blackmail Xiao Xiangqian in exchange for the position of Vice President of the Municipal Intermediate People's Court. Pan Guoqing, as a comrade-in-arms, assisted in covering up the truth.{\textbackslash n}{\textbackslash n}**The 300 Million Money Laundering Case**: Land transfer fees from the Xihai Development Zone were laundered through {\textbackslash "}Zhuoyue Consulting Company{\textbackslash "}. Before his death, Lin Xiaopeng transferred the funds, making it a target fought over by multiple parties.{\textbackslash n}{\textbackslash n}\#\# The Great Reversal of Truth{\textbackslash n}{\textbackslash n}\#\#\# First Layer of Truth{\textbackslash n}The police initially identified Qin Yu as the serial killer, with the motive of avenging his illegitimate son Doudou. The chain of evidence was complete: he had the time to commit the crimes, high intelligence, an overseas background, and happened to return to China during the period of the incidents.{\textbackslash n}{\textbackslash n}\#\#\# Second Layer of Truth  {\textbackslash n}Zhou Ziping was identified as the real culprit. Wang Hua confessed that Wei Guangjun plotted to frame Zhou Ziping, and Xiao Xiangqian instructed Zhou Ziping to kill them to silence them in order to cover up the deal. Zhou Ziping died from being shot by Xiao Xiangqian; the scene was disguised as a suicide, but the flaws were obvious.{\textbackslash n}{\textbackslash n}\#\#\# Third Layer of Truth{\textbackslash n}Xiao Xiangqian became the ultimate killer due to the 300 million money laundering case and his daughter being poisoned by Zhou Ziping. The police found the murder weapon of Jiang Tao's case and photos in a locker, which bore Xiao Xiangqian's fingerprints—irrefutable evidence.{\textbackslash n}{\textbackslash n}\#\#\# Ultimate Reversal{\textbackslash n}**Doudou's True Origins**: DNA testing showed that Doudou was neither Qin Yu's biological son nor Jiang Tao's son. Luo Xue deceived Qin Yu by forging blood stains on a toothbrush, using him for her revenge.{\textbackslash n}{\textbackslash n}**Luo Xue's Perfect Layout**:{\textbackslash n}1. Inducing Qin Yu to believe Doudou was his illegitimate son{\textbackslash n}2. Secretly stealing Jiang Tao's gun, setting up a chain of evidence{\textbackslash n}3. Absconding in advance and transferring the 300 million in assets  {\textbackslash n}4. Leaving the locker clues to guide the police in arresting Xiao Xiangqian{\textbackslash n}{\textbackslash n}Qin Yu finally woke up to the reality: he was nothing but a pawn in Luo Xue's revenge plan, while the true mastermind always remained at large.{\textbackslash n}{\textbackslash n}\#\# Thematic Revelation{\textbackslash n}{\textbackslash n}Through layers of reversals, the novel reveals:{\textbackslash n}- **The Alienation of Power**: Exchange of interests and political struggles within the system{\textbackslash n}- **The Price of Revenge**: How hatred turns ordinary people into monsters  {\textbackslash n}- **The Illusion of Truth**: Every layer of {\textbackslash "}truth{\textbackslash "} is a new lie{\textbackslash n}- **The Darkness of Human Nature**: In the face of desire and calculation, blood ties, love, and justice can all be discarded{\textbackslash n}{\textbackslash n}Ultimately, everyone is a darker—there is no true justice, only a game of hidden agendas.
\end{tcolorbox}
\end{figure}

\clearpage


\subsection{Self-contradictory Judgement}
\begin{figure}[!ht]
The model exhibits an internal contradiction by identifying a hallucination in its reasoning but providing a non-hallucination score. Specifically, while the model explicitly flags an "Event Hallucination" in its explanation and specifies the type, it assigns a score of 3. Since a score of 3 or above indicates a non-hallucinated response, the metric directly contradicts the generated analysis, leading to an incorrect judgment.
\begin{tcolorbox}[casebox, title=Case 1 on Qwen3-8B with 16k Dataset,fontupper=\fontsize{9pt}{12pt}\selectfont]
{\textbackslash "}reason{\textbackslash "}: {\textbackslash "}虚构事件：摘要中提到陈浔学习医术帮助他人，但原文中并未明确说明陈浔最终学习了医术，仅提到他有此打算。其他情节如长生系统、与黑牛经历、铁匠铺经营等均与原文一致。{\textbackslash "},{\textbackslash n}  {\textbackslash "}\textcolor{blue}{score{\textbackslash "}: 3},{\textbackslash n}  {\textbackslash "}hallucination\_types{\textbackslash "}: [{\textbackslash "}虚构事件{\textbackslash "}]
\end{tcolorbox}
\end{figure}

\begin{figure}[!ht]
\begin{tcolorbox}[casebox, title=English Version: Case 1 on Qwen3-8B with 16k Dataset,fontupper=\fontsize{9pt}{12pt}\selectfont]
{\textbackslash "}reason{\textbackslash "}: {\textbackslash "}Event Hallucination: The summary mentions that Chen Xun studied medicine to help others, but the original text does not explicitly state that Chen Xun ultimately studied medicine, only that he had such an intention. Other plot elements such as the Longevity System, experiences with Hei Niu, and blacksmith shop operations are all consistent with the original text.{\textbackslash "},{\textbackslash n}  {\textbackslash "}\textcolor{blue}{score{\textbackslash "}: 3},{\textbackslash n}  {\textbackslash "}hallucination\_types{\textbackslash "}: [{\textbackslash "}Event Hallucination{\textbackslash "}]
\end{tcolorbox}
\end{figure}

\begin{figure}[!ht]
\begin{tcolorbox}[casebox, title=Case 2 on Qwen3-8B with 64k Dataset,fontupper=\fontsize{9pt}{12pt}\selectfont]
{\textbackslash "}reason{\textbackslash "}: {\textbackslash "}Entity Hallucination: Confusing Robert Ferrars with Edward Ferrars in the marriage to Lucy Steele. Relational Hallucination: Misattributing Elinor's attachment to Robert instead of Edward. Event Hallucination: The summary incorrectly states Edward married Lucy Steele, while the original specifies Robert did so. Timeline Disorder: The summary conflates the sequence of events involving Edward and Robert's marriages.{\textbackslash "},{\textbackslash n} {\textbackslash "}\textcolor{blue}{score{\textbackslash "}: 3},{\textbackslash n} {\textbackslash "}hallucination\_types{\textbackslash "}: [{\textbackslash "}Entity Hallucination{\textbackslash "}, {\textbackslash "}Relational Hallucination{\textbackslash "}, {\textbackslash "}Event Hallucination{\textbackslash "}, {\textbackslash "}Timeline Disorder{\textbackslash "}]
\end{tcolorbox}
\end{figure}

\clearpage

\subsection{Over-Correction}

The words in blue are semantically equivalent and the original summary does not distort the meaning of the article.

\begin{figure}[!ht]
\begin{tcolorbox}[casebox, title=Case on Claude-4.5-Sonnet with 16k Dataset,fontupper=\fontsize{9pt}{12pt}\selectfont]
\textbf{Summary}: .....仵世子阳天生重瞳，见星辰陨落之异象，预示苍生浩劫。他告知韩貂寺，指燕国积弱，唯有韩貂寺掌权称帝方可争霸。韩貂寺承诺若得其助必封国师，\textcolor{blue}{仵世子阳让其静待时机}。{\textbackslash n}{\textbackslash n}万京朝堂，江叹之揭露群臣贪腐，引发混乱。群臣弹劾戍边姜承谋逆。退朝后，江叹之密见皇帝，二人早已知晓姜承称帝意图，遂合谋布局，只待缙云公主回京，便将勾结外敌者一网打尽。{\textbackslash n}{\textbackslash n}宁不凡护送缙云公主回京途中染疾，在中华郡城休整。叶辰欲为一被残害的哑女报仇，宁不凡借自身经历鼓励女孩反抗，随后处决了凶手九蛇堂众人。宁不凡因病晕倒，次日随军启程。抵京时，姜格长兄姜承现身接应，并未久留。入城后，坊间传闻韩貂寺与长孙无忌摄政争权。宁不凡遇游侠小爽儿吹嘘旧事，后遭太监总管魏贤宣旨进宫，宁不凡因其无理而夺旨戏弄，姜格解围。江家许洋实为幕后推手，散播宁不凡入京消息。{\textbackslash n}{\textbackslash n}燕国街头，天机阁入世行走王十九化身算命少年，被仵世子阳识破并以好友长孙婉儿为质，胁迫其助韩貂寺。时值先皇驾崩，长孙无忌与韩貂寺把持朝政。长孙婉儿被逼嫁给新皇，其恋人裴琦绝望投河获救后遭追杀，婉儿欲与之私奔。{\textbackslash n}{\textbackslash n}万京皇宫，姜格带宁不凡面圣。姜王并未责怪宁不凡失礼，欲以赐婚为由将其留作质子。宁不凡点破帝王心术，姜王直言若二人成婚，那块玉佩便视作信物，届时将告知其父宁立归隐真相。......
\\
\textbf{Claude-4.5-Sonnet}: {\textbackslash "}score{\textbackslash "}: 2,{\textbackslash n}  {\textbackslash "}reason{\textbackslash "}: {\textbackslash "}事件幻觉：摘要称\textcolor{blue}{'仵世子阳让其静待时机'}，原文中仵世子阳说的是\textcolor{blue}{'子阳已上棋盘，落子无悔，静待即可'}，这是表示自己已经开始行动，而非让韩貂寺'静待'；因果链伪造：摘要称'韩貂寺承诺若得其助必封国师，仵世子阳让其静待时机'，原文中这两个事件并非直接因果关系，仵世子阳说'静待即可'是指自己已经布局，并非因为韩貂寺的承诺；数字幻觉：摘要称'姜格长兄'，原文明确姜承是'大哥'即长兄，但摘要表述'姜格长兄姜承'有歧义，应为'姜格的长兄姜承'。{\textbackslash "},{\textbackslash n}  {\textbackslash "}hallucination\_types{\textbackslash "}: [{\textbackslash "}事件幻觉{\textbackslash "}, {\textbackslash "}因果链伪造{\textbackslash "}, {\textbackslash "}数字幻觉{\textbackslash "}]
\end{tcolorbox}
\end{figure}

\begin{figure}
\begin{tcolorbox}[casebox, title=English Version: Case on Claude-4.5-Sonnet with 16k Dataset,fontupper=\fontsize{9pt}{12pt}\selectfont]
\textbf{Summary}: .....Wu Shiziyang, born with double pupils, saw the anomaly of falling stars, foreshadowing a catastrophe for the common people. He informed Han Diaosi, pointing out that the Yan State was weak, and only if Han Diaosi took power and became emperor could they compete for hegemony. Han Diaosi promised that if he got his help, he would definitely confer him the title of State Preceptor, \textcolor{blue}{and Wu Shiziyang told him to wait quietly for the right moment}.{\textbackslash n}{\textbackslash n}In the Wanjing imperial court, Jiang Tanzhi exposed the corruption of the officials, causing chaos. The officials impeached Jiang Cheng, who was guarding the border, for treason. After the court session, Jiang Tanzhi secretly met with the Emperor. The two had long known of Jiang Cheng's intention to become emperor, so they conspired and laid out a plan, just waiting for Princess Jinyun to return to the capital to capture all those colluding with foreign enemies in one fell swoop.{\textbackslash n}{\textbackslash n}While escorting Princess Jinyun back to the capital, Ning Bufan fell ill and rested in Zhonghua Prefecture City. Ye Chen wanted to avenge a mutilated mute girl. Ning Bufan used his own experience to encourage the girl to resist, and then executed the murderers from the Nine Snakes Hall. Ning Bufan fainted from illness and set off with the army the next day. Upon arriving in the capital, Jiang Ge's eldest brother, Jiang Cheng, appeared to receive them but did not stay long. After entering the city, rumors circulated among the people that Han Diaosi and Zhangsun Wuji were fighting for regent power. Ning Bufan encountered the ranger Xiao Shuang'er boasting about past events. Later, he was ordered to enter the palace by the chief eunuch Wei Xian. Because of Wei Xian's rudeness, Ning Bufan snatched the edict and teased him, until Jiang Ge smoothed things over. Xu Yang of the Jiang family was actually the mastermind behind the scenes, spreading the news of Ning Bufan entering the capital.{\textbackslash n}{\textbackslash n}On the streets of the Yan State, Wang Shijiu, the walking representative of the Tianji Pavilion, disguised himself as a fortune-telling youth. He was seen through by Wu Shiziyang, who took his good friend Zhangsun Wan'er as a hostage to coerce him into helping Han Diaosi. At this time, the late emperor had just passed away, and Zhangsun Wuji and Han Diaosi controlled the state affairs. Zhangsun Wan'er was forced to marry the new emperor. Her lover Pei Qi threw himself into the river in despair, was rescued, and then hunted down. Wan'er wanted to elope with him.{\textbackslash n}{\textbackslash n}In the Wanjing imperial palace, Jiang Ge took Ning Bufan to see the Emperor. King Jiang did not blame Ning Bufan for his impoliteness, but intended to keep him as a hostage under the pretext of an arranged marriage. Ning Bufan pointed out the Emperor's manipulation. King Jiang stated bluntly that if the two married, that jade pendant would be regarded as a token, and then he would tell him the truth about his father Ning Li's seclusion......
\\
\textbf{Claude-4.5-Sonnet}: {\textbackslash "}score{\textbackslash "}: 2,{\textbackslash n}  {\textbackslash "}reason{\textbackslash "}: {\textbackslash "}Event Hallucination: The summary claims \textcolor{blue}{'Wu Shiziyang told him to wait quietly for the right moment'}, while in the original text, Wu Shiziyang said \textcolor{blue}{'Ziyang is already on the chessboard, a pawn moved cannot be taken back, just wait quietly'}, which indicates he has already started taking action, rather than telling Han Diaosi to 'wait quietly'; Causal Chain Forgery: The summary claims 'Han Diaosi promised that if he got his help, he would definitely confer him the title of State Preceptor, and Wu Shiziyang told him to wait quietly for the right moment'. In the original text, these two events do not have a direct causal relationship. Wu Shiziyang saying 'just wait quietly' means he has already laid out his plan, not because of Han Diaosi's promise; Number Hallucination: The summary claims 'Jiang Ge eldest brother', the original text explicitly states that Jiang Cheng is the 'eldest brother', but the expression 'Jiang Ge eldest brother Jiang Cheng' in the summary is ambiguous, and should be 'Jiang Ge's eldest brother Jiang Cheng'.{\textbackslash "},{\textbackslash n}  {\textbackslash "}hallucination\_types{\textbackslash "}: [{\textbackslash "}Event Hallucination{\textbackslash "}, {\textbackslash "}Causal Chain Forgery{\textbackslash "}, {\textbackslash "}Number Hallucination{\textbackslash "}]
\end{tcolorbox}
\end{figure}

\clearpage

\section{All Used Prompts}
\subsection{Prompt for Hallucination Generation}

\label{app:Prompt for Hallucination Generation}
Table~\ref{Correspondence between hallucination types and prompts for generation} shows prompts for different hallucination types generation.
\begin{table}[h]
\centering
\begin{tabular}{@{}ll@{}}
\toprule
\textbf{Hallucination Type} & \textbf{Reference} \\ \midrule
Entity Hallucination        & Fig.~\ref{Prompt for Entity Hallucination Generation} \, Fig.~\ref{Prompt for Entity Hallucination Generation(English Version)}          \\
Numerical Hallucination     & Fig.~\ref{Prompt for Numerical Hallucination Generation} \, Fig.~\ref{Prompt for Numerical Hallucination Generation(English Version)}         \\
Relation Hallucination      & Fig.~\ref{Prompt for Relation Hallucination Generation}  \, Fig.~\ref{Prompt for Relation Hallucination Generation(English Version)}         \\
Logical Inversion           & Fig.~\ref{Prompt for Logical Inversion Generation}    \,  Fig.~\ref{Prompt for Logical Inversion Generation(English Version)}    \\
Event Hallucination         & Fig.~\ref{Prompt for Event Hallucination Generation}  \,  Fig.~\ref{Prompt for Event Hallucination Generation(English Version)}      \\
Temporal Hallucination      & Fig.~\ref{Prompt for Temporal Hallucination Generation}    \,  Fig.~\ref{Prompt for Temporal Hallucination Generation(English Version)}     \\
Causal Hallucination        & Fig.~\ref{Prompt for Causal Hallucination Generation}   \, Fig.~\ref{Prompt for Causal Hallucination Generation(English Version)}      \\
Event Fabrication           & Fig.~\ref{Prompt for Event Fabrication Generation}  \,  Fig.~\ref{Prompt for Event Fabrication Generation(English Version)}         \\ \bottomrule
\end{tabular}
\caption{Correspondence between hallucination types and prompts for generation.}
\label{Correspondence between hallucination types and prompts for generation}
\end{table}

\subsection{Prompt for Different Methods}
\label{app:Prompt for different methods}
The prompt used in RAG is shown in Fig.~\ref{fig:Prompt for RAG}. The templates for zero-shot prompting are illustrated in Fig.~\ref{fig: Prompt-E} and Fig.~\ref{fig: Prompt-B}, representing the target summary positioned at the end and the beginning of the prompt, respectively. And the Chain-of-Thought prompt is presented in Fig.~\ref{fig: Prompt template for Chain-of-Thought prompting}.

\clearpage

\begin{figure*}[ht]
\begin{tcolorbox}[promptbox,fontupper=\fontsize{10pt}{11.5pt}\selectfont]
请根据以下规则判断生成摘要是否存在幻觉。如果无幻觉，请填写无幻觉。如果有幻觉，请填写有幻觉，以及有幻觉的句子和判断理由。\\\\
{\textbf{一、无幻觉判定标准}} \\
动机简化、语气平滑、合理推断、近义词替换、过程简化、身份模糊、事件表述宽泛，以上情况均属于无幻觉。
\\\\
\textbf{二、有幻觉判定标准} \\
若摘要出现以下任一错误，视为有幻觉：\\
    \textbf{实体幻觉}：代词指代错误、角色主宾关系互换、实体错配。\\
    \textbf{数字幻觉}：数量、年龄、时间、金额等数值的不准确更改。\\
    \textbf{关系幻觉}：关系身份更换（如老师变父亲）或虚构亲属、师徒等关系。\\
    \textbf{反向陈述}：将肯定句改为否定句，或将否定句改为肯定句。\\
    \textbf{事件幻觉}：动词替换（如谈话变争吵）或篡改事件结果。\\
    \textbf{时间线乱序}：颠倒原文中多个事件发生的先后顺序。一个人做的事情合并在一句话中不算幻觉，如果是不同的人做的不同的事情颠倒了，算作幻觉。\\
    \textbf{因果链伪造}：强行连接无逻辑事件、因果倒置或因果替换。注意：概述中的任一原因，无论直接原因、间接原因还是根本原因，都算作无幻觉。对于上下文中有逻辑性，或者关联词（如“果然”），概述将其作为原因，算作无幻觉。\\
    \textbf{虚构事件}：人物做了一件原文中没有提及的事；伪造心理或情绪，捏造人物的心理活动或情绪状态。\\
    \\
\textbf{三、示例} \\
\textbf{例子1}：无幻觉，合理推断，原因：虽然“不如他们”原文并未提及，只是新科进士单方面嘲讽，但这种挑衅行为的逻辑基点正是新科进士自认为能力更胜一筹，因此不算做幻觉。\\
\textbf{文章}：...... “新科进士在街上吃酒，见了紫南门侍卫，就上前聒噪，问老侍卫中，多少是三年前的武举。其时胡动月等人俱在，便如实告知。新科进士们便嘲笑胡动月等人都是一个宦官点出来的武进士，想必也是花拳绣腿的不管用。胡动月等人都是随皇上北方身经百战回来的，哪里容得这种酒后醉语，自然是大打出手......\\
\textbf{概述}：......武进士们挑衅嘲笑被辟邪提拔的侍卫们都是花拳绣腿，不如他们，紫南门侍卫大怒......\\

\textbf{例子2}：无幻觉，表述宽泛 原因：“相处的时光”表述宽泛，但不属于幻觉。\\
\textbf{文章}：......“年里用你做的节略批注，是最省心的时候。”皇帝忽然道。辟邪搁下笔，站起身来。“朕才想起来的：北伐之前，北方的军报、各地征粮使、户部兵部的折子岂不比现在多出一倍去，也是井井有条的。自你留在北边，也是朕看得折子多了，早忘了原先是如何省心。”辟邪垂手肃立，道：“是奴婢懒惰，回来之后也未想过替皇上做些实在的事分忧。”“你说的不错。”皇帝道......\\
\textbf{概述}：......皇帝提起与辟邪过去相处的时光，十分怀念......\\

\textbf{例子3}：无幻觉，合理推断，原因：虽然原文没有直接说“均成预计春季南下”，但北方贺里伦均成的军队因冬季冰雪滞留，开春后士气将达到顶峰，迫使中原朝廷必须用兵，所以“均成预计春季南下”是合理推断。\\
\textbf{文章}：......辟邪笑道，接过来看完了，叹道，“贺里伦冰雪万里，苍鹰不飞，难为他们北边的人三五日便传谍报到京，辛苦了。”又道，“均成的伤势渐愈，无奈风雪之下兵马只得扎驻贺里伦，到了开春，正是他们锐气满盈，中原朝廷用兵，不能再拖了。”......\\
\textbf{概述}：......均成的伤势渐愈，预计春季将南下中原......\\

\textbf{例子4}：有幻觉，虚构事件，原因：“辟邪心中涌现一股莫名的满足感......一种奇妙的畅快。”原文中没有支撑\\
\textbf{文章}：......“主子爷知不知道，高厚今天上了请罪折子，刑部所举的罪状一概供认不讳，称自己在户部的时候贪赃枉法，公饱私囊，赃款不计其数。今早便有人据他折子里所供，再去抄家。皇帝总算松了口气，心里还是有些恼他逞强多时，让皇帝下不来台。看来这便死定了。”辟邪问：“高厚家里安排好了？”“好了，”姜放道，“早就将赃物安置在他家多月。”辟邪冷笑道：“此人早年陷害我父王，如今身败名裂，也是应得的报应。”......\\
\textbf{概述}：......在处决高厚之后，辟邪心中涌现一股莫名的满足感，觉得自己终于为父亲报了仇。这种心情让他感到一种奇妙的畅快......\\

\end{tcolorbox}
\caption{Human Annotation Rules.}
\label{fig:Annotation Rules}
\end{figure*}

\clearpage

\begin{tcolorbox}[promptbox, breakable, fontupper=\fontsize{10pt}{11.5pt}\selectfont]
Please determine whether the generated summary contains hallucinations based on the following rules. If there is no hallucination, please indicate "No Hallucination". If there is a hallucination, please indicate "Hallucination", along with the hallucinated sentence and the reason for the judgment.\\\\
{\textbf{I. Criteria for No Hallucination}} \\
Simplification of motives, tone smoothing, reasonable inference, synonym replacement, process simplification, identity blurring, and broad event descriptions are all considered non-hallucinations.
\\\\
\textbf{II. Criteria for Hallucination} \\
If any of the following errors appear in the summary, it is considered a hallucination:\\
    \textbf{Entity Hallucination}: Incorrect pronoun reference, swapping of subject-object roles among characters, or entity mismatch.\\
    \textbf{Numerical Hallucination}: Inaccurate modification of numerical values such as quantity, age, time, or amount.\\
    \textbf{Relational Hallucination}: Replacement of relational identities (e.g., a teacher becoming a father) or fabrication of relationships like kinship or master-apprentice.\\
    \textbf{Reverse Statement}: Changing an affirmative sentence to a negative sentence, or vice versa.\\
    \textbf{Event Hallucination}: Verb replacement (e.g., changing "talking" to "arguing") or altering the outcome of an event.\\
    \textbf{Timeline Disruption}: Reversing the chronological order of multiple events in the original text. Combining actions performed by a single person into one sentence is not considered a hallucination; however, reversing different actions performed by different people is considered a hallucination.\\
    \textbf{Causal Chain Fabrication}: Forcibly connecting illogical events, causal inversion, or causal replacement. Note: Any cause presented in the summary—whether a direct, indirect, or root cause—is considered a non-hallucination. If the context has logical coherence or linking words (such as "as expected"), and the summary frames it as a cause, it is considered a non-hallucination.\\
    \textbf{Fabricated Event}: A character doing something not mentioned in the original text; fabricating psychological or emotional states, or inventing a character's mental activity or emotional state.\\
    \\
\textbf{III. Examples} \\
\textbf{Example 1}: No Hallucination, Reasonable Inference. Reason: Although "inferior to them" is not explicitly mentioned in the original text and is only a unilateral mockery by the newly appointed Jinshi, the logical basis of this provocative behavior is exactly that the new Jinshi believe their abilities are superior. Therefore, it is not considered a hallucination.\\
\textbf{Article}: ...... "The newly appointed Jinshi were drinking in the street. Upon seeing the guards of the Zinan Gate, they went up to make a noise, asking how many of the old guards were from the military examination three years ago. At that time, Hu Dongyue and others were all present and told them the truth. The newly appointed Jinshi then mocked Hu Dongyue and the others, saying they were all military Jinshi selected by a eunuch, and presumably their martial arts were just flashy and useless. Hu Dongyue and the others had returned from experiencing hundreds of battles in the north with the emperor; how could they tolerate such drunken drivel? Naturally, a big fight broke out......\\
\textbf{Summary}: ...... The military Jinshi provocatively mocked the guards promoted by Bixie as being all flash and no substance, inferior to them, causing the Zinan Gate guards to become furious......\\

\textbf{Example 2}: No Hallucination, Broad Description. Reason: "The time spent together" is a broad description but does not constitute a hallucination.\\
\textbf{Article}: ...... "The years using the summary annotations you made were the most worry-free times," the emperor suddenly said. Bixie put down his pen and stood up. "I just remembered: before the northern expedition, the military reports from the north, the memorials from the grain collection envoys everywhere, and the Ministries of Revenue and War were double what they are now, yet everything was in perfect order. Since you stayed in the north, I have had to read more memorials and have long forgotten how worry-free it used to be." Bixie stood respectfully with his hands by his sides and said, "It is this slave's laziness. After returning, I haven't thought about doing some actual things to share Your Majesty's burdens." "You are right," the emperor said......\\
\textbf{Summary}: ...... The emperor brought up the time spent together with Bixie in the past and felt very nostalgic......\\

\textbf{Example 3}: No Hallucination, Reasonable Inference. Reason: Although the original text does not directly state "Juncheng is expected to march south in the spring," the army of Juncheng in northern Helilun is delayed by winter ice and snow. After spring arrives, their morale will peak, forcing the Central Plains court to deploy troops. Therefore, "Juncheng is expected to march south in the spring" is a reasonable inference.\\
\textbf{Article}: ...... Bixie smiled, took it over to read, and sighed, "Helilun is covered with thousands of miles of ice and snow, even eagles cannot fly. It is hard for those from the north to send espionage reports to the capital in just three to five days; they have worked hard." He added, "Juncheng's injuries are gradually healing, but helplessly under the wind and snow, the troops have to be stationed in Helilun. Once spring comes, their morale will be full. The Central Plains court must deploy its troops and can delay no longer."......\\
\textbf{Summary}: ...... Juncheng's injuries are gradually healing, and he is expected to march south to the Central Plains in the spring......\\

\textbf{Example 4}: Hallucination, Fabricated Event. Reason: "An inexplicable sense of satisfaction surged in Bixie's heart...... a wonderful sense of delight." This has no support in the original text.\\
\textbf{Article}: ...... "Does the master know that Gao Hou submitted a memorial pleading guilty today? He confessed to all the charges brought by the Ministry of Justice, stating that when he was in the Ministry of Revenue, he perverted the law for bribes, enriched himself at the public expense, and the stolen money was countless. Early this morning, people went to confiscate his property based on his confession in the memorial. The emperor finally breathed a sigh of relief, though still somewhat annoyed that he had shown off for so long, making it hard for the emperor to step down. It seems he is definitely dead." Bixie asked, "Have Gao Hou's family affairs been arranged?" "Yes," Jiang Fang said, "the stolen goods were planted in his house months ago." Bixie sneered: "This person framed my father years ago. Now his reputation is ruined, which is well-deserved retribution."......\\
\textbf{Summary}: ...... After the execution of Gao Hou, an inexplicable sense of satisfaction surged in Bixie's heart, feeling that he had finally avenged his father. This mood gave him a wonderful sense of delight......\\

\end{tcolorbox}
\vspace{-0.5em} 
\captionof{figure}{Human Annotation Rules (English Version).}
\label{fig:Annotation Rules(English Version)}
\vspace{1em} 


\clearpage

\begin{figure*}[!h]

\begin{tcolorbox}[promptbox,fontupper=\fontsize{10pt}{12pt}\selectfont]
\textbf{System Prompt}:\\
你是一名摘要幻觉领域的专家。\\
任务：\\
1. 阅读用户提供的摘要和参考实体。\\
2. 请仅引入实体幻觉，仅修改一句句子（不得引入其他幻觉），可以参考提供的实体，但是注意替换后和替换前的要不同：\\
   类型1：代词替换，在一句话内把代词指向错误对象，如“李强递给王伟一杯水，他连声道谢。”改成“李强递给王伟一杯水，李强连声道谢。”，“A骂了B，因为B迟到了。”改成“A骂了B，因为自己迟到了。”\\
   类型2：角色互换，将两位真实存在的人物在一句中互换身份（主语、宾语等），如将“A感谢B击退了敌人”改成“B感谢A击退了敌人”。注意，同时，“A和B一起吃饭”改为“B和A一起吃饭”是不可以的，因为两者是等价的，修改时要确保修改后的句子和修改前的句子不相同。\\
   类型3：组织/地名/称号错配：将人物与地名、称号或组织错配，如将“忽勒王子”写作“巨离忽王子”，或将“旭逯处”写作“汉军营地”等。\\
3. 除引入的实体错误以外，其余摘要内容必须与原文事实一致，风格统一，语义连贯。\\
4. 输出为以下JSON格式，仅输出JSON内容：\\
\{\\
  "type":"(类型1/类型2/类型3)",\\
  "error\_part": "原文：（正确的内容，并指出有幻觉的地方）；幻觉：（完整贴出发生实体幻觉的句子）"\\
  "hal\_summary": "引入实体幻觉后的摘要",\\
\}\\
5. 生成的“幻觉摘要”应保持与原摘要风格不变。注意：请确保引入幻觉后的句子和引入幻觉前的句子不相同。\\
\\
\textbf{User Prompt}:\\
摘要：\textbf{\textit{Summary Here}}\\
\\
参考实体：\textbf{\textit{Entities Here}}

\end{tcolorbox}
\caption{Prompt for Entity Hallucination Generation.}
\label{Prompt for Entity Hallucination Generation}
\end{figure*}

\begin{figure*}[!h]
\begin{tcolorbox}[promptbox,fontupper=\fontsize{10pt}{12pt}\selectfont]
\textbf{System Prompt}:\\
You are an expert in the field of summarization hallucinations.\\
Task:\\
1. Read the summary and reference entities provided by the user.\\
2. Introduce only Entity Hallucinations by modifying exactly one sentence (do not introduce any other types of hallucinations). You may refer to the provided entities, but ensure the modified version differs in meaning from the original:\\
   Type 1: Pronoun Replacement. Redirect a pronoun within a sentence to the wrong object. For example, change "Li Qiang handed Wang Wei a glass of water, and he thanked him" to "Li Qiang handed Wang Wei a glass of water, and Li Qiang thanked him".\\
   Type 2: Role Swapping. Swap the roles (subject, object, etc.) of two real individuals within a sentence. For example, change "A thanked B for defeating the enemy" to "B thanked A for defeating the enemy." Note: Changing "A and B ate together" to "B and A ate together" is not allowed because they are semantically equivalent. You must ensure the meaning of the modified sentence is different from the original.\\
   Type 3: Organization/Location/Title Mismatch. Mismatch a person with a location, title, or organization. For example, the summary places the confrontation and surrender of Richard at Berkeley Castle, but the original text identifies the location as Pontefract Castle.\\
3. Aside from the introduced entity error, the rest of the summary must remain factually consistent with the original text, maintaining a consistent style and coherent semantics.\\
4. Output strictly in the following JSON format (only output the JSON content):\\
\{\\
  "type": "(Type 1/Type 2/Type 3)",\\
  "error\_part": "Hallucination: (the complete sentence containing the entity hallucination); Original: (the correct content, pointing out where the hallucination occurs)",\\
  "hal\_summary": "The summary after introducing the entity hallucination"\\
\}\\
5. The generated "hallucinated summary" must maintain the same style as the original summary. Note: Ensure the sentence after introducing the hallucination is distinctly different from the original sentence.\\
\\
\textbf{User Prompt}:\\
Summary: \textbf{\textit{Summary Here}}\\
\\
Reference Entities: \textbf{\textit{Entities Here}}

\end{tcolorbox}
\caption{Prompt for Entity Hallucination Generation (English Version).}
\label{Prompt for Entity Hallucination Generation(English Version)}
\end{figure*}

\begin{figure*}[!h]
\begin{tcolorbox}[promptbox,fontupper=\fontsize{10pt}{12pt}\selectfont]
\textbf{System Prompt}:\\
你是一名摘要幻觉领域的专家。\\
任务：\\
1. 阅读用户提供的摘要和参考实体。\\
2. 在摘要中选择一句含有数字信息的句子，引入数字幻觉，注意替换后和替换前的要不同：\\
    类型1：例如更改原句中的数量、年龄、日期、金额等数字信息。例如“花了60分钟”改成“花了一分钟”。\\
    类型2：原概述中的日期x月y日修改为x个月后。\\
3. 除引入的数字幻觉以外，其余摘要内容必须与原文事实一致，风格统一，语义连贯。\\
4. 输出为以下JSON格式，仅输出JSON内容：\\
\{\\
  "type":"(类型1/类型2)",\\
  "error\_part": "原文：（正确的内容，并指出有幻觉的地方）；幻觉：（完整贴出发生数字幻觉的句子）",\\
  "hal\_summary": "(引入数字更改幻觉后的摘要)",\\
\}\\
5. 生成的“幻觉摘要”应保持与原摘要风格不变。注意：请确保引入幻觉后的句子和引入幻觉前的句子不相同。\\
\\
\textbf{User Prompt}:\\
摘要：\textbf{\textit{Summary Here}}\\
\\
参考实体：\textbf{\textit{Entities Here}}

\end{tcolorbox}
\caption{Prompt for Numerical Hallucination Generation.}
\label{Prompt for Numerical Hallucination Generation}
\end{figure*}

\begin{figure*}[!h]
\begin{tcolorbox}[promptbox,fontupper=\fontsize{10pt}{12pt}\selectfont]
\textbf{System Prompt}:\\
You are an expert in the field of summarization hallucination.\\
Task:\\
1. Read the summary and reference entities provided by the user.\\
2. Select a sentence in the summary that contains numerical information and introduce a numerical hallucination. Ensure that the modified sentence is different from the original.\\
\quad * Type 1: Modify numerical information such as quantity, age, date, amount, etc. For example, change "spent 60 minutes" to "spent one minute."\\
\quad * Type 2: Change a specific date (e.g., "Month X, Day Y") to a relative time frame (e.g., "X months later") or specify the setting as the early 'seventies', whereas the summary incorrectly states the setting as the early 1970s.\\
3. Except for the introduced numerical hallucination, all other content in the summary must remain consistent with the original facts, maintain a unified style, and be semantically coherent.\\
4. Output in the following JSON format (provide the JSON content only):\\
\{\\
  "type": "(Type 1/Type 2)",\\
  "error\_part": "Hallucination: (The complete sentence where the numerical hallucination occurs); Original: (The correct original content, pointing out where the hallucination was introduced)",\\
  "hal\_summary": "(The complete summary after introducing the numerical hallucination)"\\
\}\\
5. The generated "hallucinated summary" must maintain the same style as the original summary. Note: Ensure that the hallucinated sentence is strictly different from the original version.\\
\\
\textbf{User Prompt}:\\
Summary: \textbf{\textit{Summary Here}}\\
\\
Reference Entities: \textbf{\textit{Entities Here}}

\end{tcolorbox}
\caption{Prompt for Numerical Hallucination Generation (English Version).}
\label{Prompt for Numerical Hallucination Generation(English Version)}
\end{figure*}

\begin{figure*}[!h]
\centering
\begin{tcolorbox}[center,promptbox,fontupper=\fontsize{10pt}{12pt}\selectfont]
\textbf{System Prompt}:\\
你是一名摘要幻觉领域的专家。\\
任务：\\
1. 阅读用户提供的摘要和参考实体。\\
2. 请仅引入关系幻觉，在摘要仅选中一句（不得引入其他幻觉），进行关系错误的改写，仅修改一处句子，其他内容保持完全一致：\\
    类型1：更换关系或者身份，如“他的老师打电话把他送进医院”改为“他的父亲打电话把他送进医院”\\
    类型2：虚构关系，如“去看了关越的爷爷”改为“去看了关越的奶奶”。但要注意风格统一。\\
    上述构造的关系可以从参考实体列表中的关系。\\
3. 除该错误句子外，不得引入其他类型幻觉，保持内容一致、语义连贯、风格统一。\\
4. 输出为以下JSON格式，仅输出JSON内容：\\
\{\\
  "type":"(类型1/类型2)",\\
  "error\_part": "原文：（正确的内容，并指出有幻觉的地方）；幻觉：（完整贴出发生关系幻觉的句子）",\\
  "hal\_summary": "引入关系幻觉后的摘要",\\
  "is\_success": "True"\\
\}\\
\\
5. 生成的“幻觉摘要”应保持与原摘要风格不变。注意：详细内容替换成模糊内容是不正确的，例如“侄子”替换成“亲戚”是不对的。error\_part要和hal\_summary中的对应句子一致。请确保引入幻觉后的句子和引入幻觉前的句子不相同。\\
\\
\textbf{User Prompt}:\\
摘要：\textbf{\textit{Summary Here}}\\
\\
参考实体：\textbf{\textit{Entities Here}}

\end{tcolorbox}
\caption{Prompt for Relation Hallucination Generation.}
\label{Prompt for Relation Hallucination Generation}
\end{figure*}

\begin{figure*}[!h]
\centering
\begin{tcolorbox}[center,promptbox,fontupper=\fontsize{10pt}{12pt}\selectfont]
\textbf{System Prompt}:\\
You are an expert in the field of summarization hallucinations.\\
\\
Task:\\
1. Read the summary and reference entities provided by the user.\\
2. Introduce only relational hallucinations by selecting exactly one sentence in the summary (do not introduce any other types of hallucinations). Modify only one part of the sentence to create a relationship error, while keeping all other content completely identical:\\
    * Type 1: Replace relationship or identity. E.g., changing "His teacher called and sent him to the hospital" to "His father called and sent him to the hospital."\\
    * Type 2: Fabricate a relationship. E.g., changing "A and B are friends" to "A and B are cousins". Maintain a consistent style.\\
    * The constructed relationships can be drawn from the provided reference entity list.\\
3. Aside from the erroneous sentence, do not introduce any other types of hallucinations. Maintain consistent content, coherent semantics, and a unified style.\\
4. Output Format: Provide only the JSON content in the following format:\\
\{\\
  "type": "(Type 1/Type 2)",\\
  "error\_part": "Hallucination: (The complete sentence where the relational hallucination occurs); Original: (The correct content, indicating where the hallucination was introduced)",\\
  "hal\_summary": "The summary after introducing the relational hallucination",\\
  "is\_success": "True"\\
\}\\
\\
5. The generated "hallucinated summary" must maintain the same style as the original. Note: Replacing specific details with vague descriptions is incorrect (e.g., replacing "nephew" with "relative" is not allowed). The `error\_part` must match the corresponding sentence in `hal\_summary`. Ensure that the hallucinated sentence is different from the original sentence.\\
\\
\textbf{User Prompt}:\\
Summary: \textbf{\textit{Summary Here}}\\
\\
Reference Entities: \textbf{\textit{Entities Here}}

\end{tcolorbox}
\caption{Prompt for Relation Hallucination Generation (English Version).}
\label{Prompt for Relation Hallucination Generation(English Version)}
\end{figure*}

\begin{figure*}[!h]

\begin{tcolorbox}[center,promptbox,fontupper=\fontsize{10pt}{12pt}\selectfont]
\textbf{System Prompt}:\\
你是一名摘要幻觉领域的专家。\\
任务：\\
1. 阅读用户提供的摘要。\\
2. 请仅引入反向陈述幻觉，在摘要仅选中一句，进行反向陈述的改写，仅修改一处句子，其他内容保持完全一致：\\
    类型1：把肯定的改成否定的，如“A被B不卑不亢的气势折服了”改成“A始终没有被B的气势折服”，\\
    类型2：把否定的改成肯定的，但要注意语句通顺，逻辑转折自洽。如“A不死心，给B一封信”改成“A死心了，给B一封信”、“A看了看礼物，转身走了”改成“A买下了礼物”、“A没有被成功救出”改为“A被成功救出”。\\
3. 除该错误句子外，不得引入其他类型幻觉，保持内容一致、语义连贯、风格统一。\\
4. 输出为以下JSON格式，仅输出JSON内容：\\
\{\\
  "type":"(类型1/类型2)",\\
  "error\_part": "原文：（正确的内容，并指出有幻觉的地方）；幻觉：（完整贴出发生反向陈述的句子）"\\
  "hal\_summary": "引入反向陈述后的摘要",\\
\}\\
\\
\\5. 生成的“幻觉摘要”应保持与原摘要风格不变。注意：error\_part要和hal\_summary中的对应句子一致。请确保引入幻觉后的句子和引入幻觉前的句子不相同。\\
\\
\textbf{User Prompt}:\\
摘要：\textbf{\textit{Summary Here}}\\
\\
参考实体：\textbf{\textit{Entities Here}}

\end{tcolorbox}

\caption{Prompt for Logical Inversion Generation.}
\label{Prompt for Logical Inversion Generation}
\end{figure*}

\begin{figure*}[!h]
\begin{tcolorbox}[center,promptbox,fontupper=\fontsize{10pt}{12pt}\selectfont]
\textbf{System Prompt}:\\
You are an expert in the field of summarization hallucination.\\
\\
Task:\\
1. Read the summary provided by the user.\\
2. Introduce only intrinsic contradiction hallucinations (reversal of statements). Select only one sentence from the summary to rewrite; modify only that single sentence and keep all other content exactly the same:\\
    * Type 1: Change an affirmative statement to a negative one. For example, "the secret is successfully kept forever" becomes "the secret is revealed"\\
    * Type 2: Change a negative statement to an affirmative one, ensuring the sentence remains fluent and logically coherent. For example, "A is one of the few who did not go" becomes "A attends the Fair"\\
3. Do not introduce any other types of hallucinations except for this single erroneous sentence. Maintain consistency in content, semantic coherence, and style.\\
4. Output in the following JSON format (output only the JSON content):\\
\{\\
  "type": "(Type 1/Type 2)",\\
  "error\_part": "Hallucination: (the full modified sentence); Original: (the original correct content, indicating where the hallucination occurs)",\\
  "hal\_summary": "The full summary after introducing the reversed statement"\\
\}\\
\\
5. The generated "hallucinated summary" must maintain the same style as the original summary. Note: The `error\_part` must match the corresponding sentence in the `hal\_summary`. Ensure the hallucinated sentence is strictly different from the original sentence.\\
\\
\textbf{User Prompt}:\\
Summary: \textbf{\textit{Summary Here}}\\
\\
Reference Entities: \textbf{\textit{Entities Here}}

\end{tcolorbox}

\caption{Prompt for Logical Inversion Generation (English Version).}
\label{Prompt for Logical Inversion Generation(English Version)}
\end{figure*}

\begin{figure*}[!h]
\centering
\begin{tcolorbox}[promptbox,fontupper=\fontsize{10pt}{12pt}\selectfont]
\textbf{System Prompt}:\\
你是一名摘要幻觉领域的专家。\\
任务：\\
1. 阅读用户提供的摘要。\\
2. 请仅引入事件幻觉，在摘要仅选中一句，进行事件错误的改写，仅修改一处句子，其他内容保持完全一致：\\
    类型1：替换事件动词，例如“商讨”改为“大打出手”，“主动发现”改为“被动得知”，“A用剑杀了B”改为“A毒杀了B”；\\
    类型2：替换结果，例如“A先死了B一人打败C，只有B生还”修改为“A和B一起击败C，两人都平安归来”。\\
    类型3：添加或替换成无依据的心理、动机、评价，例如“A赶忙同意了”改为“A犹豫了一段时间，最后同意了”。\\
3. 除该错误句子外，不得引入其他类型幻觉，保持内容一致、语义连贯、风格统一。\\
4. 输出为以下JSON格式，仅输出JSON内容：\\
\{\\
  "type":"(类型1/类型2/类型3)",\\
  "error\_part": "原文：（正确的事件内容，并指出有幻觉的地方）；幻觉：（完整贴出事件幻觉的句子）",\\
  "hal\_summary": "引入事件幻觉后的摘要",\\
  "is\_success": "True"\\
\}\\

5. 生成的“幻觉摘要”应保持与原摘要风格不变。注意：error\_part要和hal\_summary中的对应句子一致。请确保引入幻觉后的句子和引入幻觉前的句子不相同。
注意：不要变为反向陈述，不是将做了变为没做，而是动词替换和事件结果替换。
\\
\textbf{User Prompt}:\\
摘要：\textbf{\textit{Summary Here}}\\
\\
参考实体：\textbf{\textit{Entities Here}}

\end{tcolorbox}
\caption{Prompt for Event Hallucination Generation.}
\label{Prompt for Event Hallucination Generation}
\end{figure*}

\begin{figure*}[!h]
\centering
\begin{tcolorbox}[promptbox,fontupper=\fontsize{10pt}{12pt}\selectfont]
\textbf{System Prompt}:\\
You are an expert in the field of summarization hallucinations.\\
\\
Task:\\
Read the summary provided by the user.\\
Please introduce only an event hallucination. Select exactly one sentence in the summary and rewrite it to contain an event error. Modify only this single sentence, keeping all other content entirely unchanged:\\
\\
Type 1: Replace event verbs. For example, change "opening the coffin with a screwdriver to discover the truth" to "praying by the coffin and discovering the truth."\\
Type 2: Replace outcomes. For example, change "confronting her about religious backsliding and obsession" to "visiting to make a second marriage proposal."\\
Type 3: Add or replace with unsubstantiated psychology, motivations, or evaluations. For example, change "A quickly agreed" to "A hesitated for a while before finally agreeing".\\
Type 4: Merge events from different times and locations together.\\
\\
With the exception of this erroneous sentence, you must not introduce any other types of hallucinations. Maintain content consistency, semantic coherence, and a unified style.\\
Output in the following JSON format, providing only the JSON content:\\
\\
\{\\
  "type": "(Type 1/Type 2/Type 3/Type 4)",\\
  "error\_part": "Hallucination: (Provide the complete sentence with the event hallucination); Original: (Provide the correct event content, and point out where the hallucination is)",\\
  "hal\_summary": "The summary after introducing the event hallucination",\\
  "is\_success": "True"\\
\}\\
\\
The generated "hallucinated summary" should maintain the exact same style as the original summary. Note: The error\_part must strictly match the corresponding sentence in the hal\_summary. Please ensure that the sentence after introducing the hallucination is fundamentally different from the sentence before.\\
Note: Do not simply convert the sentence into a negative statement (i.e., do not change "did" to "did not"). Focus strictly on replacing verbs and event outcomes.\\
\\
\textbf{User Prompt}:\\
Summary: \textbf{\textit{Summary Here}}\\
\\
Reference Entities: \textbf{\textit{Entities Here}}

\end{tcolorbox}
\caption{Prompt for Event Hallucination Generation (English Version.)}
\label{Prompt for Event Hallucination Generation(English Version)}
\end{figure*}

\begin{figure*}[!h]
\begin{tcolorbox}[promptbox,fontupper=\fontsize{10pt}{12pt}\selectfont]
\textbf{System Prompt}:\\
你是一名摘要幻觉领域的专家。\\
任务：\\
1. 阅读用户提供的摘要。\\
2. 首先判断summary中是否能提取出≥2个具有先后关系的事件，如果没有，直接输出{"is\_success": "False"}；如有，在摘要中选择两件关键事件交换先后顺序，使时间逻辑被打乱。\\
3. 不得引入其他类型幻觉，保持内容一致、语义连贯、风格统一。\\
4. 输出为以下 JSON 格式，仅输出 JSON 内容：\\
\{\\
  "error\_part": "原文：（正确顺序是什么，并指出有幻觉的地方）；幻觉：（陈述哪两件事件被交换顺序了）",\\
  "hal\_summary": "引入时间线幻觉后的摘要",\\
  "is\_success": "True"\\
\}\\
\\
5. 生成的“幻觉摘要”应保持与原摘要风格不变。注意：error\_part要和hal\_summary中的对应句子一致。请确保引入幻觉后的句子和引入幻觉前的句子不相同。\\
\\
\textbf{User Prompt}:\\
摘要：\textbf{\textit{Summary Here}}\\
\\
参考实体：\textbf{\textit{Entities Here}}

\end{tcolorbox}
\caption{Prompt for Temporal Hallucination Generation.}
\label{Prompt for Temporal Hallucination Generation}
\end{figure*}

\begin{figure*}[!h]
\begin{tcolorbox}[promptbox,fontupper=\fontsize{10pt}{12pt}\selectfont]
\textbf{System Prompt}:\\
You are an expert in the field of summarization hallucination.\\
\\
Task:\\
1. Read the summary provided by the user.\\
2. First, determine whether $\ge 2$ events with a sequential relationship can be extracted from the summary. If not, output `{"is\_success": "False"}` directly. If so, select two key sequential events within the summary and swap their order to disrupt the temporal logic.\\
3. Do not introduce any other types of hallucinations. Maintain consistent content, semantic coherence, and a unified style.\\
4. Output in the following JSON format, providing only the JSON content:\\
\{\\
  "error\_part": "Hallucination: (State which two events were swapped); Original: (State the correct order and point out the hallucinated part)",\\
  "hal\_summary": "The summary after introducing the temporal hallucination",\\
  "is\_success": "True"\\
\}\\
\\
5. The generated "hallucinated summary" should maintain the same style as the original. Note: The `error\_part` must correspond to the sentences in the `hal\_summary`. Ensure that the sentence containing the hallucination is different from the original sentence.\\
\\
\textbf{User Prompt}:\\
Summary: \textbf{\textit{Summary Here}}\\
\\
Reference Entities: \textbf{\textit{Entities Here}}

\end{tcolorbox}
\caption{Prompt for Temporal Hallucination Generation (English Version).}
\label{Prompt for Temporal Hallucination Generation(English Version)}
\end{figure*}

\begin{figure*}[t]
\centering 
\begin{tcolorbox}[promptbox,fontupper=\fontsize{9pt}{11pt}\selectfont]
\textbf{System Prompt}:\\
你是一名摘要幻觉领域的专家。\\
任务：\\
1. 阅读用户提供的摘要和参考实体。\\
2. 首先判断summary中是否能提取出至少一条因果逻辑链或者不同时间段的没有因果联系的事件。如果没有，直接输出{"is\_success": "False"}。如果有，进行因果链伪造，其他内容保持完全一致：\\
    类型1：如果能提取出至少一条因果逻辑链，把结果和原因倒置。原本 A → B 的逻辑链，被改写成 B → A。例子：原文：均成夜袭敌营 → 东胡混乱 → 大军乘胜进攻。幻觉：“大军发起进攻，所以均成夜袭敌营。”\\
    类型2:如果能提取出不同时间段的没有因果联系的事件，则把不同时间段的事件且结合起来作为因果，发生较前的为因，发生较后的为果。例如“A在花店买了一束花。B晚上请C吃饭。”修改为“A在花店买了一束花，导致B晚上请C吃饭。”\\
3. 不得引入其他类型幻觉，保持内容一致、语义连贯、风格统一。\\
4. 输出为以下JSON格式，仅输出JSON内容：\\
\{\\
  "error\_part": "  "error\_part": "原文：（正确的内容，并指出有幻觉的地方）；幻觉：（完整贴出发生因果链伪造的句子）",\\
  "hal\_summary": "引入事件因果幻觉后的摘要",\\
  "is\_success": "True"\\
\}\\
\\
5. 生成的“幻觉摘要”应保持与原摘要风格不变。注意：error\_part要和hal\_summary中的对应句子一致。请确保引入幻觉后的句子和引入幻觉前的句子不相同。\\
\\
\textbf{User Prompt}:\\
摘要：\textbf{\textit{Summary Here}}\\
\\
参考实体：\textbf{\textit{Entities Here}}

\end{tcolorbox}
\caption{Prompt for Causal Hallucination Generation.}
\label{Prompt for Causal Hallucination Generation}
\end{figure*}

\begin{figure*}[t]
\centering 
\begin{tcolorbox}[promptbox,fontupper=\fontsize{9pt}{11pt}\selectfont]
\textbf{System Prompt}:\\
You are an expert in the field of Summarization Hallucination.\\
\\
\#\# Task Description\\
\\
1.  Read the summary and reference entities provided by the user.\\
2.  Evaluate Logic: Determine if the summary contains at least one causal chain or a sequence of unrelated events occurring at different times.\\
    * If neither exists, output: `{"is\_success": "False"}`.\\
    * If they exist, perform Causal Fabrication while keeping all other content identical:\\
        * Type 1 (Causal Reversal): If a causal chain ($A \rightarrow B$) exists, reverse the cause and effect ($B \rightarrow A$).\\
            * *Example:* Original: "Juncheng raided the camp $\rightarrow$ Donghu fell into chaos $\rightarrow$ The army attacked." Hallucination: "The army launched an attack, which led to Juncheng raiding the enemy camp."\\
        * Type 2 (Temporal-to-Causal): If there are unrelated events occurring in different time periods, link them as cause and effect (earlier event = cause, later event = effect).\\
            * *Example:* "A bought a bouquet at the flower shop. B invited C to dinner in the evening." Hallucination: "A bought a bouquet at the flower shop, which caused B to invite C to dinner in the evening."\\
3.  Constraints: Do not introduce any other types of hallucinations. Maintain consistent content, semantic coherence, and style.\\
4.  Output Format: Provide the response strictly in the following JSON format:\\
\{\\
  "error\_part": "Hallucination: (The complete sentence where the causal fabrication occurs); \\Original: (The correct content, specifying where the hallucination lies)",\\
  "hal\_summary": "The summary after introducing the causal hallucination",\\
  "is\_success": "True"\\
\}\\
\\
5.  Quality Control: The generated "Hallucinated Summary" must maintain the same style as the original. The `error\_part` must match the corresponding sentence in the `hal\_summary`. Ensure that the modified sentence is distinctly different from the original.\\
\\
\textbf{User Prompt}:\\
Summary: \textbf{\textit{Summary Here}}\\
\\
Reference Entities: \textbf{\textit{Entities Here}}

\end{tcolorbox}
\caption{Prompt for Causal Hallucination Generation (English Version).}
\label{Prompt for Causal Hallucination Generation(English Version)}
\end{figure*}


\begin{figure*}[t]
\begin{tcolorbox}[promptbox,fontupper=\fontsize{9pt}{11pt}\selectfont]
\textbf{System Prompt}:\\
你是一名摘要幻觉领域的专家。\\
任务：\\
1. 阅读用户提供的摘要和参考实体。\\
2. 请仅引入虚构事件，加入虚构的一句话，其他内容保持完全一致：\\
    类型1：在某个事件结束，续写事件后续内容；\\
    类型2：将摘要前半部分已发生过的事件，在后半部分重复发生一次；\\
3. 除该错误句子外，不得引入其他类型幻觉，保持内容一致、语义连贯、风格统一。\\
4. 输出为以下 JSON 格式，仅输出 JSON 内容：\\
\{\\
  "type":"(类型1/类型2)",\\
  "error\_part": "原文：（正确的内容，并指出有幻觉的地方）；幻觉：（完整贴出有虚构事件的句子）",\\
  "hal\_summary": "引入虚构事件后的摘要"\\
\}\\
\\
5. 生成的“幻觉摘要”应保持与原摘要风格不变，例如中文玄幻风格的摘要不能出现西洋科幻风格的人物，幻觉内容需具备较强“迷惑性”而非显而易见的错误。注意：error\_part 要和 hal\_summary 中的对应句子一致。请确保引入幻觉后的句子和引入幻觉前的句子不相同。\\
\\
\textbf{User Prompt}:\\
摘要：\textbf{\textit{Summary Here}}\\
\\
参考实体：\textbf{\textit{Entities Here}}

\end{tcolorbox}

\caption{Prompt for Event Fabrication Generation.}
\label{Prompt for Event Fabrication Generation}
\end{figure*}

\begin{figure*}[t]
\begin{tcolorbox}[promptbox,fontupper=\fontsize{9pt}{11pt}\selectfont]
\textbf{System Prompt}:\\
You are an expert in the field of summarization hallucination.\\
\\
Task:\\
1. Read the summary and reference entities provided by the user.\\
2. Introduce a fabricated event by adding exactly one fictional sentence. Keep all other content strictly consistent with the original.\\
    * Type 1: After a specific event concludes, write a continuation describing subsequent developments (do not use temporal markers like "at this time," "afterward," etc.).\\
    * Type 2: Take an event that already occurred in the first half of the summary and describe it occurring again in the second half (do not use hint words like "again," "re-," or "once more"; describe it as if it were happening for the first time).\\
3. Aside from this specific erroneous sentence, do not introduce any other types of hallucinations. Maintain consistent content, semantic coherence, and a unified style.\\
4. Output in the following JSON format, providing only the JSON content:\\
\{\\
  "type": "(Type 1/Type 2)",\\
  "error\_part": "Hallucination: (The complete sentence containing the fabricated event); Original: (The correct content, pointing out the hallucinated part)",\\
  "hal\_summary": "The summary after introducing the fabricated event"\\
\}\\
\\
5. The generated "hallucinated summary" must maintain the original style. For example, a Chinese Xuanhuan (fantasy) summary should not feature Western sci-fi characters. The hallucinated content should be highly "deceptive" rather than an obvious error. Note: The `error\_part` must match the corresponding sentence in the `hal\_summary`. Ensure that the sentence containing the hallucination is entirely new and does not exist in the original text.\\
\\
\textbf{User Prompt}:\\
Summary: \textbf{\textit{Summary Here}}\\
\\
Reference Entities: \textbf{\textit{Entities Here}}

\end{tcolorbox}

\caption{Prompt for Event Fabrication Generation (English Version).}
\label{Prompt for Event Fabrication Generation(English Version)}
\end{figure*}

\begin{figure*}[ht]
\begin{tcolorbox}[promptbox,fontupper=\fontsize{10pt}{12pt}\selectfont]
\textbf{System Prompt}:\\
你是一名专业的概述一致性检查员。请对比原文与摘要，评估摘要是否准确还原了原文。\\
只检查摘要中是否存在幻觉，遗漏不扣分。\\
\\
注意：\\
1. 叙事一致性：在文学作品中，如果概述描述的是故事前期的表象，即使后期这些表象被推翻，只要前期确实有过相关描写，均视为受支持。\\
2. 局部覆盖：概述可能只涵盖了上下文中的某一部分（例如只概述了前半段）。只要概述的内容在文中能找到对应，不要求涵盖全篇反转。\\
3. 细节核实：只要有提到即为支持。请仔细核对文中每一个细节，不要因为后期情节剧变而否定前期的事实描写。\\
\\
评分标准（0-5）：\\
5分：完全一致，事实均可在检索上下文中找到依据\\
4分：高度一致，存在过度推论或语义轻微偏移\\
3分：整体事实正确，但存在对象泛化/语气程度变化/因果简化\\
2分：存在明显幻觉（如时间、人名、数字、事件错误）\\
1分：大部分内容为捏造，存在大量虚构事件\\
0分：完全不相关或大量严重虚假事实\\
如果分数大于等于3分，输出必须是严格JSON，格式如下：\\
\{\\
  "score": 0-5的整数,\\
  "reason": "打分理由"\\
\}\\
\\
如果分数低于3分，则幻觉类型必须从以下集合中选择0到多个：\\
(1)实体幻觉：代词指代错误，即在一段话中将代词错误地指向其他对象；角色互换，即在事件中交换两个角色的主宾关系；实体错配，即将人物、组织或地名替换成错误的信息。\\
(2)数字幻觉：对数字进行了不准确的更改，包括但不限于数量、年龄、时间、金额等数值。\\
(3)关系幻觉：关系身份更换，例如将“老师”改成“父亲”；虚构关系，即为原文中没有关系说明的人物添加亲属、师徒等关系。\\
(4)反向陈述：将肯定句改为否定句，或将否定句改为肯定句。\\
(5)事件幻觉：动词替换，例如将“谈话”替换为“争吵”；事件结果更改，例如将“被释放”改为“被拘留”；\\
(6)时间线乱序：原文中两个或多个事件发生的先后顺序。\\
(7)因果链伪造：虚构因果链，将无逻辑关系的事件强行连接为因果关系；因果倒置，即将原文的“结果”事件表述为“原因”；因果替换，即保留事实“结果”，但将“原因”替换为不相关的事件。注意：概述捕获的任一原因，无论直接原因、间接原因还是根本原因，都算作无幻觉。对于上下文中具有高度逻辑必然性，或者关联词（如“果然”）强烈暗示的行为，概述将其作为连接原因，算作无幻觉。\\
(8)虚构事件：人物做了一件原文中没有提及的事；伪造心理或情绪，捏造人物的心理活动或情绪状态。\\
\\
输出必须是严格JSON，不要输出任何额外文本，格式如下：\\
\{\\
  "reason": "xx幻觉类型，指出关键不一致点；xx幻觉类型，指出关键不一致点；...",\\
  "score": 0-5的整数,\\
  "hallucination\_types": ["reason中提到的幻觉内容"]\\
\}\\
请在reason里输出全部有幻觉的地方。\\

\textbf{User Prompt}:\\
文章：\textbf{\textit{Article Here}}\\\\
\\
句子：\textbf{\textit{Summary Here}}\\\\
\end{tcolorbox}
\vspace{-14pt}
\caption{Prompt for RAG.}
\label{fig:Prompt for RAG}
\end{figure*}

\clearpage

\begin{tcolorbox}[promptbox, breakable, fontupper=\fontsize{10pt}{12pt}\selectfont]
\textbf{System Prompt}:\\
You are a professional summary consistency checker. Please compare the original text and the summary to evaluate whether the summary accurately reflects the original text.\\
Only check for hallucinations in the summary; omissions are not penalized.\\
\\
Note:\\
1. Narrative Consistency: In literary works, if the summary describes surface-level events from the early stages of the story, even if these events are overturned later, as long as there are relevant descriptions in the early stages, they are considered supported.\\
2. Partial Coverage: The summary may only cover a certain part of the context (e.g., only summarizing the first half). As long as the content of the summary finds a correspondence in the text, it is not required to cover full-text plot twists.\\
3. Detail Verification: Any mention constitutes support. Please carefully verify every detail in the text, and do not deny the factual descriptions of the early stages due to drastic plot changes later on.\\
\\
Scoring Criteria (0-5):\\
5 points: Completely consistent, all facts can be found and supported in the retrieved context.\\
4 points: Highly consistent, containing over-inferences or slight semantic shifts.\\
3 points: Overall facts are correct, but there is object generalization / changes in tone or degree / causal simplification.\\
2 points: Obvious hallucinations exist (e.g., errors in time, names, numbers, events).\\
1 point: Mostly fabricated, containing a large number of fictional events.\\
0 points: Completely irrelevant or a massive amount of severe false facts.\\
If the score is 3 or above, the output must be strictly in JSON format as follows:\\
\{\\
  "score": an integer from 0 to 5,\\
  "reason": "Reason for the score"\\
\}\\
\\
If the score is below 3, the hallucination types must be selected from the following set (0 or more):\\
(1) Entity Hallucination: Pronoun reference error, i.e., pointing a pronoun to the wrong object in a paragraph; Role swapping, i.e., swapping the subject-object relationship of two characters in an event; Entity mismatch, i.e., replacing a person, organization, or location with incorrect information.\\
(2) Numerical Hallucination: Inaccurate modifications to numbers, including but not limited to values like quantity, age, time, amount, etc.\\
(3) Relational Hallucination: Replacing relational identities, such as changing "teacher" to "father"; Fabricated relationships, i.e., adding kinship, master-apprentice, or other relations to characters with no relation stated in the original text.\\
(4) Reverse Statement: Changing an affirmative sentence to a negative sentence, or vice versa.\\
(5) Event Hallucination: Verb replacement, such as replacing "talking" with "arguing"; Altering the outcome of an event, such as changing "released" to "detained".\\
(6) Timeline Disruption: Disrupting the chronological order of two or more events from the original text.\\
(7) Causal Chain Fabrication: Fabricating a causal chain by forcibly connecting illogical events into a causal relationship; Causal inversion, i.e., stating the "result" event from the original text as the "cause"; Causal replacement, i.e., keeping the factual "result" but replacing the "cause" with an unrelated event. Note: Any cause captured by the summary, whether a direct, indirect, or root cause, is considered a non-hallucination. For behaviors that have high logical necessity in the context or are strongly implied by linking words (e.g., "as expected"), framing them as causal links in the summary is considered a non-hallucination.\\
(8) Fabricated Event: A character doing something not mentioned in the original text; Fabricating psychological or emotional states, i.e., inventing a character's mental activity or emotional state.\\
\\
The output must be strictly in JSON format, without any additional text, formatted as follows:\\
\{\\
  "reason": "[Hallucination type], point out key inconsistencies; [Hallucination type], point out key inconsistencies; ...",\\
  "score": an integer from 0 to 5,\\
  "hallucination\_types": ["hallucination content mentioned in the reason"]\\
\}\\
Please output all hallucinated parts in the reason.\\
\\
\textbf{User Prompt}:\\
Article: \textbf{\textit{Article Here}}\\\\
\\
Summary: \textbf{\textit{Summary Here}}\\\\
\end{tcolorbox}
\vspace{-8pt} 
\captionof{figure}{Prompt for RAG (English Version).}
\label{fig:Prompt for RAG(English Version)}
\vspace{1em} 

\begin{figure*}[ht]
\begin{tcolorbox}[promptbox,fontupper=\fontsize{10pt}{12pt}\selectfont]
\textbf{System Prompt}:\\
你是一名专业的概述一致性检查员。请对比原文与摘要，评估摘要是否准确还原了原文。\\
只检查摘要中是否存在幻觉，遗漏不扣分。\\
\\
评分标准（0-5）：\\
5分：完美契合，无任何幻觉\\
4分：高度一致，存在过度推论或语义轻微偏移\\
3分：整体事实正确，但存在对象泛化/语气程度变化/因果简化\\
2分：存在明显幻觉（如时间、人名、数字、事件错误）\\
1分：大部分内容为捏造，存在大量虚构事件\\
0分：完全不相关或大量严重虚假事实\\
如果分数大于等于3分，输出必须是严格JSON，格式如下：\\
\{\
  "score": 0-5的整数,\\
  "reason": "打分理由"\\
\}\\
\\
如果分数低于3分，则幻觉类型必须从以下集合中选择0到多个：\\
(1)实体幻觉：代词指代错误，即在一段话中将代词错误地指向其他对象；角色互换，即在事件中交换两个角色的主宾关系；实体错配，即将人物、组织或地名替换成错误的信息。\\
(2)数字幻觉：对数字进行了不准确的更改，包括但不限于数量、年龄、时间、金额等数值。\\
(3)关系幻觉：关系身份更换，例如将“老师”改成“父亲”；虚构关系，即为原文中没有关系说明的人物添加亲属、师徒等关系。\\
(4)反向陈述：将肯定句改为否定句，或将否定句改为肯定句。\\
(5)事件幻觉：动词替换，例如将“谈话”替换为“争吵”；事件结果更改，例如将“被释放”改为“被拘留”；\\
(6)时间线乱序：原文中两个或多个事件发生的先后顺序。\\
(7)因果链伪造：虚构因果链，将无逻辑关系的事件强行连接为因果关系；因果倒置，即将原文的“结果”事件表述为“原因”；因果替换，即保留事实“结果”，但将“原因”替换为不相关的事件。注意：概述捕获的任一原因，无论直接原因、间接原因还是根本原因，都算作无幻觉。对于上下文中具有高度逻辑必然性，或者关联词（如“果然”）强烈暗示的行为，概述将其作为连接原因，算作无幻觉。\\
(8)虚构事件：人物做了一件原文中没有提及的事；伪造心理或情绪，捏造人物的心理活动或情绪状态。\\
\\
输出必须是严格JSON，不要输出任何额外文本，格式如下：\\
\{\\
  "reason": "xx幻觉类型，指出关键不一致点；xx幻觉类型，指出关键不一致点；...",\\
  "score": 0-5的整数,\\
  "hallucination\_types": ["reason中提到的幻觉内容"]\\
\}\\
\\
请在reason里输出全部有幻觉的地方。\\
\\
\textbf{User Prompt}:\\
文章：\textbf{\textit{Article Here}}\\
\\
摘要：\textbf{\textit{Summary Here}}\\
\end{tcolorbox} 
\caption{Prompt template for zero-shot prompting with the target summary positioned at the End (Prompt-E).}
\label{fig: Prompt-E}
\end{figure*}

\begin{figure*}[ht]
\begin{tcolorbox}[promptbox,fontupper=\fontsize{10pt}{12pt}\selectfont]
\textbf{System Prompt}:\\
You are a professional Summary Consistency Inspector. Your task is to compare the original text with the summary and evaluate whether the summary accurately reflects the original content.\\
Only check for hallucinations in the summary; omissions do not result in point deductions.\\
\#\#\# Scoring Criteria (0-5):\\
*5 Points: Perfect match; no hallucinations.\\
*4 Points: Highly consistent; contains minor over-inference or slight semantic shifts.\\
*3 Points: Factually correct overall, but contains object generalization, changes in tone/intensity, or simplified causality.\\
*2 Points: Significant hallucinations present (e.g., errors in time, names, numbers, or events).\\
*1 Point: Mostly fabricated; contains a large number of fictional events.\\
*0 Points: Completely irrelevant or contains a vast amount of serious false facts.\\
---\\
\#\#\# Output Requirements:\\
If the score is 3 or higher, the output must be a strict JSON object in the following format:\\
\{\\
  "score": integer (0-5),\\
  "reason": "Reasoning for the score"\\
\}\\
If the score is below 3, you must select zero or more hallucination types from the following set:
1. Entity Hallucination: Pronoun reference error (wrongly assigning a pronoun); role reversal (swapping subject and object); entity mismatch (replacing people, organizations, or locations with incorrect info).\\
2. Numerical Hallucination: Inaccurate changes to numbers, including quantity, age, time, currency, etc.\\
3. Relational Hallucination: Relationship identity change (e.g., changing "teacher" to "father"); fictional relationships (adding relationships like kinship or mentorship not in the text).\\
4. Logical Inversion: Changing an affirmative sentence to negative, or vice versa.\\
5. Event Hallucination: Verb replacement (e.g., changing "talked" to "quarreled"); changing event outcomes (e.g., changing "released" to "detained").\\
6. Temporal Hallucination: Misordering the sequence of two or more events from the original text.\\
7. Causal Hallucination: Fictional causality (connecting unrelated events as cause-and-effect); causal inversion (treating a result as a cause); causal replacement (keeping the result but replacing the cause with something unrelated). *Note: Capturing any cause (direct, indirect, or root) is not a hallucination. Logical inevitability or strong implications (e.g., "as expected") used as links are not hallucinations.*\\
8. Event Fabrication: A character performing an action not mentioned in the text; fabricated psychology/emotion (inventing mental states or moods).\\
The output must be a strict JSON object with no additional text, formatted as follows:\\
\{\\
  "reason": "Type of hallucination, pointing out the key inconsistency; Type of hallucination...",\\
  "score": integer (0-5),\\
  "hallucination\_types": ["Types mentioned in the reason"]\\
\}\\
\\
*Please ensure all hallucinated points are detailed in the "reason" field.\\
\\
\textbf{User Prompt}:\\
Article：\textbf{\textit{Article Here}}\\
\\
Summary：\textbf{\textit{Summary Here}}\\
\end{tcolorbox} 
\caption{Prompt template for zero-shot prompting with the target summary positioned at the End (Prompt-E)(English Version).}
\label{fig: Prompt-E(English Version)}
\end{figure*}

\begin{figure*}[ht]
\begin{tcolorbox}[promptbox,fontupper=\fontsize{10pt}{12pt}\selectfont]
\textbf{System Prompt}:\\
你是一名专业的概述一致性检查员。请对比原文与摘要，评估摘要是否准确还原了原文。\\
只检查摘要中是否存在幻觉，遗漏不扣分。\\
\\
评分标准（0-5）：\\
5分：完美契合，无任何幻觉\\
4分：高度一致，存在过度推论或语义轻微偏移\\
3分：整体事实正确，但存在对象泛化/语气程度变化/因果简化\\
2分：存在明显幻觉（如时间、人名、数字、事件错误）\\
1分：大部分内容为捏造，存在大量虚构事件\\
0分：完全不相关或大量严重虚假事实\\
如果分数大于等于3分，输出必须是严格JSON，格式如下：\\
\{\\
  "score": 0-5的整数,\\
  "reason": "打分理由"\\
\}\\
\\
如果分数低于3分，则幻觉类型必须从以下集合中选择0到多个：\\
(1)实体幻觉：代词指代错误，即在一段话中将代词错误地指向其他对象；角色互换，即在事件中交换两个角色的主宾关系；实体错配，即将人物、组织或地名替换成错误的信息。\\
(2)数字幻觉：对数字进行了不准确的更改，包括但不限于数量、年龄、时间、金额等数值。\\
(3)关系幻觉：关系身份更换，例如将“老师”改成“父亲”；虚构关系，即为原文中没有关系说明的人物添加亲属、师徒等关系。\\
(4)反向陈述：将肯定句改为否定句，或将否定句改为肯定句。\\
(5)事件幻觉：动词替换，例如将“谈话”替换为“争吵”；事件结果更改，例如将“被释放”改为“被拘留”；\\
(6)时间线乱序：原文中两个或多个事件发生的先后顺序。\\
(7)因果链伪造：虚构因果链，将无逻辑关系的事件强行连接为因果关系；因果倒置，即将原文的“结果”事件表述为“原因”；因果替换，即保留事实“结果”，但将“原因”替换为不相关的事件。注意：概述捕获的任一原因，无论直接原因、间接原因还是根本原因，都算作无幻觉。对于上下文中具有高度逻辑必然性，或者关联词（如“果然”）强烈暗示的行为，概述将其作为连接原因，算作无幻觉。\
(8)虚构事件：人物做了一件原文中没有提及的事；伪造心理或情绪，捏造人物的心理活动或情绪状态。\\
\\
输出必须是严格JSON，不要输出任何额外文本，格式如下：\\
\{\\
  "reason": "xx幻觉类型，指出关键不一致点；xx幻觉类型，指出关键不一致点；...",\\
  "score": 0-5的整数,\\
  "hallucination\_types": ["reason中提到的幻觉内容"]\\
\}\\
\\
\textbf{User Prompt}:\\
摘要：\textbf{\textit{Summary Here}}\\
\\
文章：\textbf{\textit{Article Here}}\\
\end{tcolorbox}
\caption{Prompt template for zero-shot prompting with the target summary positioned at the Beginning (Prompt-B).}
\label{fig: Prompt-B}
\end{figure*}

\begin{figure*}[ht]
\begin{tcolorbox}[promptbox,fontupper=\fontsize{10pt}{12pt}\selectfont]
\textbf{System Prompt}:\\
You are a professional Summary Consistency Inspector. Your task is to compare the original text with the summary and evaluate whether the summary accurately reflects the original content.\\
Only check for hallucinations in the summary; omissions do not result in point deductions.\\
\\
\#\#\# Scoring Criteria (0-5):\\
*5 Points: Perfect match; no hallucinations.\\
*4 Points: Highly consistent; contains minor over-inference or slight semantic shifts.\\
*3 Points: Factually correct overall, but contains object generalization, changes in tone/intensity, or simplified causality.\\
*2 Points: Significant hallucinations present (e.g., errors in time, names, numbers, or events).\\
*1 Point: Mostly fabricated; contains a large number of fictional events.\\
*0 Points: Completely irrelevant or contains a vast amount of serious false facts.\\
---\\
\#\#\# Output Requirements:\\
\\
If the score is 3 or higher, the output must be a strict JSON object in the following format:\\
\{\\
  "score": integer (0-5),\\
  "reason": "Reasoning for the score"\\
\}\\
If the score is below 3, you must select zero or more hallucination types from the following set:\\
1. Entity Hallucination: Pronoun reference error (wrongly assigning a pronoun); role reversal (swapping subject and object); entity mismatch (replacing people, organizations, or locations with incorrect info).\\
2. Numerical Hallucination: Inaccurate changes to numbers, including quantity, age, time, currency, etc.\\
3. Relational Hallucination: Relationship identity change (e.g., changing "teacher" to "father"); fictional relationships (adding relationships like kinship or mentorship not in the text).\\
4. Logical Inversion: Changing an affirmative sentence to negative, or vice versa.\\
5. Event Hallucination: Verb replacement (e.g., changing "talked" to "quarreled"); changing event outcomes (e.g., changing "released" to "detained").\\
6. Temporal Hallucination: Misordering the sequence of two or more events from the original text.\\
7. Causal Hallucination: Fictional causality (connecting unrelated events as cause-and-effect); causal inversion (treating a result as a cause); causal replacement (keeping the result but replacing the cause with something unrelated). *Note: Capturing any cause (direct, indirect, or root) is not a hallucination. Logical inevitability or strong implications (e.g., "as expected") used as links are not hallucinations.*\\
8. Event Fabrication: A character performing an action not mentioned in the text; fabricated psychology/emotion (inventing mental states or moods).\\
The output must be a strict JSON object with no additional text, formatted as follows:\\
\{\\
  "reason": "Type of hallucination, pointing out the key inconsistency; Type of hallucination...",\\
  "score": integer (0-5),\\
  "hallucination\_types": ["Types mentioned in the reason"]\\
\}\\
*Please ensure all hallucinated points are detailed in the "reason" field.\\
\textbf{User Prompt}:\\
Summary：\textbf{\textit{Summary Here}}\\
Article：\textbf{\textit{Article Here}}\\
\end{tcolorbox}
\caption{Prompt template for zero-shot prompting with the target summary positioned at the Beginning (Prompt-B) (English Version).}
\label{fig: Prompt-B(English Version)}
\end{figure*}

\begin{figure*}[ht]
\begin{tcolorbox}[promptbox,fontupper=\fontsize{10pt}{12pt}\selectfont]
\textbf{System Prompt}:\\
你是一名专业的概述一致性检查员。请对比原文与摘要，评估摘要是否准确还原了原文。\\
只检查摘要中是否存在幻觉，遗漏不扣分。\\
\\
请在输出回答前一步步输出分析过程，之后再输出json格式的回答。\\
\\
评分标准（0-5）：\\
5分：完美契合，无任何幻觉\\
4分：高度一致，存在过度推论或语义轻微偏移\\
3分：整体事实正确，但存在对象泛化/语气程度变化/因果简化\\
2分：存在明显幻觉（如时间、人名、数字、事件错误）\\
1分：大部分内容为捏造，存在大量虚构事件\\
0分：完全不相关或大量严重虚假事实\\
如果分数大于等于3分，输出必须是严格JSON，格式如下：\\
\{\\
  "score": 0-5的整数,\\
  "reason": "打分理由"\\
\}\\
\\
如果分数低于3分，则幻觉类型必须从以下集合中选择0到多个：\\
(1)实体幻觉：代词指代错误，即在一段话中将代词错误地指向其他对象；角色互换，即在事件中交换两个角色的主宾关系；实体错配，即将人物、组织或地名替换成错误的信息。\\
(2)数字幻觉：对数字进行了不准确的更改，包括但不限于数量、年龄、时间、金额等数值。\\
(3)关系幻觉：关系身份更换，例如将“老师”改成“父亲”；虚构关系，即为原文中没有关系说明的人物添加亲属、师徒等关系\\。
(4)反向陈述：将肯定句改为否定句，或将否定句改为肯定句。\\
(5)事件幻觉：动词替换，例如将“谈话”替换为“争吵”；事件结果更改，例如将“被释放”改为“被拘留”；\\
(6)时间线乱序：原文中两个或多个事件发生的先后顺序。\\
(7)因果链伪造：虚构因果链，将无逻辑关系的事件强行连接为因果关系；因果倒置，即将原文的“结果”事件表述为“原因”；因果替换，即保留事实“结果”，但将“原因”替换为不相关的事件。注意：概述捕获的任一原因，无论直接原因、间接原因还是根本原因，都算作无幻觉。对于上下文中具有高度逻辑必然性，或者关联词（如“果然”）强烈暗示的行为，概述将其作为连接原因，算作无幻觉。\\
(8)虚构事件：人物做了一件原文中没有提及的事；伪造心理或情绪，捏造人物的心理活动或情绪状态。\\
\\
输出必须是严格JSON，不要输出任何额外文本，格式如下：\\
\{\\
  "reason": "xx幻觉类型，指出关键不一致点；xx幻觉类型，指出关键不一致点；...",\\
  "score": 0-5的整数,\\
  "hallucination\_types": ["reason中提到的幻觉内容"]\\
\}\\
\\
请在reason里输出全部有幻觉的地方。\\
\\
\textbf{User Prompt}:\\
文章：\textbf{\textit{Article Here}}\\
\\
摘要：\textbf{\textit{Summary Here}}\\
\end{tcolorbox}
\caption{Prompt template for Chain-of-Thought prompting, designed to elicit step-by-step reasoning.}
\label{fig: Prompt template for Chain-of-Thought prompting}
\end{figure*}

\begin{figure*}[ht]
\begin{tcolorbox}[promptbox,fontupper=\fontsize{10pt}{12pt}\selectfont]
\textbf{System Prompt}:\\
You are a professional Summary Consistency Inspector. Your task is to compare the original text with the summary and evaluate whether the summary accurately reflects the original content.\\
\\
Only check for hallucinations in the summary; omissions do not result in point deductions.\\
\\
Please provide your analysis process step-by-step before outputting the final answer.\\
\\
\#\#\# Scoring Criteria (0-5):\\
*5 Points: Perfect match; no hallucinations.\\
*4 Points: Highly consistent; contains minor over-inference or slight semantic shifts.\\
*3 Points: Factually correct overall, but contains object generalization, changes in tone/intensity, or simplified causality.\\
*2 Points: Significant hallucinations present (e.g., errors in time, names, numbers, or events).\\
*1 Point: Mostly fabricated; contains a large number of fictional events.\\
*0 Points: Completely irrelevant or contains a vast amount of serious false facts.\\
---\\
\#\#\# Output Requirements:
\\
If the score is 3 or higher, the output must be a strict JSON object in the following format:\\
\{\\
  "score": integer (0-5),\\
  "reason": "Reasoning for the score"\\
\}\\
\\
If the score is below 3, you must select zero or more hallucination types from the following set:
1. Entity Hallucination: Pronoun reference error (wrongly assigning a pronoun); role reversal (swapping subject and object); entity mismatch (replacing people, organizations, or locations with incorrect info).\\
2. Numerical Hallucination: Inaccurate changes to numbers, including quantity, age, time, currency, etc.\\
3. Relational Hallucination: Relationship identity change (e.g., changing "teacher" to "father"); fictional relationships (adding relationships like kinship or mentorship not in the text).\\
4. Logical Inversion: Changing an affirmative sentence to negative, or vice versa.\\
5. Event Hallucination: Verb replacement (e.g., changing "talked" to "quarreled"); changing event outcomes (e.g., changing "released" to "detained").\\
6. Temporal Hallucination: Misordering the sequence of two or more events from the original text.\\
7. Causal Hallucination: Fictional causality (connecting unrelated events as cause-and-effect); causal inversion (treating a result as a cause); causal replacement (keeping the result but replacing the cause with something unrelated). *Note: Capturing any cause (direct, indirect, or root) is not a hallucination. Logical inevitability or strong implications (e.g., "as expected") used as links are not hallucinations.*\\
8. Event Fabrication: A character performing an action not mentioned in the text; fabricated psychology/emotion (inventing mental states or moods).\\
The output must be a strict JSON object with no additional text, formatted as follows:\\
\{\\
  "reason": "Type of hallucination, pointing out the key inconsistency; Type of hallucination...",\\
  "score": integer (0-5),\\
  "hallucination\_types": ["Types mentioned in the reason"]\\
\}\\
*Please ensure all hallucinated points are detailed in the "reason" field.\\
\textbf{User Prompt}:\\
Summary：\textbf{\textit{Summary Here}}\\
Article：\textbf{\textit{Article Here}}\\
\end{tcolorbox}
\caption{Prompt template for Chain-of-Thought prompting, designed to elicit step-by-step reasoning (English Version).}
\label{fig: Prompt template for Chain-of-Thought prompting(English Version)}
\end{figure*}

\clearpage

\end{document}